\documentclass[11pt]{article}

\usepackage[utf8]{inputenc} % allow utf-8 input
\usepackage[T1]{fontenc}    % use 8-bit T1 fonts
\usepackage{lmodern}
\usepackage[hidelinks]{hyperref}       % hyperlinks
\usepackage{url}            % simple URL typesetting
\usepackage{booktabs}       % professional-quality tables
\usepackage{microtype}      % microtypography

\usepackage[numbers]{natbib}

\usepackage{algorithm,algpseudocode}
\usepackage{amsfonts}
\usepackage{amsmath}
\usepackage{amsthm}
\usepackage{amssymb}
\usepackage{mathtools}
\usepackage[title]{appendix}
\usepackage{bm}
\usepackage{courier}
\usepackage[usenames,dvipsnames]{color}
\usepackage{enumitem}
\usepackage{graphicx}
\usepackage[caption=false,font=footnotesize]{subfig}
\usepackage{url}
\usepackage{nicefrac}

\usepackage[margin=1in]{geometry}

\input{./Definitions}

\graphicspath{ {./figures/} }
\DeclareGraphicsExtensions{.pdf, .jpeg, .png}

\allowdisplaybreaks

\title{
Randomized Iterative Algorithms for
Fisher Discriminant Analysis
}

\author{
	Agniva Chowdhury%
	\thanks{Department of Statistics, Purdue University,
		West Lafayette, IN. Emails:\,\url{{chowdhu5, yang768}@purdue.edu}.}
	\and
	Jiasen Yang%
	\footnotemark[1]
	\and
	Petros Drineas%
	\thanks{Department of Computer Science, Purdue University,
		West Lafayette, IN. Email:~\url{pdrineas@purdue.edu}.}
}

\date{}

\begin{document}

\maketitle

%!TEX root = arxiv_LDA.tex

\begin{abstract}
Fisher discriminant analysis (FDA) is a widely used method for classification and dimensionality reduction. When the number of predictor variables greatly exceeds the number of observations, one of the alternatives for conventional FDA is regularized Fisher discriminant analysis (RFDA). In this paper, we present a simple, iterative, sketching-based algorithm for RFDA
that comes with provable accuracy guarantees when compared to the conventional approach.  Our analysis builds upon two simple structural results that boil down to randomized matrix multiplication, a fundamental and well-understood primitive of randomized linear algebra. We analyze the behavior of RFDA when the ridge leverage and the standard leverage scores are used to select predictor variables and we prove that accurate approximations can be achieved by a sample whose size depends on the effective degrees of freedom of the RFDA problem. Our results yield significant improvements over existing approaches and our empirical evaluations support our theoretical analyses.
\end{abstract}

%!TEX root = arxiv_LDA.tex

\section{Introduction}\label{sxn:intro}
In multivariate statistics and machine learning, Fisher's linear discriminant analysis (FDA) is a widely used method for classification and dimensionality reduction. The main idea is to project the data onto a lower dimensional space such that the separability of points \textit{between} the different classes is maximized while the separability of points \textit{within} each class is minimized.

%\todo{I still think it would be helpful to start with a paragraph (\eg, half-a-column) introducing Fisher's LDA in the conventional setup and showing how that leads to the eigen-problem (see Wikipedia or Bishop's book or the intro to Zhang's/Ye's paper).
%Not only would this be helpful to readers who are not familiar with the eigen-formulation, this would also sound better than directly starting with saying that we build upon Ye's previous paper which makes our contribution seem incremental.
%}
Let $\Ab\in\RR{n}{d}$ be the \textit{centered} data matrix whose rows represent $n$ points in $\mathbb{R}^d$. We assume that $\Ab$ is centered around $\mb\in\R{d}$, with $\mb$ being the \emph{grand-mean} of the original raw (non-centered) data-points.%
\footnote{If the original data were represented by the matrix $\widehat{\Ab}\in \mathbb{R}^{n \times d}$, then $\mb$ is the row-wise mean of $\widehat{\Ab}$ and $\Ab = \widehat{\Ab} - \mathbf{1}_n \mb^\ts$, where $\mathbf{1}_n$ is the all-ones vector. As a result of mean-centering, $\rank(\Ab)\le \min\{n-1,d-1\}$.}
Suppose there are $c$ disjoint classes with $n_j$ observations belonging to the $j$-th class and  $\sum_{j=1}^cn_j=n$.
Further, let $\mb_j\in\R{d}$ denote the mean vector of the raw (non-centered) data-points corresponding to the $j$-th class, $j=1,2,\dots, c$.
Define the \emph{total scatter matrix}
\[
\Sigmab_t \defeq \sum_{i=1}^{n}(\ab_i-\mb)(\ab_i-\mb)^\ts\in\RR{d}{d},
\]
where $\ab_i$ is the $i$-th raw data-point. Similarly, define the
\emph{between scatter matrix}
\[
\Sigmab_b \defeq \sum_{j=1}^{c}n_j(\mb_j-\mb)(\mb_j-\mb)^\ts\in\RR{d}{d}.
\]
%\todo{The above were for un-centered $\Ab$, right?}
%
Under these notations, conventional FDA solves the generalized eigen-problem
\[
\Sigmab_b\,\xb_i=\lambda_i\,\Sigmab_t\,\xb_i, \quad i=1,2,\dots q,
\]
where $\xb_i$ is called the $i$-th \emph{discriminant direction},
with $q\le\min\{d,c-1\}$ and $\lambda_1\ge\lambda_2\ge\cdots\ge\lambda_q>0$\,. We can further express this problem in matrix form as
\begin{flalign}
	\Sigmab_b\Xb=\Sigmab_t\,\Xb\Lambdab,\label{eq:fda_preq}
\end{flalign}
where $\Xb \defeq \begin{bmatrix*}[c]
\xb_1 & \xb_2 &\cdots & \xb_q
\end{bmatrix*}\in\RR{d}{q}$ and $\Lambdab \defeq \diag\{\lambda_1,\dots,\lambda_q\}$\,. An elegant linear algebraic formulation of eqn.~\eqref{eq:fda_preq} was presented in~\cite{ZDXJ10}:
\begin{flalign}
	(\Ab^\ts\Omegab\Omegab^\ts\Ab)\Xb=(\Ab^\ts\Ab)\,\Xb\Lambdab,\label{eq:fda}
\end{flalign}
where
%$\Ab=\Hb\Ab$ is the centered data matrix and $\Hb=\Ib_n-\frac{1}{n}\mathbf{1}_n\mathbf{1}_n^\ts\in\RR{n}{n}$ is the corresponding centering matrix
%\footnote{Note that $\Hb$ is symmetric, idempotent and $\rank(\Hb)=n-1$ which implies $\rank(\Ab)\le n-1$. Therefore, $\Ab$ can not be full rank and we assume it to be $\rho$.}\,.
$\Sigmab_t=\Ab^\ts\Ab$ and $\Sigmab_b=\Ab^\ts\Omegab\Omegab^\ts\Ab$.
%\todo{We need to be consistent with the above.}
Here, $\Omegab\in\RR{n}{c}$ denotes the rescaled \emph{class membership matrix}, with $\Omegab_{ij}=1/\sqrt{n_j}$ if the $i$-th row of $\Ab$ (\ie, the $i$-th data point) is a member of the $j$-th class; otherwise $\Omegab_{ij}=0$.

If $\Ab^\ts\Ab$ is non-singular, then the pairs $(\lambda_i,\xb_i)$ for $i=1,2, \dots, q$ are the eigen-pairs of the matrix $(\Ab^\ts\Ab)^{-1}\Ab^\ts\Omegab\Omegab^\ts\Ab$. However, in many applications, such as micro-array analysis~\cite{GHT07}, information retrieval~\cite{Deerwester90}, face recognition~\cite{EC97,ZY08}, etc, the underlying $\Ab^\ts\Ab$ is ill-conditioned as the number of predictors greatly exceeds the number of observations, i.e., $d\gg n$. This makes the computation of $(\Ab^\ts\Ab)^{-1}\Ab^\ts\Omegab\Omegab^\ts\Ab$ numerically unstable. A popular alternative to FDA that addresses this problem
% of $\Ab^\ts\Ab$ being ill-conditioned
is \emph{regularized Fisher discriminant analysis}~(RFDA)~\cite{Friedman89,GHT07}.%
\footnote{We note that another variant is pseudo-inverse FDA~\cite{WEBB02}, which replaces
	$(\Ab^\ts\Ab)^{-1}$ by $(\Ab^\ts\Ab)^{\dagger}$.}

% \paragraph{Regularized Fisher discriminant analysis (RFDA).}
In RFDA, $(\Ab^\ts\Ab)^{-1}$ is replaced by $(\Ab^\ts\Ab+\lambda\Ib_d)^{-1}$, where $\lambda>0$ is a regularization parameter. In this case, eqn.~\eqref{eq:fda} becomes
\begin{flalign}
	\Gb~\Omegab^\ts\Ab\Xb=\Xb\Lambdab\label{eq:rfda3},
\end{flalign}
where $\Gb=(\Ab^\ts\Ab+\lambda\Ib_d)^{-1}\Ab^\ts\Omegab=\Ab^\ts(\Ab\Ab^\ts+\lambda\Ib_n)^{-1}\Omegab$. (The last equality can be easily verified using the SVD of $\Ab$.)
Note that the inverse of $\Ab^\ts\Ab+\lambda\Ib_d$ always exists for $\lambda>0$. We define the \textit{effective degrees of freedom} of RFDA as
\begin{flalign}
	d_\lambda= \sum_{i=1}^{\rho}\frac{\sigma_i^2}{\sigma_i^2+\lambda}\leq \rho\,. \label{eqn:df2}
\end{flalign}
Here $\rho$ is the rank of the matrix $\Ab$ and we note that $d_\lambda$ depends  on both the value of the regularization parameter $\lambda$ and $\sigma_i^2,~i=1,2,\dots,\rho$\,,~\ie~ the non-zero singular values of $\Ab$.

\emph{Solving the RFDA problem of eqn.~\eqref{eq:rfda3}.}
Notice that the solution $(\Xb, \Lambdab)$ to  eqn.~\eqref{eq:rfda3} may not be unique. Indeed, if $\Xb$ is a solution to eqn.~\eqref{eq:rfda3}, then for any non-singular diagonal matrix $\Db\in\RR{q}{q}$, $\Xb\Db$ is also a solution. \cite{ZDXJ10} proposed an eigenvalue decomposition~(EVD)-based algorithm (see Algorithm~\ref{algo:EVD_RFDA} in Appendix~\ref{app:evd}) which not only returns $\Xb$ as a solution to eqn.~\eqref{eq:rfda3} but also guarantees that for any two data points $\wb_1, \wb_2\in \R{d}$, $\Xb$ satisfies $\|(\wb_1-\wb_2)^\ts\Xb\|_2=\|(\wb_1-\wb_2)^\ts\Gb\|_2$ (see Theorem~\ref{thm:EVD_RFDA}). This implies that instead of using the actual solution $\Xb$, if we project the points using $\Gb$, the distances between the projected points would also be preserved. Thus, for any distance-based classification method (e.g., $k$-nearest-neighbors), both $\Xb$ and $\Gb$ would result in the same predictions. Therefore, when solving eqn.~(\ref{eq:rfda3}) it is reasonable to shift our interest from $\Xb$ to $\Gb$. However, due to the high dimensionality~$d$ of the input data, exact computation of $\Gb$ is expensive, taking time $\Ocal(n^2d+n^3+ndc)$.

%!TEX root = arxiv_LDA.tex

\subsection{Our Contributions}
\label{sec:contribs}
We present a simple, iterative, sketching-based algorithm for the  RFDA problem that guarantees highly accurate solutions when compared to conventional approaches.
Our analysis builds upon simple structural conditions that boil down to randomized matrix multiplication, a fundamental and well-understood primitive of randomized linear algebra.
Our main algorithm (see Algorithm~\ref{algo:iterative_rda}) is analyzed in light of the following structural constraint, which constructs a sketching matrix $\Sb \in \mathbb{R}^{d \times s}$ (for an appropriate choice of the sketching dimension $s \ll d$), such that
\begin{flalign}
\norm{\Sigmab_{\lambda}\Vb^\ts\Sb\Sb^\ts\Vb\Sigmab_{\lambda} - \Sigmab_{\lambda}^2}_2 \leq \frac{\ve}{2} \,.
\label{eq:struct2}
\end{flalign}
Here, $\Vb\in\RR{d}{\rho}$ contains the right singular vectors of $\Ab$ and $\Sigmab_{\lambda}\in\RR{\rho}{\rho}$ is a diagonal matrix with
\begin{flalign}
(\Sigmab_{\lambda})_{ii}=\frac{\sigma_i}{\sqrt{\sigma_i^2+\lambda}},
\qquad i=1,\dots, \rho\,.\label{eq:sigmalambda}
\end{flalign}
Notice that $\norm{\Sigmab_{\lambda}}_F^2=d_\lambda$, which is defined to be the effective degrees of freedom of the RFDA problem (see eqn.~(\ref{eqn:df2})). Eqn.~(\ref{eq:struct2}) can be satisfied by sampling with respect to the \textit{ridge leverage scores} of~\cite{Alaoui2014,Co17} or by oblivious sketching matrix constructions (\eg, \emph{count-sketch} \cite{ClaWoo13} or \emph{sub-sampled randomized Hadamard transform}~(SRHT)~\cite{AilCha09,Drineas2011,Woodruff2014}) for $\Sb$ with column sizes $s$ depending on $d_\lambda$. Recall that $d_\lambda$ is upper bounded by $\rho$ but could be significantly smaller depending on the distribution of the singular values and the choice of $\lambda$. Indeed, it follows that by sampling-and-rescaling $\Ocal(d_\lambda\ln d_\lambda)$ predictor variables from the matrix $\Ab$ (using either exact or approximate ridge leverage scores~\cite{Alaoui2014,Co17}), we can satisfy the constraint of eqn.~(\ref{eq:struct2}) and Algorithm~\ref{algo:iterative_rda} returns an estimator\smash{ $\widehat{\Gb}$} satisfying
\begin{flalign}
\nbr{(\wb-\mb)^\ts(\widehat{\Gb}-\Gb)}_2\le\frac{\ve^{t}}{\sqrt{\lambda}}\nbr{\Vb\Vb^T(\wb-\mb)}_2 \,.\label{eq:bound_rfda}
\end{flalign}
Here $\wb\in\R{d}$ is any test data point and $\Vb\Vb^T(\wb-\mb)$ is the part of $\wb-\mb$ that lies within the range of $\Ab^\ts$ (see footnote 1 for the definition of $\mb$). We note that the dependency of the error on $\ve$ drops \emph{exponentially} fast as the number of iterations $t$ increases. See Section~\ref{sxn:sketch} for constructions of $\Sb$ and Section~\ref{sxn:prior} for a comparison of this bound with prior work.

Additionally, we complement the bound of eqn.~\eqref{eq:bound_rfda} with a second bound subject to a different structural condition, namely
\begin{flalign}
\nbr{\Vb^\ts\Sb\Sb^\ts\Vb-\Ib_\rho}_2\le\frac{\ve}{2}\,.\label{eq:struct1}
\end{flalign}
Indeed, assuming that the rank of $\Ab$ is much smaller than $\min\{n,d\}$, one can use the (exact or approximate) column \emph{leverage scores}~\citep{Mahoney2009,Mah-mat-rev_BOOK} of the matrix $\Ab$ to satisfy the aforementioned constraint by sampling $\Ocal(\rho\ln \rho)$ columns, in which case $\Sb$ is a sampling-and-rescaling matrix. Perhaps more interestingly, a variety of oblivious sketching matrix constructions for $\Sb$ can also be used to satisfy eqn.~(\ref{eq:struct1}) (see Section~\ref{sxn:sketch} for specific constructions of $\Sb$). In either case, under this structural condition, the output of Algorithm~\ref{algo:iterative_rda} satisfies
\begin{flalign}
\nbr{(\wb-\mb)^\ts(\widehat{\Gb}-\Gb)}_2\le\frac{\ve^{t}}{2\sqrt{\lambda}}\nbr{\Vb\Vb^T(\wb-\mb)}_2.\label{eq:rdares}
\end{flalign}
The above guarantee is essentially identical to the guarantee of eqn.~\eqref{eq:bound_rfda} and the approximation error decays exponentially fast as the number of iterations $t$ increases. However, this second bound exhibits a worse dependency on the sketching size $s$. Indeed, eqn.~\eqref{eq:struct1} can be satisfied by sampling-and-rescaling $\Ocal(\rho\ln \rho)$ predictor variables from the matrix $\Ab$, which could be much larger than the sketch size needed when sampling with respect to the ridge leverage scores.

To the best of our knowledge, our bounds are the first attempt to provide general structural results that guarantee provable, high-quality solutions  for the RFDA problem. To summarize,
our first structural result (Theorem~\ref{thm:iterative2}) can be satisfied by sampling with respect ro ridge leverage scores or by the use of oblivious sketching matrices whose size depends on the effective degrees of freedom of the RFDA problem and results in a highly accurate guarantee in terms of ``distance  distortion'' caused by iterative sketching. While ridge leverage scores have been used in a number of applications involving matrix approximation, cost-preserving projections, clustering, etc.~\cite{Co17}, their performance in the context of RFDA has not been analyzed in prior work.  Our second structural result (Theorem~\ref{thm:iterative_rda}) complements the analysis of Theorem~\ref{thm:iterative2} subject to a second structural condition (eqn.~\eqref{eq:struct1}) which can be satisfied by sampling with respect to standard leverage scores using a sketch size that depends on the rank of the centered data matrix.

%!TEX root = arxiv_LDA.tex

\subsection{Prior Work}\label{sxn:prior}

The work most closely related to ours is~\cite{LCZ17}, where the authors proposed a fast random-projection-based algorithm to accelerate RFDA. Their theoretical analysis showed that random projections (and in particular the count-min sketch) preserve the generalization ability of FDA on the original training data.  However, for the $d\gg n$ case, the error bound in their work  (Theorem~3 of~\cite{LCZ17}) depends on the condition number of the centered data matrix $\Ab$. More precisely, they proved that their method computes a matrix \smash{$\widehat{\Gb}$} in time $\Ocal(\nnz{\Ab})+\Ocal(n^2q+n^3+ndc)$,  which, for any test data point $\wb\in\R{d}$, satisfies
\begin{flalign*}
\|(\wb-\mb)^\ts(\widehat{\Gb}-\Gb)\|_2\le\frac{\kappa\,\ve}{1-\ve}\|\Vb\Vb^\ts(\wb-\mb)\|_2\,,
\end{flalign*}
with high probability for any $\epsilon \in (0,1]$. Here $\kappa$ is the condition number of $\Ab$; thus, their random-projection-based RFDA approach well-approximates the original RFDA problem only when $\Ab$ is well-conditioned ($\kappa$ small).

Our work was heavily motivated by~\cite{ZDXJ10}, where the authors presented a flexible and efficient implementation of RFDA through an EVD-based algorithm. In addition, \cite{ZDXJ10} uncovered a general relationship between RFDA and ridge regression that explains how matrix $\Gb$ has similar properties with the solution matrix $\Wb$ in terms of distance-based classification methods. We also note that using their linear algebraic formulation and the proposed EVD-based framework,~\cite{LCZ17} presented a fast implementation of FDA. Another line of work that motivated our approach was the framework of leverage score sampling and the relatively recent introduction of ridge leverage scores~\cite{Alaoui2014,Co17}. Indeed, our Theorems~\ref{thm:iterative2} and \ref{thm:iterative_rda} present  structural results that can be satisfied (with high probability) by sampling columns of $\Ab$ with probabilities proportional to (exact or approximate) ridge leverage scores and leverage scores, respectively (see Section~\ref{sxn:sketch}). To the best of our knowledge, ours are the first results showing a strong accuracy guarantee for RFDA problems when ridge leverage scores are used to sample predictor variables, in one or more iterations.

% In this context, there is another recent paper \cite{CYD18} where the authors presented an iterative ridge regression algorithm in a sketching-based framework for $d\gg n$. They proved that the output of their proposed algorithm closely approximates the true solution of ridge regression problem if the columns of the data-matrix are sampled with probability proportional to column ridge leverage scores and the number of samples depends on the \textit{effective degrees of freedom} of the problem. However, a key advantage of our work over the aforementioned paper \cite{CYD18} is that their main result (Theorem~2 in \cite{CYD18}) holds under the assumption that $\lambda$ lies on the spectrum of the non-zero singular values of $\Ab$, whereas our main result (Theorem~1) does not rely on such a condition and thus is valid for any $\lambda>0$. Furthermore, in terms of the linear algebraic manipulations as well as the complexity of the problem in general, we emphasize that the transition from regularized regression problems to RFDA is far from being trivial.

Under a different context, in a recent paper~\cite{CYD18}, we presented an iterative algorithm for ridge regression problems with $d\gg n$ in a sketching-based framework. There, we proved that the output of our proposed algorithm closely approximates the true solution of the ridge regression problem if the columns of the data matrix are sampled with probabilities proportional to the column ridge leverage scores. The number of samples required depends on the \textit{effective degrees of freedom} of the problem. However, a key advantage of our current work is that the main result (Theorem~2) in~\cite{CYD18} holds under assumptions on $\lambda$ and the singular values of $\Ab$, whereas our main result here (Theorem~\ref{thm:iterative2}) is valid for any $\lambda>0$. Furthermore, the transition from regularized regression problems to RFDA is far from trivial.

Among other relevant works, \cite{BZSH14} addressed the scalability of FDA by developing a random projection-based FDA algorithm and presented a theoretical analysis of the approximation error involved. However, their framework applies exclusively to the \textit{two-stage} FDA problem \cite{BHK97,YL05}, where the issue of singularity is addressed before the actual FDA stage.
Another line of research \cite{SP12,WT15} dealt with the fast implementation of null-space based FDA \cite{Chen00} for $d\gg n$ using random matrices.
Nevertheless, their approach is quite different from ours and does not come with provable guarantees.
Finally,~\cite{YJ05} proposed an iterative approach to address the singularity of $\Ab^\ts\Ab$, where the underlying data representation model is different from conventional FDA.
Although, the running time of their proposed algorithm is empirically lower than the original approach and its classification accuracy is  competitive, it does not yield a closed form solution of the discriminant directions.
%!TEX root = arxiv_LDA.tex

\subsection{Notations}
We use $\ab,\bb,\dots$ to denote vectors and $\Ab,\Bb,\dots$ to denote matrices. For a matrix $\Ab$, $\Ab_{*i}$ ($\Ab_{i*}$) denotes the $i$-th column (row) of $\Ab$ as a column (row) vector. For a vector $\ab$, $\nbr{\ab}_2$ denotes its Euclidean norm; for a matrix $\Ab$, $\nbr{\Ab}_2$ denotes its spectral norm and $\nbr{\Ab}_F$ denotes its Frobenius norm. We refer the reader to~\cite{Golub1996} for properties of norms that will be quite useful in our work. For a matrix $\Ab \in \mathbb{R}^{n \times d}$ with $d\geq n$ of rank $\rho$, its (thin) Singular Value Decomposition (SVD) is the product $\Ub \Sigmab \Vb^\ts$, with $\Ub \in \mathbb{R}^{n \times \rho}$ (the matrix of the left singular vectors), $\Vb \in \mathbb{R}^{d \times \rho}$ (the matrix of the right singular vectors), and $\Sigmab \in \mathbb{R}^{\rho \times \rho}$ a diagonal matrix whose diagonal entries are the non-zero singular values of $\Ab$ arranged in a non-increasing order. Computation of the SVD takes, in this setting, $\Ocal(n^2d)$ time. 
%We will use the notation $\Ub_k \in \mathbb{R}^{n \times k}$ to denote the matrix of the top $k$ left singular vectors and $\Ub_{k,\perp} \in \mathbb{R}^{n \times (n-k)}$ to denote the matrix of the bottom $n-k$ left singular vectors. 
We will often use $\sigma_i$ to denote the singular values of a matrix implied by context. Additional notation will be introduced as needed.

\section{Iterative Approach}
%\subsection{Regularized Discriminant Analysis}

%\begin{minipage}{.47\linewidth}
%
Our main algorithm (Algorithm~\ref{algo:iterative_rda}) solves a \textit{sketched} RFDA problem in each iteration while updating the (rescaled) class membership matrix to account for the information already captured in prior iterations.
More precisely, our algorithm iteratively computes  a sequence of matrices $\widetilde{\Gb}^{(j)}\in\RR{d}{c}$ for $j=1,\dots, t$ and returns the estimator $\widehat{\Gb}=\sum_{j=1}^{t}\widetilde{\Gb}^{(j)}$ to the original matrix $\Gb$ of eqn.~\eqref{eq:rfda3}.
Our main quality-of-approximation results (Theorems~\ref{thm:iterative2} and~\ref{thm:iterative_rda}) argue that returning the \textit{sum} of those intermediate matrices results in highly accurate approximations when compared to the original approach.
%
%\end{minipage}
%\hspace{.01\linewidth}
%
%\begin{minipage}{.5\linewidth}
%
\begin{algorithm}[H]
	\caption{Iterative RFDA Sketch}\label{algo:iterative_rda}
	\begin{algorithmic}
		\State \textbf{Input:}
		$\Ab\in\RR{n}{d}$,  $\Omegab \in \RR{n}{c}$, $\lambda>0$;
		number of iterations $t>0$;
		sketching matrix $\Sb \in \mathbb{R}^{d\times s}$;
		\State \textbf{Initialize:}
		$\Lb^{(0)} \gets \Omegab $,
		$\widetilde{\Gb}^{(0)} \gets \zero_{d\times c}$, \\
		\qquad \qquad \ \  $\Yb^{(0)} \gets \zero_{n\times c}$;

		\For{$j=1$ \textbf{to} $t$}
		\State $\Lb^{(j)} \gets \Lb^{(j-1)}-\lambda\Yb^{(j-1)}-\Ab\widetilde{\Gb}^{(j-1)}$;
		\State $\Yb^{(j)} \gets (\Ab \Sb\Sb^\ts \Ab^\ts+\lambda\Ib_n)^{-1}\Lb^{(j)}$;
		\State $\widetilde{\Gb}^{(j)} \gets \Ab^\ts \Yb^{(j)}$;
		\EndFor

		\State \textbf{Output:} $\widehat{\Gb} = \sum_{j=1}^{t} \widetilde{\Gb}^{(j)}$;
	\end{algorithmic}
\end{algorithm}
%
%\end{minipage}

Theorem~\ref{thm:iterative2} presents our approximation guarantees under the assumption that the sketching matrix $\Sb$ satisfies the constraint of eqn.~(\ref{eq:struct2}).

\begin{theorem}\label{thm:iterative2}
Let $\Ab \in \mathbb{R}^{n \times d}$ and $\Gb\in\RR{d}{c}$ be as defined in Section~\ref{sxn:intro}. Assume that for some constant $0<\ve<1$ the sketching matrix $\Sb \in \mathbb{R}^{d\times s}$ satisfies eqn.~\eqref{eq:struct2}. Then, for any test data point $\wb\in\R{d}$, the estimator \smash{$\widehat{\Gb}$} returned by Algorithm~\ref{algo:iterative_rda} satisfies
\begin{flalign*}
\nbr{(\wb-\mb)^\ts(\widehat{\Gb}-\Gb)}_2\le\frac{\ve^{t}}{\sqrt{\lambda}}\, \nbr{\Vb\Vb^\ts(\wb-\mb)}_2\,.
\end{flalign*}
Recall that $\Vb\Vb^\ts(\wb-\mb)$ is the projection of the vector $\wb-\mb$ onto the row space of $\Ab$.
\end{theorem}

Similarly, Theorem~\ref{thm:iterative_rda} presents our accuracy guarantees under the assumption that the sketching matrix $\Sb$ satisfies the constraint of eqn.~(\ref{eq:struct1}).
\begin{theorem}\label{thm:iterative_rda}
Let $\Ab \in \mathbb{R}^{n \times d}$ and $\Gb\in\RR{d}{c}$ be as defined in Section~\ref{sxn:intro}. Assume that for some constant $0<\ve<1$ the sketching matrix $\Sb \in \mathbb{R}^{d\times s}$ satisfies eqn.~\eqref{eq:struct1}. Then, for any test data point $\wb\in\R{d}$, the estimator \smash{$\widehat{\Gb}$} returned by Algorithm~\ref{algo:iterative_rda} satisfies
\begin{flalign*}
\nbr{(\wb-\mb)^\ts(\widehat{\Gb}-\Gb)}_2\le\frac{\ve^{t}}{2\sqrt{\lambda}}\, \nbr{\Vb\Vb^\ts(\wb-\mb)}_2 \,.
\end{flalign*}
Recall that $\Vb\Vb^\ts(\wb-\mb)$ is the projection of the vector $\wb-\mb$ onto the row space of $\Ab$.
\end{theorem}

\emph{Running time of Algorithm~\ref{algo:iterative_rda}.}
First, we need to compute $\Ab\widetilde{\Gb}^{(j-1)}$ which takes time $\Ocal(c\cdot\nnz{\Ab})$. Then, computing the sketch $\Ab\Sb \in \mathbb{R}^{n \times s}$ takes $T(\Ab,\Sb)$ time which depends on the particular construction of $\Sb$ (see Section~\ref{sxn:sketch}). In order to invert the matrix $\Thetab = \Ab\Sb\Sb^\ts\Ab^\ts+\lambda\Ib_n$, it suffices to compute the SVD of the matrix $\Ab\Sb$. Notice that given the singular values of $\Ab\Sb$ we can compute the singular values of $\Thetab$ and also notice that the left and right singular vectors of $\Thetab$ are the same as the left singular vectors of $\Ab\Sb$. Interestingly, we do not need to compute $\Thetab^{-1}$: we can store it implicitly by storing its left (and right) singular vectors $\Ub_{\Thetab}$ and its singular values $\Sigmab_{\Thetab}$. Then, we can compute all necessary matrix-vector products using this implicit representation of $\Thetab^{-1}$. Thus, inverting $\Thetab$ takes $\Ocal(sn^2)$ time. Updating the matrices \smash{$\Lb^{(j)}$}, \smash{$\Yb^{(j)}$}, and \smash{$\widetilde{\Gb}^{(j)}$} is dominated by the aforementioned running times. Thus, summing over all $t$ iterations, the running time of Algorithm~\ref{algo:iterative_rda} is
\begin{flalign}
\label{eqn:runtime} \Ocal(t\,c\cdot\nnz{\Ab})+ \Ocal(s\,n^2) + T(\Ab,\Sb),
\end{flalign}
which should be compared to the $\Ocal(n^2d)$ time that would be needed by standard RFDA approaches.

We note that our algorithm can also be viewed as a \emph{preconditioned Richardson iteration} with step-size equal to one for solving the linear system $(\Ab\Ab^\ts+\lambda\Ib_n)\Fb=\Omegab$ in $\Fb\in\RR{n}{c}$ with randomized pre-conditioner $\Pb^{-1}=(\Ab\Sb\Sb^\ts\Ab^\ts+\lambda\Ib_n)^{-1}$.
However, our objective and analysis are significantly different compared to the conventional preconditioned Richardson iteration. \textit{First}, our matrix of interest is $\Gb=\Ab^\ts\Fb\in\RR{d}{c}$, whereas standard convergence analysis of preconditioned Richardson's method is with respect to $\Fb$.
Specifically, in the context of discriminant analysis, for a new observation $\wb\in\R{d}$, we are interested in understanding whether the output of our algorithm closely approximates the original point in the projected space, \ie, if $\|(\wb-\mb)^\ts(\widehat{\Gb}-\Gb)\|_2$ is sufficiently small. To the best of our knowledge, standard analysis of preconditioned Richardson iteration does not yield a bound for $\|(\wb-\mb)^\ts(\widehat{\Gb}-\Gb)\|_2$. \textit{Second}, our analysis is with respect to the Euclidean norm whereas the standard convergence analysis of preconditioned Richardson iteration is in terms of the energy-norm of $(\Ab\Ab^\ts+\lambda\Ib_n)$, as the matrix $\Pb^{-1}(\Ab\Ab^\ts+\lambda\Ib_n)$ is not symmetric positive definite.

We conclude the section by noting that our proof would also work when \textit{different} sampling matrices $\Sb_j$ (for $j=1,\dots, t$) are used in each iteration, as long as they satisfy the constraints of eqns.~(\ref{eq:struct2}) or~(\ref{eq:struct1}). As a matter of fact, the sketching matrices $\Sb_j$ do not even need to have the same number of columns. See Section~\ref{sxn:conclusions} for an interesting open problem in this setting.

\paragraph{Satisfying structural conditions~\eqref{eq:struct2}~and~\eqref{eq:struct1}.}
%
%\subsection{Sketching-based approaches to construct $\Sb$}
\label{sxn:sketch}
The conditions of eqns.~\eqref{eq:struct2}~and~\eqref{eq:struct1} essentially boil down to randomized, approximate matrix multiplication~\cite{Dr01,Dr06}, a task that has received much attention in the randomized linear algebra community.
We discuss general \textit{sketching-based} approaches here and
defer the discussion of \emph{sampling-based} approaches and the corresponding results to Appendix~\ref{sxn:appendix:sampling}.
% delegate sampling-based approaches to Appendix~\ref{sxn:appendix:sampling}:
% discuss sampling-based approaches in the next section;
A particularly useful result for our purposes appeared in~\citet{Cohen2016}. Under our notation,~\cite{Cohen2016} proved that for $\Zb \in \mathbb{R}^{d \times n}$ and for a (suitably constructed) sketching matrix $\Sb \in \mathbb{R}^{d \times s}$, with probability at least $1-\delta$,
\begin{flalign}
\nbr{\Zb^\ts \Sb \Sb^\ts \Zb - \Zb^\ts \Zb}_2\le \ve \left(\nbr{\Zb}_2^2+\frac{\nbr{\Zb}_F^2}{r}\right).\label{eqn:mm}
\end{flalign}
The above bound holds for a very broad family of constructions for the sketching matrix $\Sb$
%: picking $\Sb$ from any oblivious subspace embedding (OSE) distribution (with parameters $\ve$, $\delta$, and $2r$) over matrices in %$\mathbb{R}^{d \times s}$ suffices
%
(see~\cite{Cohen2016} for details). In particular,~\cite{Cohen2016} demonstrated a construction for $\Sb$ with $s=\Ocal(r/\ve^2)$ columns such that, for any $n\times d$ matrix $\Ab$, the product $\Ab\Sb$ can be computed in time $\Ocal(\nnz{\Ab})+\tilde{\Ocal}((r^3+r^2n)/\ve^{\gamma})$ for some constant $\gamma$. Thus, starting with eqn.~(\ref{eq:struct1}) and using this particular construction for $\Sb$, let $\Zb=\Vb\Sigmab_{\lambda}$ and note that $\nbr{\Vb\Sigmab_{\lambda}}_F^2=d_\lambda$ and $\nbr{\Vb\Sigmab_{\lambda}}_2\le1$. Setting $r=d_\lambda$, eqn.~(\ref{eqn:mm}) implies that
$$\nbr{\Sigmab_{\lambda}\Vb^\ts \Sb \Sb^\ts \Vb\Sigmab_{\lambda} - \Sigmab_{\lambda}^2}_2\le 2\,\ve.$$
In this case, the running time needed to compute the sketch equals \smash{$T(\Ab,\Sb) = \Ocal(\nnz{\Ab})+\widetilde{\Ocal}(d_\lambda^2 n/\ve^{\gamma}).$}
The running time of the overall algorithm follows from eqn.~(\ref{eqn:runtime}) and our choices for $s$ and $r$:
$$\Ocal(t\,c\cdot \nnz{\Ab})+\widetilde{\Ocal}(d_\lambda n^2/\ve^{\max\{2,\gamma\}}).$$
The failure probability (hidden in the polylogarithmic terms) can be easily controlled using a union bound. Finally, a simple change of variables (using $\ve/4$ instead of $\ve$) suffices to satisfy the structural condition of eqn.~(\ref{eq:struct2}) without changing the above running time.

Similarly, starting with eqn.~(\ref{eq:struct1}), let $\Zb=\Vb$ and note that $\nbr{\Vb}_F^2=\rho$ and $\nbr{\Vb}_2=1$. Setting $r=\rho$, eqn.~(\ref{eqn:mm}) implies that
$\nbr{\Vb^\ts \Sb \Sb^\ts \Vb - \Ib_{\rho}}_2\le 2\ve.$
In this case, the running time of the sketch computation is equal to $T(\Ab,\Sb) = \Ocal(\nnz{\Ab})+\widetilde{\Ocal}(\rho^2n/\ve^{\gamma}).$
The running time of the overall algorithm follows from eqn.~(\ref{eqn:runtime}) and our choices for $s$ and $r$:
$$\Ocal(t\,c\cdot \nnz{\Ab})+\widetilde{\Ocal}(\rho n^2/\ve^{\max\{2,\gamma\}}).$$
Again, a simple change of variables suffices to satisfy eqn.~(\ref{eq:struct1}) without changing the running time.

We note that the above running times can be slightly improved if $s$ is smaller than $n$, since $s$ depends only on the \emph{effective degrees of freedom $(d_\lambda)$} of the problem (or, on the rank $\rho$ of the data matrix $\Ab$).
In this case, the SVD of $\Ab\Sb$ can be computed in $\Ocal(n s^2)$ time, and the running time of our algorithm is given by $\Ocal(t\,c\cdot\nnz{\Ab})+\widetilde{\Ocal}(d_\lambda^2 n/\ve^{\max\{4,\gamma\}})$ (or, $\Ocal(t\,c\cdot\nnz{\Ab})+\widetilde{\Ocal}(\rho^2 n/\ve^{\max\{4,\gamma\}})$).
%!TEX root = arxiv_LDA.tex

\section{Sketching the Proof of Theorem~\ref{thm:iterative2}}
%\textcolor{red}{PD: Include 1.5 page outline of the proof of the above theorem --- include as much of the proof as you can. I did not edit the experiments sections, but it should not be longer than one page, it just support the existing theory.}
Due to space considerations, most of our proofs have been delegated to the Appendix. However, to provide a flavor of the mathematical derivations underlying our contributions, we will present an outline of the proof of Theorem~\ref{thm:iterative2}.

Using the quantities defined in Algorithm~\ref{algo:iterative_rda}, let
\begin{flalign}\label{eqn:pdits}
\Gb^{(j)}=\Ab^\ts(\Ab\Ab^\ts+\lambda\Ib_n)^{-1}\Lb^{(j)}~~~\text{for }j=1,\dots, t.
\end{flalign}
Note that $\Gb=\Gb^{(1)}$. We remind the reader that $\Ub \in \mathbb{R}^{n \times \rho}$, $\Vb \in \mathbb{R}^{d \times \rho}$ and $\Sigmab \in \mathbb{R}^{\rho \times \rho}$ are, respectively, the matrices of the left singular vectors, right singular vectors  and singular values of $\Ab$. We will make extensive use of the matrix $\Sigmab_{\lambda}$ defined in eqn.~\eqref{eq:sigmalambda}. The following result provides an alternative expression for $\Gb^{(j)}$ which is easier to work with (see Appendix~\ref{app:ridlev} for the proof).
\begin{lemma}\label{lem:alt_xopt}
	For $j=1,\dots, t$, let $\Lb^{(j)}$ be the intermediate matrices in Algorithm~\ref{algo:iterative_rda} and $\Gb^{(j)}$ be the matrix defined in eqn.~\eqref{eqn:pdits}.
	Then for any $j=1,\dots, t$, $\Gb^{(j)}$ can also be expressed as
	\begin{flalign}
	\Gb^{(j)}=\Vb\Sigmab_\lambda^2\Sigmab^{-1}\Ub^\ts\Lb^{(j)}.
	%\label{eq:fda12}\nonumber
	\end{flalign}
\end{lemma}
Our next result (see Appendix~\ref{app:ridlev} for a detailed proof) provides a bound which later on plays a very important role in showing that the underlying error decays exponentially as the number of iterations in Algorithm~\ref{algo:iterative_rda} increases. We state the lemma and briefly outline its proof.
\begin{lemma}\label{lem:struct3}
	%%%%%%%%%%%%%%%%%%%%%%%%%%%%%%%%%%%%%%%%%%%%%%%%%%%%%%%%%%%%%%%%%%%%%%%%%%%%%%
	For $j=1,\dots, t$, let $\Lb^{(j)}$ be as defined in Algorithm~\ref{algo:iterative_rda} and let $\widetilde{\Gb}^{(j)}$
be defined as in eqn.~\eqref{eqn:pdits}. Further, let $\Sb\in\RR{d}{s}$ be the sketching matrix and let $\Eb=\Sigmab_\lambda\Vb^\ts\Sb\Sb^\ts\Vb\Sigmab_\lambda-\Sigmab_\lambda^2$. If eqn.~\eqref{eq:struct2} is satisfied, i.e., $\|\Eb\|_2\le\frac{\ve}{2}$, then, for all $j=1,\dots, t$,
	\begin{flalign} \nbr{(\wb-\mb)^\ts(\widetilde{\Gb}^{(j)}-\Gb^{(j)})}_2\le\ve\,\nbr{\Vb\Vb^\ts(\wb-\mb)}_2\,\nbr{\Sigmab_\lambda\Sigmab^{-1}\Ub^\ts\Lb^{(j)}}_2 \,.
	\end{flalign}
\end{lemma}
\begin{proof}[Proof sketch]
Applying Lemma~\ref{lem:alt_xopt} and using the SVD of $\Ab$ and the fact that \smash{$\|\Eb\|_2<1$}, we first express the intermediate matrices $\widetilde{\Gb}^{(j)}$ of Algorithm~\ref{algo:iterative_rda} in terms of the matrices $\Gb^{(j)}$ of eqn.~\eqref{eqn:pdits} as
\begin{flalign}
\smash{\widetilde{\Gb}^{(j)}=\Gb^{(j)}+\Vb\Sigmab_\lambda\Qb\Sigmab_\lambda\Sigmab^{-1}\Ub^\ts\Lb^{(j)}}\label{eq:th1_g_j},
\end{flalign}
where \smash{$\Qb=\sum_{\ell=1}^\infty(-1)^\ell\Eb^\ell$}.
Notice that
\begin{flalign}
\nbr{\Qb}_2
& = \Big\|\sum_{\ell=1}^\infty(-1)^\ell\Eb^\ell\Big\|_2
\le\sum_{\ell=1}^\infty\nbr{\Eb^\ell}_2
\le\sum_{\ell=1}^\infty\nbr{\Eb}_2^\ell
\le\sum_{\ell=1}^\infty\left(\frac{\ve}{2}\right)^\ell
= \frac{\ve/2}{1-\ve/2}
% \le\frac{\frac{\ve}{4\sqrt{2}}}{1/2}
\le \ve \,.\label{eq:q_j}
\end{flalign}
In the above, we used the triangle inequality, submultiplicativity of the spectral norm, and the fact that $\ve \le 1$.
%q
Next, we plug-in eqn.~\eqref{eq:th1_g_j} and apply submultiplicativity to conclude
\begin{align*}
 \nbr{(\wb-\mb)^\ts(\widetilde{\Gb}^{(j)}-\Gb^{(j)})}_2
=\ & \nbr{(\wb-\mb)^\ts\Vb\Sigmab_\lambda\Qb\Sigmab_\lambda\Sigmab^{-1}\Ub^\ts\Lb^{(j)}}_2 \\
\le\ & \nbr{(\wb-\mb)^\ts\Vb}_2 \, \nbr{\Sigmab_\lambda}_2 \, \nbr{\Qb}_2 \, \nbr{\Sigmab_\lambda\Sigmab^{-1}\Ub^\ts\Lb^{(j)}}_2\\
% \le \ve~\nbr{(\wb-\mb)^\ts\Vb}_2\nbr{\Sigmab_\lambda\Sigmab^{-1}\Ub^\ts\Lb^{(j)}}_2 \\
\le\ & \ve\,\nbr{\Vb\Vb^\ts(\wb-\mb)}_2 \, \nbr{\Sigmab_\lambda\Sigmab^{-1}\Ub^\ts\Lb^{(j)}}_2,
%=&~\ve~\|\alphab\|_2\nbr{\Sigmab_\lambda\Sigmab^{-1}\Ub^\ts\Lb^{(j)}}_2
%=&~2\ve~\nbr{(\wb-\mb)^\ts\Vb}_2=~2\ve~\nbr{\alphab}_2
\end{align*}
where the last inequality follows from eqn.~\eqref{eq:q_j} and the fact that $\nbr{\Sigmab_\lambda}_2\le1$.
\end{proof}
The next lemma (see Appendix~\ref{app:ridlev} for its proof) presents a structural result for the matrix $\Gb$.
\begin{lemma}\label{lem:induction}
	Let $\widetilde{\Gb}^{(j)}$, $j=1,\dots, t$ be the sequence of matrices introduced in Algorithm~\ref{algo:iterative_rda} and let $\Gb^{(t)}\in\R{d}$ be defined as in eqn.~\eqref{eqn:pdits}. Then,
	the matrix $\Gb$ in eqn.~\eqref{eq:rfda3} can be expressed as
	\begin{flalign}
	\Gb=\Gb^{(t)}+\sum_{j=1}^{t-1}\widetilde{\Gb}^{(j)}.
	\label{eq:iterative2_1}
	\end{flalign}
\end{lemma}
Repeated application of Lemmas~\ref{lem:induction} and~\ref{lem:struct3} yields:
	\begin{align}
	\nonumber \nbr{(\wb-\mb)^\ts(\widehat{\Gb} -\Gb)}_2
	= \ & \|(\wb-\mb)^\ts(\textstyle\sum_{j=1}^{t}\widetilde{\Gb}^{(j)}-\Gb)\|_2\\
	= \ & \|(\wb-\mb)^\ts(\widetilde{\Gb}^{(t)}-(\Gb-\textstyle\sum_{j=1}^{t-1}\widetilde{\Gb}^{(j)}))\|_2\nonumber\\
	\le\ & \nbr{(\wb-\mb)^\ts(\widetilde{\Gb}^{(t)}-\Gb^{(t)})}_2\nonumber\\
	\le\ & \ve\,\nbr{\Vb\Vb^\ts(\wb-\mb)}_2\nbr{\Sigmab_\lambda\Sigmab^{-1}\Ub^\ts\Lb^{(t)}}_2.\label{eq:fda_induction}
	\end{align}
The next bound (see Appendix~\ref{app:ridlev} for its detailed proof) provides a critical inequality that can be used recursively in order to establish Theorem~\ref{thm:iterative2}.
\begin{lemma}\label{lem:fda_recur}
	Let $\Lb^{(j)}$, $j=1,\dots, t$ be the matrices defined in Algorithm~\ref{algo:iterative_rda}. For any $j=1,\dots, t-1$, if eqn.~\eqref{eq:struct2} is satisfied, i.e., $\|\Eb\|_2\le\frac{\ve}{2}$, then
	\begin{flalign}
	\nbr{\Sigmab_\lambda\Sigmab^{-1}\Ub^\ts\Lb^{(j+1)}}_2\le
	\ve\, \nbr{\Sigmab_\lambda\Sigmab^{-1}\Ub^\ts\Lb^{(j)}}_2.\label{eq:fda_recur}
	\end{flalign}
\end{lemma}
%\textcolor{red}{PD: Include a sketch of the proof of the above lemma to reach the page limit.}
\begin{proof}[Proof sketch]
From Algorithm~\ref{algo:iterative_rda}, we have that for $j=1,\dots, t-1$,
\begin{flalign}
\Lb^{(j+1)}=\Lb^{(j)}-\lambda\Yb^{(j)}-\Ab\widetilde{\Gb}^{(j)}
=\Lb^{(j)}-(\Ab\Ab^\ts+\lambda\Ib_n)(\Ab\Sb\Sb^\ts\Ab^\ts+\lambda\Ib_n)^{-1}\Lb^{(j)}.\label{eq:fda_recur8}
\end{flalign}

Applying the SVD of $\Ab$
% along with further linear algebraic manipulations yield the following
it can be shown (see Appendix~\ref{app:ridlev} for details) that
\begin{flalign}
&(\Ab\Ab^\ts+\lambda\Ib_n)(\Ab\Sb\Sb^\ts\Ab^\ts+\lambda\Ib_n)^{-1}\Lb^{(j)}\nonumber\\ &~~~~~~~~~~~~~~~~~~~~~~~=\Lb^{(j)}+\Ub(\Sigmab^2+\lambda\Ib_\rho)\Sigmab^{-1}\Sigmab_\lambda\Qb\Sigmab_\lambda\Sigmab^{-1}\Ub^\ts\Lb^{(j)},\label{eq:fda_recur9}
\end{flalign}
where $\Qb=\sum_{\ell=1}^\infty(-1)^\ell\Eb^\ell$.

Combining eqns.~\eqref{eq:fda_recur8} and \eqref{eq:fda_recur9}, we get
\begin{flalign}
\Lb^{(j+1)}=-\Ub(\Sigmab^2+\lambda\Ib_\rho)\Sigmab^{-1}\Sigmab_\lambda\Qb\Sigmab_\lambda\Sigmab^{-1}\Ub^\ts\Lb^{(j)}. \label{eq:fda_recur11}
\end{flalign}
Finally, applying eqn.~\eqref{eq:fda_recur11}, we obtain
\begin{flalign}
\nbr{\Sigmab_\lambda\Sigmab^{-1}\Ub^\ts\Lb^{(j+1)}}_2=&~\nbr{\Sigmab_\lambda\Sigmab^{-1}\Ub^\ts\Ub(\Sigmab^2+\lambda\Ib_\rho)\Sigmab^{-1}\Sigmab_\lambda\Qb\Sigmab_\lambda\Sigmab^{-1}\Ub^\ts\Lb^{(j)}}_2\nonumber\\
=&~\nbr{\Sigmab_\lambda\Sigmab^{-1}(\Sigmab^2+\lambda\Ib_\rho)\Sigmab^{-1}\Sigmab_\lambda\Qb\Sigmab_\lambda\Sigmab^{-1}\Ub^\ts\Lb^{(j)}}_2\nonumber\\
=&~\nbr{\Qb\Sigmab_\lambda\Sigmab^{-1}\Ub^\ts\Lb^{(j)}}_2
\le\nbr{\Qb}_2\nbr{\Sigmab_\lambda\Sigmab^{-1}\Ub^\ts\Lb^{(j)}}_2\nonumber\\
\le&~\ve\,\nbr{\Sigmab_\lambda\Sigmab^{-1}\Ub^\ts\Lb^{(j)}}_2\nonumber
\end{flalign}
where the third equality holds since $\Sigmab_\lambda\Sigmab^{-1}(\Sigmab^2+\lambda\Ib_\rho)\Sigmab^{-1}\Sigmab_\lambda=\Ib_\rho$. The last two inequalities follow from sub-multiplicativity and the fact that $\|\Qb\|_2\le\ve$ (from eqn.~\eqref{eq:q_j}).
\end{proof}

\paragraph{Proof of Theorem~\ref{thm:iterative2}.} Applying Lemma~\ref{lem:fda_recur} iteratively, we get
\begin{flalign}
\nbr{\Sigmab_\lambda\Sigmab^{-1}\Ub^\ts\Lb^{(t)}}_2\le \ve\,\nbr{\Sigmab_\lambda\Sigmab^{-1}\Ub^\ts\Lb^{(t-1)}}_2\le
\ldots
\le\ve^{t-1}\nbr{\Sigmab_\lambda\Sigmab^{-1}\Ub^\ts\Lb^{(1)}}_2.
\label{eq:fda_rec}
\end{flalign}
Notice that $\Lb^{(1)} = \Omegab$ by definition.
% From the definition of $\Omegab$, it is easy to verify that
Also, $\Omegab^\ts\Omegab=\Ib_c$ and thus $\nbr{\Omegab}_2=1$.
Furthermore, we know that $\nbr{\Ub^\ts}_2=1$ and $\nbr{\Sigmab_\lambda\Sigmab^{-1}}_2=\displaystyle\max_{1\le i\le\rho}(\sigma_i^2+\lambda)^{-\frac{1}{2}}$. Thus, sub-multiplicativity yields
\begin{flalign}
\nbr{\Sigmab_\lambda\Sigmab^{-1}\Ub^\ts\Lb^{(1)}}_2
% \nbr{\Sigmab_\lambda\Sigmab^{-1}\Ub^\ts\Omegab}_2
\le \nbr{\Sigmab_\lambda\Sigmab^{-1}}_2\nbr{\Ub^\ts}_2\nbr{\Omegab}_2
= \max_{1\le i\le\rho}(\sigma_i^2+\lambda)^{-\frac{1}{2}}\le \lambda^{-\frac{1}{2}},
\label{eq:proof_th1}
\end{flalign}
where the last inequality holds since
$
(\sigma_i^2+\lambda)^{-\frac{1}{2}} \le \lambda^{-\frac{1}{2}}
$
for all $i=1\ldots \rho$.

Finally, combining eqns.~\eqref{eq:fda_induction},~\eqref{eq:fda_rec} and~\eqref{eq:proof_th1}, we get
\begin{flalign*}
\nbr{(\wb-\mb)^\ts(\widehat{\Gb} -\Gb)}_2\le\frac{\ve^t}{\sqrt{\lambda}}~\nbr{\Vb\Vb^\ts(\wb-\mb)}_2\,,
\end{flalign*}
which concludes the proof.
\qed

%!TEX root = arxiv_LDA.tex

\section{Empirical Evaluation}

\subsection{Experiment Setup}

We perform experiments on two real-world datasets:
ORL~\cite{ATT94} is a database of grey-scale face images with $n = 400$ examples and $d = 10,304$ features, with each example belonging to one of $c = 40$ classes;
% RNA~\cite{UCI13} is a gene expression RNA-Seq dataset, and
PEMS~\cite{UCI11} describes the occupancy rate of different car lanes in freeways of the San Francisco bay area, with $n = 440$ examples, $d = 138,672$ features, and $c = 7$ label classes.
% Table~\ref{tab:dataset} provides the summary statistics of the datasets used.

% \begin{table}[htbp]
% \caption{Dataset summary.}
% \label{tab:dataset}
% \begin{center}
% % \begin{small}
% \begin{tabular}{lcccr}
% \toprule
% Dataset & \#Examples $n$ & \#Features $d$ & \#Classes $c$ \\
% \midrule
% ORL~\cite{ATT94} & 400 & 10,304 & 40 \\
% RNA~\cite{UCI13} & 801 & 20,264 & 5 \\
% PEMS~\cite{UCI11} & 440 & 138,672 & 7 \\
% \bottomrule
% \end{tabular}
% % \end{small}
% \end{center}
% \vskip -0.1in
% \end{table}

In our experiments, we compare both sketching-based and sampling-based constructions for the sketching matrix~$\Sb$.
For sketching-based approaches (\cf Section~\ref{sxn:sketch}),
we construct $\Sb$ using either the \emph{count-sketch} matrix~\cite{ClaWoo13} as in~\cite{LCZ17}, and
the \emph{sub-sampled randomized Hadamard transform}~(SRHT)~\cite{AilCha09}.
For sampling-based approaches (\cf Appendix~\ref{sxn:appendix:sampling}), we
construct the sampling-and-rescaling matrix $\Sb$ (\cf Algorithm~\ref{algo:1} of Appendix~\ref{sxn:appendix:sampling}) using three different choices of sampling probabilities:
(\emph{i}) uniformly at random,
(\emph{ii}) proportional to column leverage scores, or
(\emph{iii}) proportional to column ridge leverage scores.
Note that constructing~$\Sb$ with uniform sampling probabilities do not in general satisfy the
structural conditions of eqns.~\eqref{eq:struct2}~and~\eqref{eq:struct1}.

For each sketching method, we run Algorithm~\ref{algo:iterative_rda} for $50$ iterations
with a variety of sketch sizes,
and measure
% (\emph{i})
the relative approximation error
$\|\widehat{\Gb} - \Gb\|_F / \|\Gb\|_F$, where
$\Gb$ is computed exactly.
% and (\emph{ii}) the objective sub-optimality $\frac{f(\widehat{\xvec}^{*})}{f(\xvec^*)} - 1$,
% where $f(\xvec) = \|\Ab\xvec - \bvec\|_2^2 + \lambda \|\xvec\|_2^2$ is the
% objective function for ridge regression.
We also randomly divide each dataset into a training set with 60\% examples and a test set of 40\% examples (stratified by label), and measure the classification accuracy on the test set with $\widehat{\Gb}$ estimated from the training set.
For each sketching method, we repeat 20 random trials and report the means and standard errors of the experiment results.

\subsection{Results and Discussion}

\begin{figure*}[!htb]
	\centering
	\subfloat{
		\includegraphics[width=0.24\columnwidth,keepaspectratio]{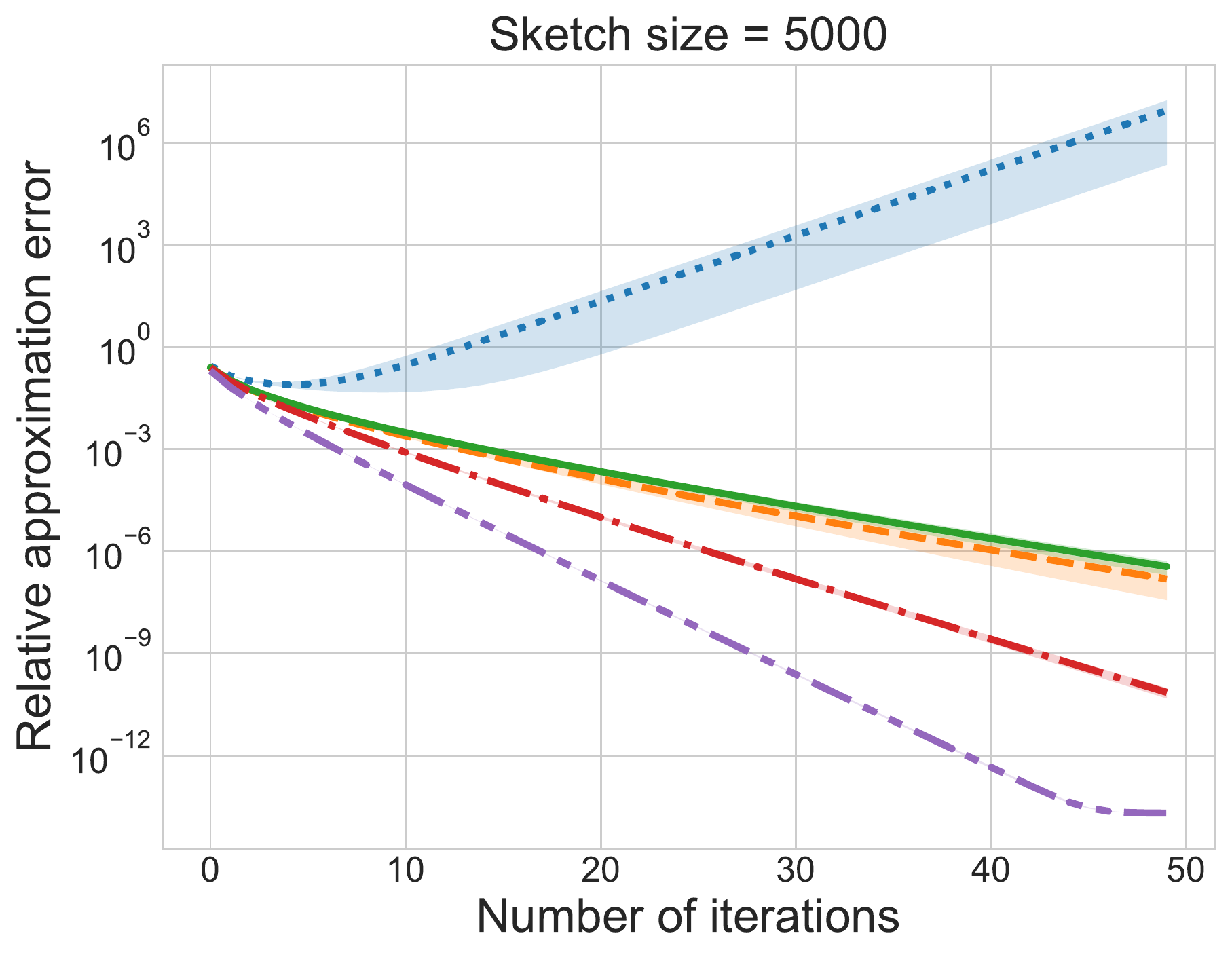}}
	\subfloat{
		\includegraphics[width=0.24\columnwidth,keepaspectratio]{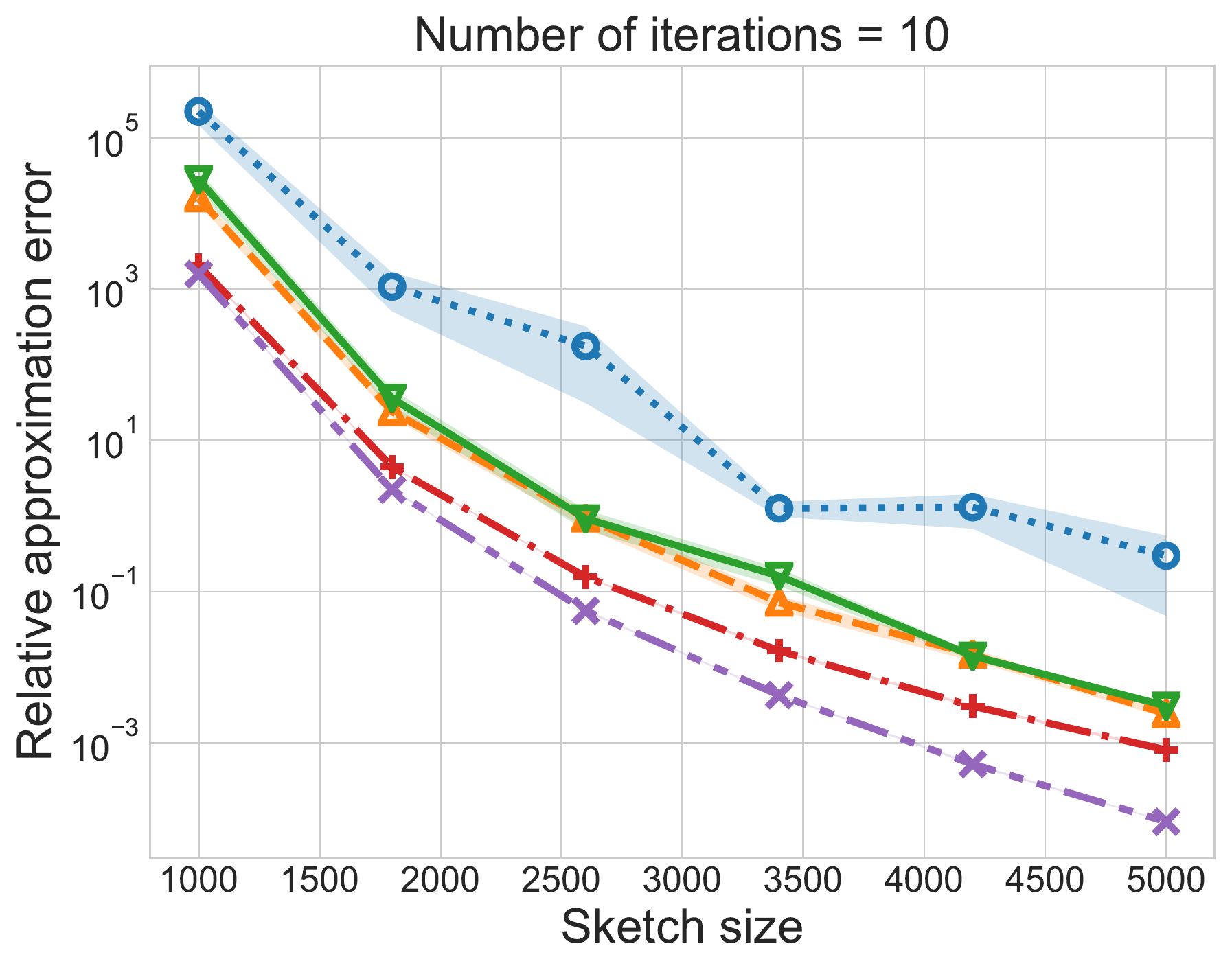}}
	\subfloat{
		\includegraphics[width=0.24\columnwidth,keepaspectratio]{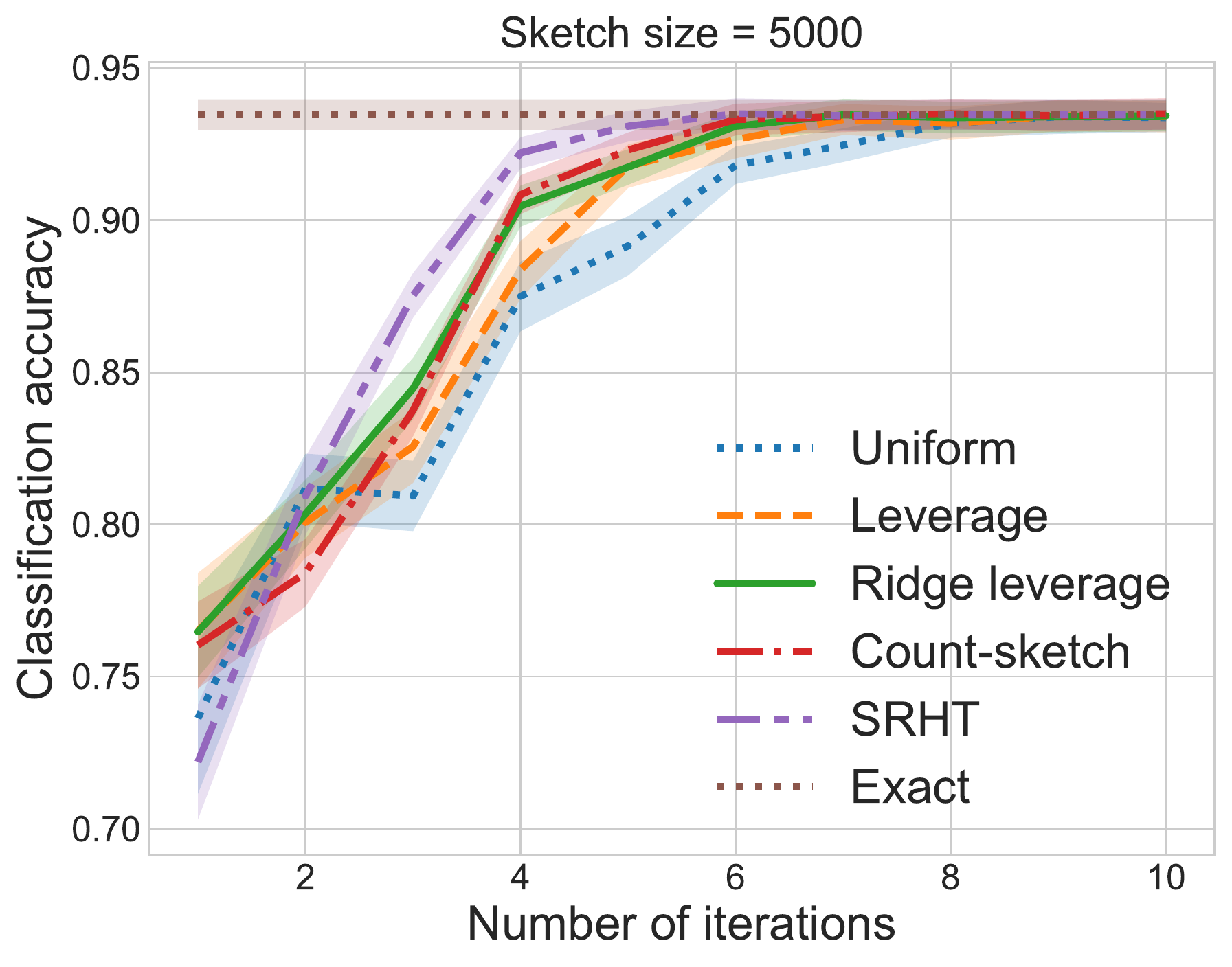}}
	\subfloat{
		\includegraphics[width=0.24\columnwidth,keepaspectratio]{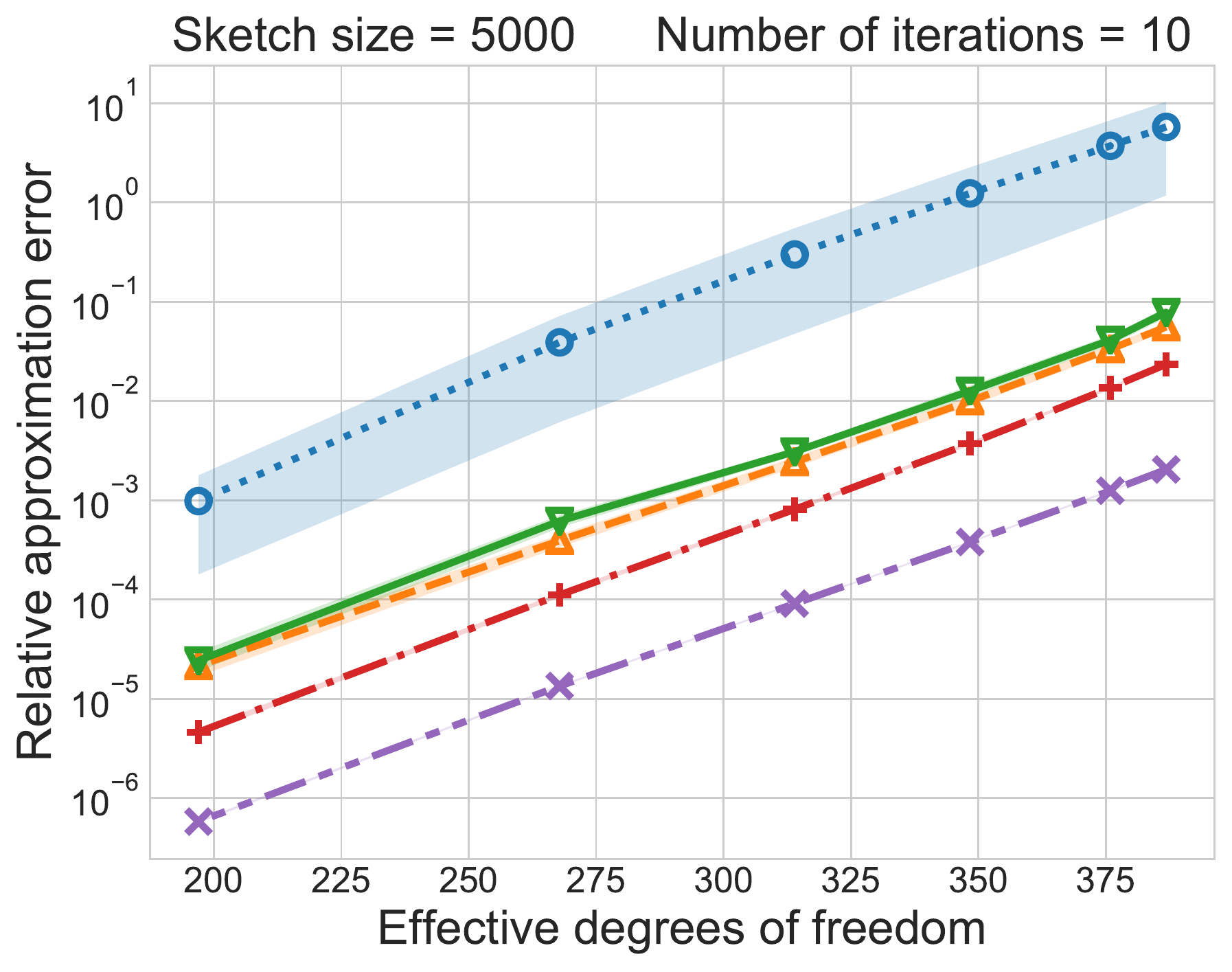}}
	\vspace{-8pt}
	\\
	\addtocounter{subfigure}{-4}
	\subfloat[Error vs.~iterations]{
		\includegraphics[width=0.24\columnwidth,keepaspectratio]{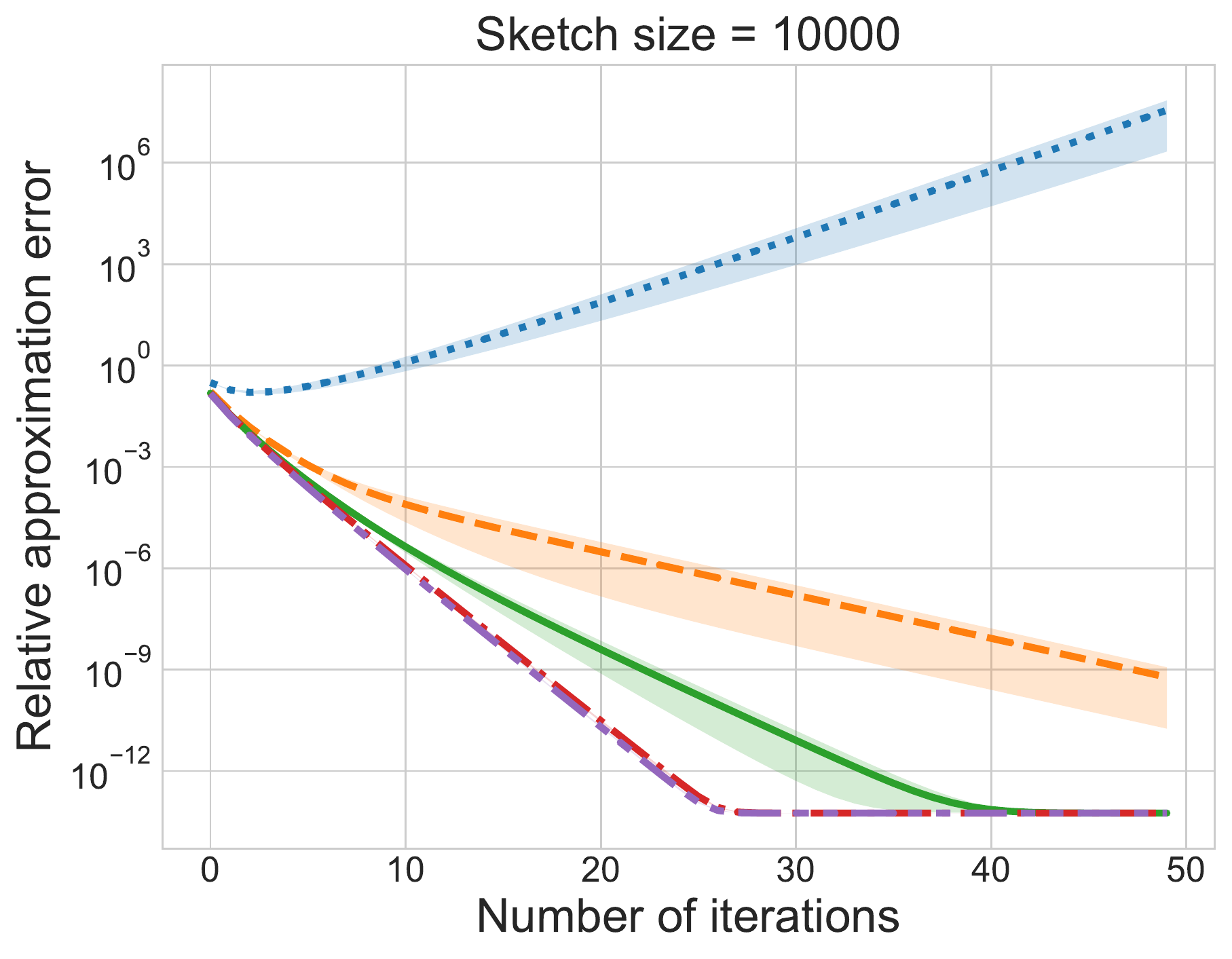}}
	\subfloat[Error vs.~sketch size]{
		\includegraphics[width=0.24\columnwidth,keepaspectratio]{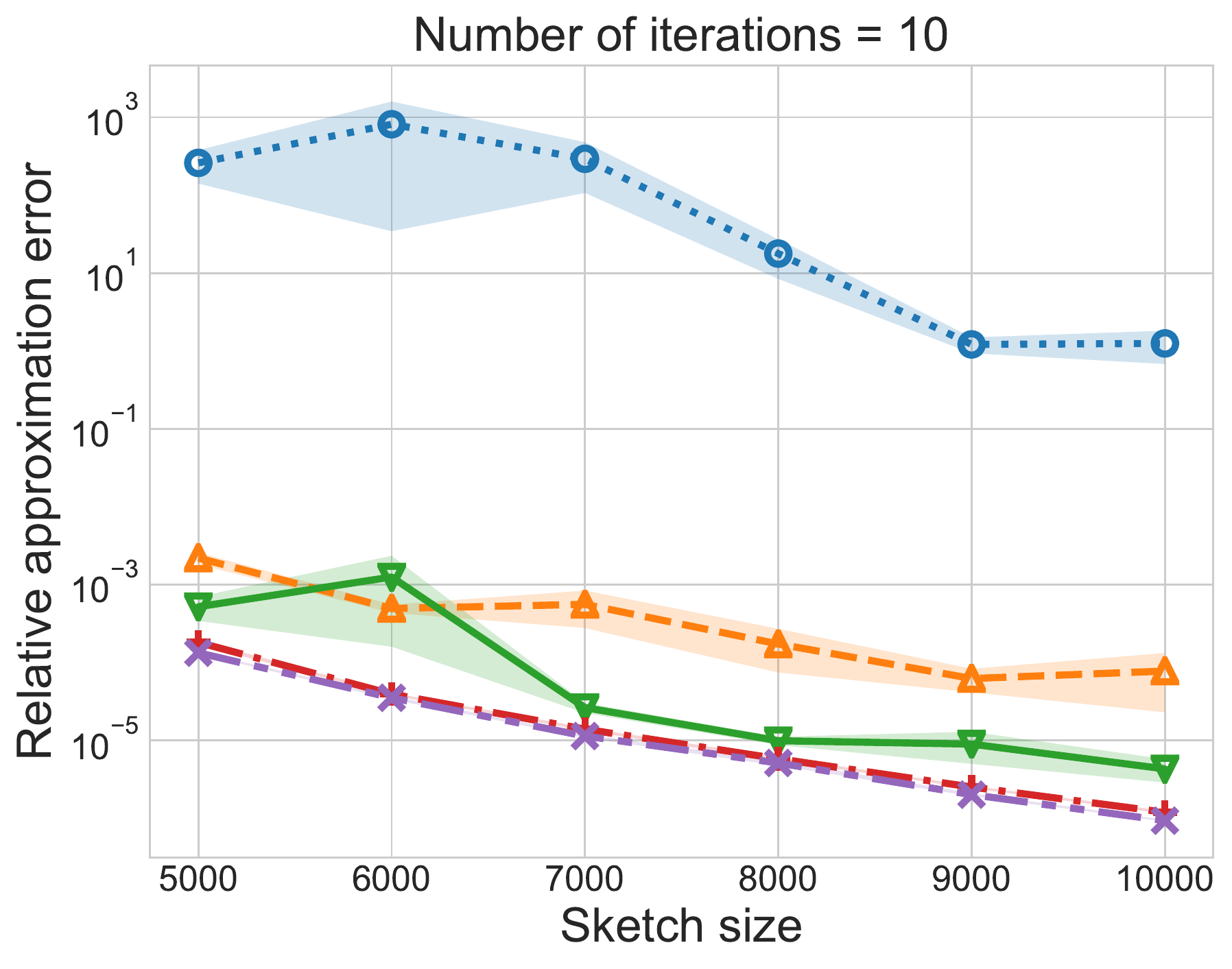}}
	\subfloat[Accuracy vs.~iterations]{
		\includegraphics[width=0.24\columnwidth,keepaspectratio]{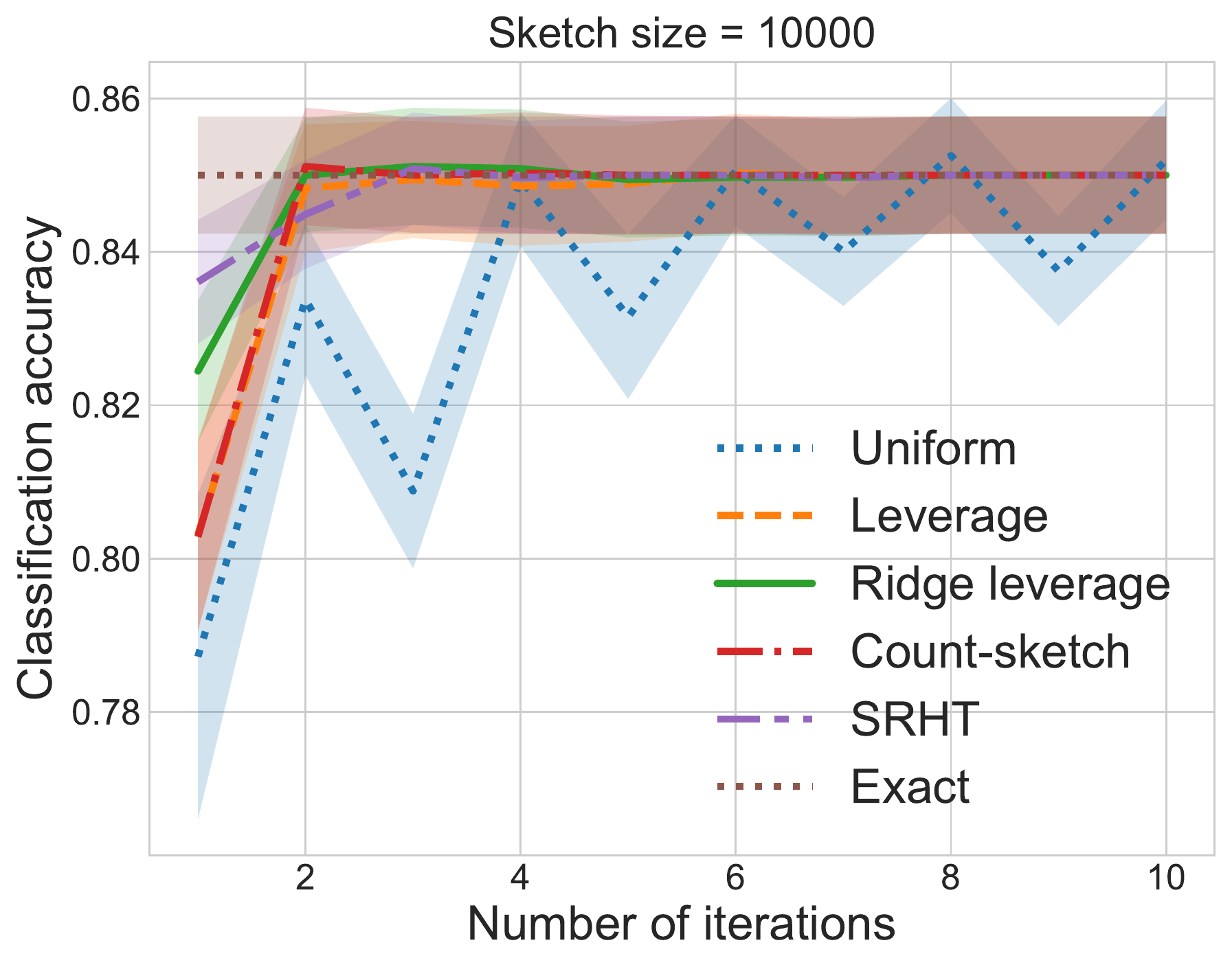}}
	\subfloat[Error vs.~$d_{\lambda}$]{
		\includegraphics[width=0.24\columnwidth,keepaspectratio]{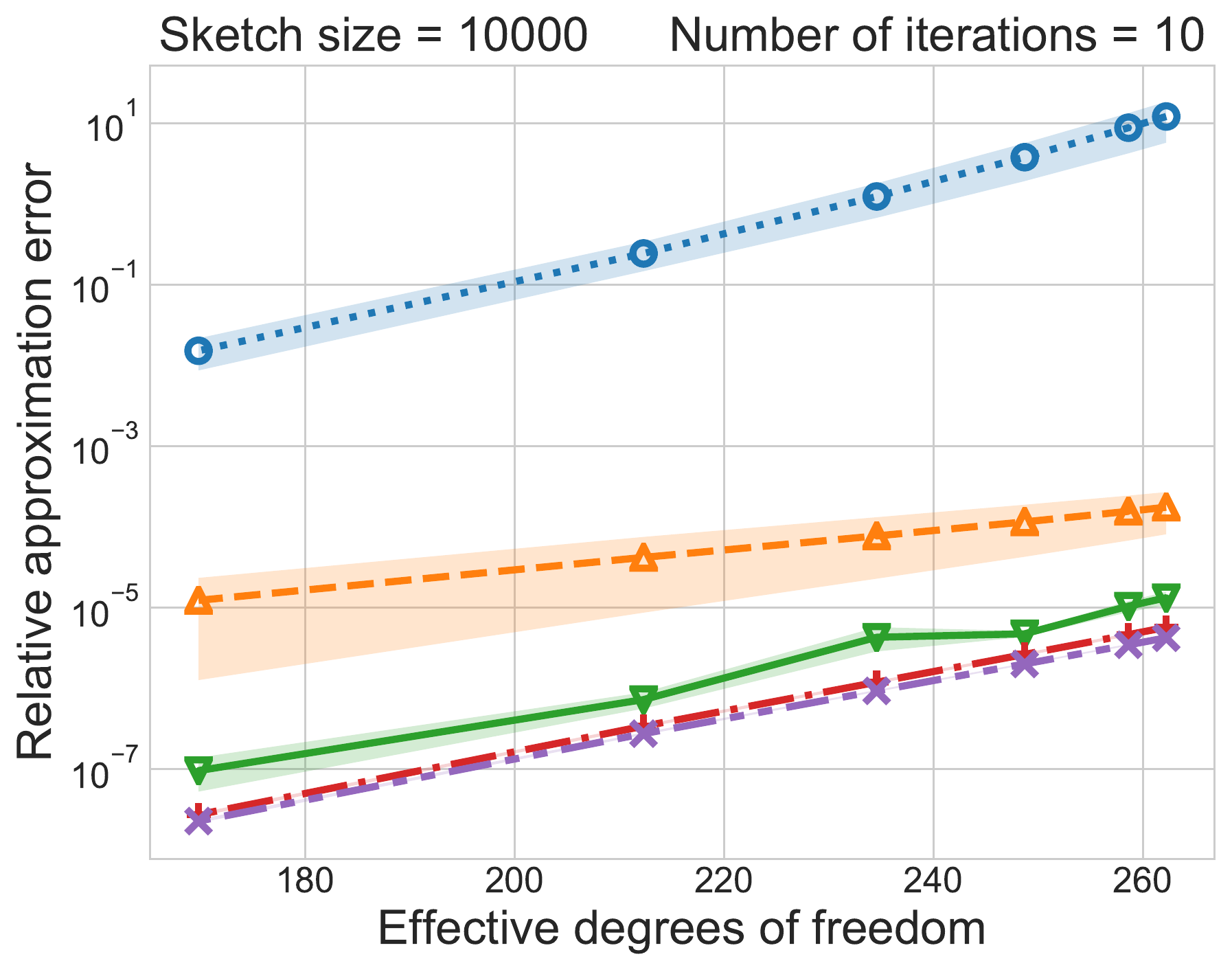}}
	\caption{
		Experiment results on ORL (top row) and PEMS (bottom row); errors are on log-scale.}
	\label{fig:results}
\end{figure*}

In Figure~\ref{fig:results},
the first column plots the relative approximation error (for a fixed sketch size) as the iterative algorithm progresses;
the second column plots the relative approximation error with respect to
varying sketch sizes; and
the third column plots the test classification accuracy obtained using the estimated \smash{$\widehat{\Gb}=\sum_{j=1}^t\widetilde{\Gb}^{(j)}$} after $t = 1, \dots, 10$ iterations.
For count-sketch, SRHT, as well as leverage score and ridge leverage score sampling, we observe that the relative approximation error decays \emph{exponentially} as our iterative algorithm progresses.%
\footnote{Except in the last column of Figure~\ref{fig:results}, we set the regularization parameter to $\lambda = 10$ in the RFDA problem as well as the ridge leverage score sampling probabilities.}
%{}
In particular, constructing the sketching matrix $\Sb$ using the sketching-based approaches appear to achieve slightly improved approximation quality over the sampling-based approaches.
Furthermore, while leverage score and ridge leverage score sampling perform comparably on the ORL dataset, the latter significantly outperforms the former on the PEMS dataset.
This confirms our discussion in Section~\ref{sec:contribs}: % \ref{sxn:appendix:sampling}:
for ridge leverage score sampling, setting $s = \Ocal(\ve^{-2} d_{\lambda} \ln d_{\lambda})$ suffices to satisfy the structural condition of eqn.~(\ref{eq:struct2}), while for leverage scores, setting $s = \Ocal(\ve^{-2}\rho\ln\rho)$ suffices to satisfy the structural condition of eqn.~(\ref{eq:struct1}).
(Recall that $\rho$ can be substantially larger than the effective degrees of freedom $d_\lambda$.)
Finally, we note that the proposed approach of~\cite{LCZ17}~(see Theorem~3 therein for the $d \gg n$ setting) corresponds to running a single iteration of Algorithm~\ref{algo:iterative_rda};
our iterative algorithm yields significant improvements in the approximation quality of the solutions.

% \todo{Do we need to also mention this somewhere in the related work and/or Section 2.1?}

In the last column of Figure~\ref{fig:results},
we keep the design matrix unchanged (fixing $n$)
while varying the regularization parameter $\lambda$, and plot the
relative approximation error against the effective degrees of freedom $d_\lambda$ of the RFDA problem.
% (for a fixed sketch size and the number of iterations).
We observe that the relative approximation error decreases exponentially as $d_\lambda$ decreases;
thus, the sketch size or number of iterations necessary to achieve a certain approximation precision also decreases with $d_\lambda$, even though $n$ remains fixed.

%!TEX root = arxiv_LDA.tex

%!TEX root = arxiv_LDA.tex

\section{Conclusion and Open Problems}\label{sxn:conclusions}
We have presented simple structural results to analyze an iterative, sketching-based RFDA algorithm that guarantees highly accurate solutions when compared to conventional approaches. An obvious open problem is to either improve on the sample size requirement of our sketching matrix or present matching lower bounds to show that our bounds are tight. A second open problem would be to explore similar approaches for other versions of regularized FDA that use, say, the pseudo-inverse of the centered data matrix (see footnote 2).

Finally, an exciting open problem would be to investigate whether the use of different sampling matrices in each iteration of Algorithm~\ref{algo:iterative_rda} (\ie, introducing new ``randomness'' in each iteration) could lead to \textit{provably} improved bounds for our main theorems. We conjecture that this is indeed the case, and we present further experiment results in Appendix~\ref{sxn:appendix-expts} which support our conjecture.
In particular, the results show that using a newly sampled sketching matrix at every iteration enables faster convergence as the iterations progress, and also reduces the minimum sketch size necessary for Algorithm~\ref{algo:iterative_rda} to converge.

%%%%%%%%%%%%%%%%%%%%%%%%

% \clearpage

\bibliographystyle{plainnat}
{
\bibliography{paper_index}
}

% \clearpage

% \begin{center}
% \textbf{\LARGE
% Appendix to \\
% Randomized Iterative Algorithms for \\
% Fisher Discriminant Analysis
% }
% \end{center}

\begin{appendices}
%
%!TEX root = arxiv_LDA.tex

\section{Preliminary results and full SVD representation}
We start by reviewing a result regarding the convergence of a matrix \textit{von Neumann} series for $\left(\Ib-\Pb\right)^{-1}$. This will be an important tool in our analysis.
\begin{proposition}\label{eq:matsum}
	Let $\Pb$ be any square matrix with $\nbr{\Pb}_2<1$. Then $\left(\Ib-\Pb\right)^{-1}$ exists and
	\begin{flalign*}
	\left(\Ib-\Pb\right)^{-1}=\Ib+\sum_{\ell=1}^{\infty}\Pb^\ell.
	\end{flalign*}
\end{proposition}

\paragraph{Full SVD representation.}
The full SVD representation of $\Ab$ is given by
$\Ab=\Ub_f\Sigmab_f\Vb_f^\ts$,
where $\Ub_f\Ub_f^\ts=\Ub_f^\ts\Ub_f=\Ib_n$, $\Vb_f\Vb_f^\ts=\Vb_f^\ts\Vb_f=\Ib_d$,
$\Sigmab_f=\begin{pmatrix}
\Sigmab & \mathbf{0}\\
\mathbf{0} & \mathbf{0}
\end{pmatrix}\in\RR{n}{d}$,
$\Ub_f=\begin{pmatrix}
\Ub & \Ub_{\perp}
\end{pmatrix}$ and $\Vb_f=\begin{pmatrix}
\Vb & \Vb_{\perp}
\end{pmatrix}$.
Here, $\Ub_{\perp}$ and $\Vb_{\perp}$ comprise of the last $n-\rho$ and $d-\rho$ columns of $\Ub_f$ and $\Vb_f$, respectively.

\section{EVD-based algorithms for FDA}\label{app:evd}
%
%\begin{minipage}{.47\linewidth}
For RFDA, we quote an EVD-based algorithm along with an important result from \citet{ZDXJ10} which together are the building blocks of our iterative framework.
%Actually, \citep{ZDXJ10} only presented the algorithm and proved the corresponding result pertaining to the RFDA problem \eqref{eq:rfda3}, not for the low-rank PFDA problem \eqref{eq:psrfda3}. However, similar results for problem \eqref{eq:psrfda3} was implicit in their paper. In our sketching-based structure, we need to treat the problems \eqref{eq:rfda3} and \eqref{eq:psrfda3} separately which enables us to state the algorithms and the corresponding theorems explicitly.
Let $\Mb\in\RR{c}{c}$
%and $\Nb\in\RR{c}{c}$
be the matrix such that $\Mb=\Omegab^\ts\Ab\Gb$.
%and $\Nb=\Omegab^\ts\Ab\Fb$
Clearly, $\Mb$ is symmetric and positive semi-definite.
%\end{minipage}
%
\hspace{.01\linewidth}
%
%\begin{minipage}{.5\linewidth}
\begin{algorithm}[H]
	\caption{Algorithm for RFDA problem \eqref{eq:rfda3}}\label{algo:EVD_RFDA}
	\begin{algorithmic}
		\State \textbf{Input:} $\Ab\in\RR{n}{d}$, $\Omegab\in\RR{n}{c}$ and $\lambda>0$;
		\State $\Gb \gets (\Ab^\ts\Ab+\lambda\Ib_d)^{-1}\Ab^\ts\Omegab$ ;
		\State $\Mb\gets \Omegab^\ts\Ab\Gb$;
		\State Compute thin SVD $\Mb=\Vb_{\Mb}\Sigmab_{\Mb}\Vb_{\Mb}^\ts$;
		\State \textbf{Output:} $\Xb=\Gb\,\Vb_{\Mb}$
	\end{algorithmic}
\end{algorithm}
%\end{minipage}

%\begin{algorithm}
%	\caption{Algorithm for Low-rank PFDA Problem \eqref{eq:psrfda3}}\label{algo:EVD_pfda}
%	\begin{algorithmic}
%		\State 1.~\textbf{Input:} $\Ab\in\RR{n}{d}$; $\Omegab\in\RR{n}{c}$, $\lambda>0$ and $k>0$;
%		\State 2.~Compute $\Ab_{k}=\Ub_k\Sigmab_k\Vb_k^\ts$: the best $k$-rank approximation of $\Ab$;
%		\State 3.~Calculate $\Fb \gets (\Ab_k^\ts\Ab_k)^{\dagger}\Ab^\ts\Omegab$ and $\Nb\gets \Omegab^\ts\Ab\Fb$;
%		\State 4.~Compute thin SVD of $\Nb=\Vb_{\Nb}\Sigmab_{\Nb}\Vb_{\Nb}^\ts$;
%		\State 5.~\textbf{Output:} $\Xb=\Fb\,\Vb_{\Nb}$ as the solution of the problem \eqref{eq:psrfda3}
%	\end{algorithmic}
%\end{algorithm}

\begin{theorem}\label{thm:EVD_RFDA}
	Using Algorithm~\ref{algo:EVD_RFDA}, let $\Xb$ be the solution of problem \eqref{eq:rfda3} , then we have
	$$\Xb\Xb^\ts=~\Gb\,\Gb^{\ts}$$
\end{theorem}

%\begin{theorem}\label{thm:EVD_pfda}
%	Using Algorithm~\ref{algo:EVD_pfda}, let $\Xb$ be the solution of problem \eqref{eq:psrfda3}, then we have
%	$$\Xb\Xb^\ts=~\Fb\,\Fb^{\ts}$$
%\end{theorem}

For any two data points $\wb_1,~\wb_2\in\R{d}$, Theorem~\ref{thm:EVD_RFDA} implies
\begin{flalign*}
&(\wb_1-\wb_2)^\ts\Xb\Xb^\ts(\wb_1-\wb_2)=(\wb_1-\wb_2)^\ts\Gb\,\Gb^{\ts}(\wb_1-\wb_2)\\
\iff&\|(\wb_1-\wb_2)^\ts\Xb\|_2=\|(\wb_1-\wb_2)^\ts\Gb\|_2.
\end{flalign*}

%Similarly, Theorem~\ref{thm:EVD_pfda} implies $\|(\ab_1-\ab_2)^\ts\Xb\|_2=\|(\ab_1-\ab_2)^\ts\Fb\|_2$.
Theorem~\ref{thm:EVD_RFDA}
%and Theorem~\ref{thm:EVD_pfda}
indicates that if we use any distance-based classification method such as $k$-nearest neighbors, both $\Xb$ and $\Gb$
%(or, $\Fb$)
shares the same property. Thus, we may shift our interest from $\Xb$ to $\Gb$.
%(or, $\Fb$ for low-rank PFDA).

%!TEX root = arxiv_LDA.tex

\section{Proof of Theorem~\ref{thm:iterative2}}\label{app:ridlev}
%where $\Rb=\Ib_n+\lambda\Sigmab^{-2}$.

\begin{proof}[Proof of Lemma~\ref{lem:alt_xopt}.]
	Using the full SVD representation of $\Ab$ we have
	\begin{flalign}
	\Gb^{(j)}=&~\Vb_f\Sigmab_f^\ts\Ub_f^\ts(\Ub_f\Sigmab_f\Sigmab_f^\ts\Ub_f^\ts+\lambda\Ub_f\Ub_f^\ts)^{-1}\Lb^{(j)}\nonumber\\
	=&~\Vb_f\Sigmab_f^\ts(\Sigmab_f\Sigmab_f^\ts+\lambda\Ib_n)^{-1}\Ub_f^\ts\Lb^{(j)}\nonumber\\
	=&~\begin{pmatrix}
	\Vb & \Vb_{\perp}
	\end{pmatrix}\begin{pmatrix}
	\Sigmab & \mathbf{0}\\
	\mathbf{0} & \mathbf{0}
	\end{pmatrix}\left[\begin{pmatrix}
	\Sigmab^2 & \mathbf{0}\\
	\mathbf{0} & \mathbf{0}
	\end{pmatrix}+\lambda\Ib_n\right]^{-1}\begin{pmatrix}
	\Ub^\ts\\
	\Ub_{\perp}^\ts
	\end{pmatrix}\Lb^{(j)}\nonumber\\
	=&~\begin{pmatrix}
	\Vb & \Vb_{\perp}
	\end{pmatrix}\begin{pmatrix}
	\Sigmab & \mathbf{0}\\
	\mathbf{0} & \mathbf{0}
	\end{pmatrix}\left[\begin{pmatrix}
	\Sigmab^2+\lambda\Ib_\rho & \mathbf{0}\\
	\mathbf{0} & \lambda\Ib_{n-\rho}
	\end{pmatrix}\right]^{-1}\begin{pmatrix}
	\Ub^\ts\\
	\Ub_{\perp}^\ts
	\end{pmatrix}\Lb^{(j)}\nonumber\\
	=&~\begin{pmatrix}
	\Vb & \Vb_{\perp}
	\end{pmatrix}\begin{pmatrix}
	\Sigmab & \mathbf{0}\\
	\mathbf{0} & \mathbf{0}
	\end{pmatrix}\begin{pmatrix}
	(\Sigmab^2+\lambda\Ib_\rho)^{-1} & \mathbf{0}\\
	\mathbf{0} & \frac{1}{\lambda}\Ib_{n-\rho}
	\end{pmatrix}\begin{pmatrix}
	\Ub^\ts\\
	\Ub_{\perp}^\ts
	\end{pmatrix}\Lb^{(j)}\nonumber\\
	=&~\begin{pmatrix}
	\Vb & \Vb_{\perp}
	\end{pmatrix}\begin{pmatrix}
	\Sigmab(\Sigmab^2+\lambda\Ib_\rho)^{-1} & \mathbf{0}\\
	\mathbf{0} & \mathbf{0}
	\end{pmatrix}\begin{pmatrix}
	\Ub^\ts\\
	\Ub_{\perp}^\ts
	\end{pmatrix}\Lb^{(j)}\nonumber\\
	=&~\Vb\Sigmab(\Sigmab^2+\lambda\Ib_\rho)^{-1}\Ub^\ts\Lb^{(j)}\nonumber\\
	=&~\Vb\Sigmab\Sigmab^{-1}(\Ib_\rho+\lambda\Sigmab^{-2})^{-1}\Sigmab^{-1}\Ub^\ts\Lb^{(j)}\nonumber\\
	=&~\Vb\Sigmab_{\lambda}^2\Sigmab^{-1}\Ub^\ts\Lb^{(j)},\label{eq:fda11}
	\end{flalign}
which completes the proof.
\end{proof}

\begin{proof}[Detailed proof of Lemma~\ref{lem:struct3}.]
	First, using SVD of $\Ab$, we express $\widetilde{\Gb}^{(j)}$ in terms of $\Gb^{(j)}$.
	\begin{flalign}
	\widetilde{\Gb}^{(j)}=&~\Vb_f\Sigmab_f^\ts\Ub_f^\ts(\Ub_f\Sigmab_f\Vb_f^\ts\Sb\Sb^\ts\Vb_f\Sigmab_f^\ts\Ub_f^\ts+\lambda\Ub_f\Ub_f^\ts)^{-1}\Lb^{(j)}\nonumber\\
	=&~\Vb_f\Sigmab_f^\ts(\Sigmab_f\Vb_f^\ts\Sb\Sb^\ts\Vb_f\Sigmab_f^\ts+\lambda\Ib_n)^{-1}\Ub_f^\ts\Lb^{(j)}\nonumber\\
	=&~\begin{pmatrix}
	\Vb & \Vb_{\perp}
	\end{pmatrix}\begin{pmatrix}
	\Sigmab & \mathbf{0}\\
	\mathbf{0} & \mathbf{0}
	\end{pmatrix}\left[\begin{pmatrix}
	\Sigmab\Vb^\ts\Sb\Sb^\ts\Vb\Sigmab & \mathbf{0}\\
	\mathbf{0} & \mathbf{0}
	\end{pmatrix}+\lambda\Ib_n\right]^{-1}\begin{pmatrix}
	\Ub^\ts\\
	\Ub_{\perp}^\ts
	\end{pmatrix}\Lb^{(j)}\nonumber\\
	=&~\begin{pmatrix}
	\Vb & \Vb_{\perp}
	\end{pmatrix}\begin{pmatrix}
	\Sigmab & \mathbf{0}\\
	\mathbf{0} & \mathbf{0}
	\end{pmatrix}\left[\begin{pmatrix}
	\Sigmab\Vb^\ts\Sb\Sb^\ts\Vb\Sigmab+\lambda\Ib_\rho & \mathbf{0}\\
	\mathbf{0} & \lambda\Ib_{n-\rho}
	\end{pmatrix}\right]^{-1}\begin{pmatrix}
	\Ub^\ts\\
	\Ub_{\perp}^\ts
	\end{pmatrix}\Lb^{(j)}\nonumber\\
	=&~\begin{pmatrix}
	\Vb & \Vb_{\perp}
	\end{pmatrix}\begin{pmatrix}
	\Sigmab & \mathbf{0}\\
	\mathbf{0} & \mathbf{0}
	\end{pmatrix}\begin{pmatrix}
	(\Sigmab\Vb^\ts\Sb\Sb^\ts\Vb\Sigmab+\lambda\Ib_\rho)^{-1} & \mathbf{0}\\
	\mathbf{0} & \frac{1}{\lambda}\Ib_{n-\rho}
	\end{pmatrix}\begin{pmatrix}
	\Ub^\ts\\
	\Ub_{\perp}^\ts
	\end{pmatrix}\Lb^{(j)}\nonumber\\
	=&~\begin{pmatrix}
	\Vb & \Vb_{\perp}
	\end{pmatrix}\begin{pmatrix}
	\Sigmab(\Sigmab\Vb^\ts\Sb\Sb^\ts\Vb\Sigmab+\lambda\Ib_\rho)^{-1} & \mathbf{0}\nonumber\\
	\mathbf{0} & \mathbf{0}
	\end{pmatrix}\begin{pmatrix}
	\Ub^\ts\\
	\Ub_{\perp}^\ts
	\end{pmatrix}\Lb^{(j)}\nonumber\\
	=&~\Vb\Sigmab(\Sigmab\Vb^\ts\Sb\Sb^\ts\Vb\Sigmab+\lambda\Ib_\rho)^{-1}\Ub^\ts\Lb^{(j)}\label{eq:fda14}\\
	=&~\Vb\Sigmab\left(\Sigmab\Sigmab_\lambda^{-1}\left(\Sigmab_\lambda\Vb^\ts\Sb\Sb^\ts\Vb\Sigmab_\lambda\right)\Sigmab_\lambda^{-1}\Sigmab+\lambda\Ib_\rho\right)^{-1}\Ub^\ts\Lb^{(j)}\nonumber\\
	=&~\Vb\Sigmab\left(\Sigmab\Sigmab_\lambda^{-1}\left(\Sigmab_\lambda^2+\Eb\right)\Sigmab_\lambda^{-1}\Sigmab+\lambda\Ib_\rho\right)^{-1}\Ub^\ts\Lb^{(j)}\label{eq:fda16}\\
	=&~\Vb\Sigmab\left(\Sigmab\Sigmab_\lambda^{-1}\left(\Sigmab_\lambda^2+\Eb\right)\Sigmab_\lambda^{-1}\Sigmab+\lambda\Sigmab\Sigmab_\lambda^{-1}\Sigmab_\lambda\Sigmab^{-2}\Sigmab_\lambda\Sigmab_\lambda^{-1}\Sigmab\right)^{-1}\Ub^\ts\Lb^{(j)}\nonumber\\
	=&~\Vb\Sigmab\left(\Sigmab\Sigmab_\lambda^{-1}\left(\Sigmab_\lambda^2+\Eb+\lambda\Sigmab_\lambda\Sigmab^{-2}\Sigmab_\lambda\right)\Sigmab_\lambda^{-1}\Sigmab\right)^{-1}\Ub^\ts\Lb^{(j)}\nonumber\\
	=&~\Vb\Sigmab\left(\Sigmab\Sigmab_\lambda^{-1}\left(\Ib_\rho+\Eb\right)\Sigmab_\lambda^{-1}\Sigmab\right)^{-1}\Ub^\ts\Lb^{(j)}.\label{eq:fda17}
	\end{flalign}
	Eqn.~\eqref{eq:fda16} used the fact that $\Sigmab_\lambda\Vb^\ts\Sb\Sb^\ts\Vb\Sigmab_\lambda=\Sigmab_\lambda^2+\Eb$.
	Eqn.~\eqref{eq:fda17} follows from the fact that $\Sigmab_\lambda^2+\lambda\Sigmab_\lambda\Sigmab^{-2}\Sigmab_\lambda\in\RR{n}{n}$ is a diagonal matrix with $i$-th diagonal element
	\begin{flalign*} \left(\Sigmab_\lambda^2+\lambda\Sigmab_\lambda\Sigmab^{-2}\Sigmab_\lambda\right)_{ii}=\frac{\sigma_i^2}{\sigma_i^2+\lambda}+\frac{\lambda}{\sigma_i^2+\lambda}=1,
	\end{flalign*}
	for any $i=1\ldots \rho$.
	Thus, we have $\left(\Sigmab_\lambda^2+\lambda\Sigmab_\lambda\Sigmab^{-2}\Sigmab_\lambda\right)=\Ib_\rho$.
	Since $\|\Eb\|_2<1$, Proposition~\ref{eq:matsum}  implies that $(\Ib_\rho+\Eb)^{-1}$ exists and $$(\Ib_\rho+\Eb)^{-1}=~\Ib_\rho+\sum_{\ell=1}^{\infty}(-1)^\ell\Eb^\ell=\Ib_\rho+\Qb.$$
	Thus, eqn.~\eqref{eq:fda17} can further be expressed as
	\begin{flalign}
	\widetilde{\Gb}^{(j)}=&~\Vb\Sigmab\Sigmab^{-1}\Sigmab_\lambda\left(\Ib_\rho+\Eb\right)^{-1}\Sigmab_\lambda\Sigmab^{-1}\Ub^\ts\Lb^{(j)}\nonumber\\
	=&~\Vb\Sigmab_\lambda\left(\Ib_\rho+\Qb\right)\Sigmab_\lambda\Sigmab^{-1}\Ub^\ts\Lb^{(j)}\nonumber\\
	=&~\Vb\Sigmab_\lambda^2\Sigmab^{-1}\Ub^\ts\Lb^{(j)}+\Vb\Sigmab_\lambda\Qb\Sigmab_\lambda\Sigmab^{-1}\Ub^\ts\Lb^{(j)}\nonumber\\
	=&~\Gb^{(j)}+\Vb\Sigmab_\lambda\Qb\Sigmab_\lambda\Sigmab^{-1}\Ub^\ts\Lb^{(j)},\label{eq:fda18}
	\end{flalign}
	where the last line follows from Lemma~\ref{lem:alt_xopt}.
	Further, we have
	\begin{flalign}
	\nbr{\Qb}_2
	& = \nbr{\sum_{\ell=1}^\infty(-1)^\ell\Eb^\ell}_2
	\le\sum_{\ell=1}^\infty\nbr{\Eb^\ell}_2
	\le\sum_{\ell=1}^\infty\nbr{\Eb}_2^\ell
	\le\sum_{\ell=1}^\infty\left(\frac{\ve}{2}\right)^\ell
	= \frac{\ve/2}{1-\ve/2}
	% \le\frac{\frac{\ve}{4\sqrt{2}}}{1/2}
	\le \ve \,, \label{eq:272}
	\end{flalign}
	where we used the triangle inequality, the sub-multiplicativity of the spectral norm, and the fact that $\ve \le 1$.
	Next, we combine eqns.~\eqref{eq:fda18} and~\eqref{eq:272} to get
	\begin{flalign}
	\nbr{(\wb-\mb)^\ts(\widetilde{\Gb}^{(j)}-\Gb^{(j)})}_2=&~\nbr{(\wb-\mb)^\ts\Vb\Sigmab_\lambda\Qb\Sigmab_\lambda\Sigmab^{-1}\Ub^\ts\Lb^{(j)}}_2\nonumber\\
	\le&~\nbr{(\wb-\mb)^\ts\Vb}_2\nbr{\Sigmab_\lambda}_2\nbr{\Qb}_2\nbr{\Sigmab_\lambda\Sigmab^{-1}\Ub^\ts\Lb^{(j)}}_2\nonumber\\
	\le&~\ve\,\nbr{(\wb-\mb)^\ts\Vb}_2\nbr{\Sigmab_\lambda\Sigmab^{-1}\Ub^\ts\Lb^{(j)}}_2\nonumber\\
	=&~\ve\,\nbr{\Vb\Vb^\ts(\wb-\mb)}_2\nbr{\Sigmab_\lambda\Sigmab^{-1}\Ub^\ts\Lb^{(j)}}_2,\label{eq:fda1_15}
%	=&~\ve~\|\alphab\|_2\nbr{\Sigmab_\lambda\Sigmab^{-1}\Ub^\ts\Lb^{(j)}}_2
	%=&~2\ve~\nbr{(\wb-\mb)^\ts\Vb}_2=~2\ve~\nbr{\alphab}_2
	\end{flalign}
	which completes the proof.
\end{proof}

\begin{proof}[Proof of Lemma~\ref{lem:induction}.]
	We prove the lemma using induction on $t$.
	Note that $\Lb^{(1)}=\Omegab$. So, for $t=1$, eqn.~\eqref{eqn:pdits} boils down to $$\Gb^{(1)}=\Ab^\ts(\Ab\Ab^\ts+\lambda\Ib_n)^{-1}\Lb^{(1)}=\Gb. $$
	For $t=2$, we get
	\begin{flalign*}
	\Gb^{(2)}
	=&~\Ab^\ts(\Ab\Ab^\ts+\lambda\Ib_n)^{-1}\Lb^{(2)} \\
	=&~\Ab^\ts(\Ab\Ab^\ts+\lambda\Ib_n)^{-1}(\Lb^{(1)}-\lambda\Yb^{(1)}-\Ab\widetilde{\Gb}^{(1)})\\
	=&~\Ab^\ts(\Ab\Ab^\ts+\lambda\Ib_n)^{-1}(\Lb^{(1)}-(\Ab\Ab^\ts+\lambda\Ib_n)(\Ab\Sb\Sb^\ts\Ab^\ts+\lambda\Ib_n)^{-1}\Lb^{(1)})\\
	=&~\Ab^\ts(\Ab\Ab^\ts+\lambda\Ib_n)^{-1}\Lb^{(1)}-\Ab^\ts(\Ab\Sb\Sb^\ts\Ab^\ts+\lambda\Ib_n)^{-1}\Lb^{(1)}\\
	=&~\Gb-\widetilde{\Gb}^{(1)} .
	\end{flalign*}
%
	% Thus, eqn.~\eqref{eq:iterative2_1} is satisfied for $t=1,2$.
	Now, suppose eqn.~\eqref{eq:iterative2_1} is also true for $t=p$, \ie,
	\begin{flalign}
	\Gb^{(p)}=\Gb-\sum_{j=1}^{p-1}\widetilde{\Gb}^{(j)}. \label{eq:iterative2_2}
	\end{flalign}
	Then, for $t=p+1$, we can express $\Gb^{(t)}$ as
	\begin{flalign*}
	\Gb^{(p+1)}
	=&~\Ab^\ts(\Ab\Ab^\ts+\lambda\Ib_n)^{-1}\Lb^{(p+1)} \\
	=&~\Ab^\ts(\Ab\Ab^\ts+\lambda\Ib_n)^{-1}(\Lb^{(p)}-\lambda\Yb^{(p)}-\Ab\widetilde{\Gb}^{(p)})\\
	=&~\Ab^\ts(\Ab\Ab^\ts+\lambda\Ib_n)^{-1}(\Lb^{(p)}-(\Ab\Ab^\ts+\lambda\Ib_n)(\Ab\Sb\Sb^\ts\Ab^\ts+\lambda\Ib_n)^{-1}\Lb^{(p)})\\
	=&~\Ab^\ts(\Ab\Ab^\ts+\lambda\Ib_n)^{-1}\Lb^{(p)}-\Ab^\ts(\Ab\Sb\Sb^\ts\Ab^\ts+\lambda\Ib_n)^{-1}\Lb^{(p)}\\
	=&~\Gb^{(p)}-\widetilde{\Gb}^{(p)}=(\Gb-\sum_{j=1}^{p-1}\widetilde{\Gb}^{(j)})-\widetilde{\Gb}^{(p)}=\Gb-\sum_{j=1}^{p}\widetilde{\Gb}^{(j)}
	\end{flalign*}
	where the second equality in the last line follows from eqn.~\eqref{eq:iterative2_2}. By the induction principle, we have proven eqn.~\eqref{eq:iterative2_1}.
\end{proof}

\begin{remark}
	Using Lemma~\ref{lem:induction} and Lemma~\ref{lem:struct3} consecutively, we have
	\begin{flalign}
	& \nbr{(\wb-\mb)^\ts(\widehat{\Gb} -\Gb)}_2 \nonumber\\
	=\ & \nbr{(\wb-\mb)^\ts(\sum_{j=1}^{t}\widetilde{\Gb}^{(j)}-\Gb)}_2
	= \nbr{(\wb-\mb)^\ts(\widetilde{\Gb}^{(t)}-(\Gb-\sum_{j=1}^{t-1}\widetilde{\Gb}^{(j)}))}_2\nonumber\\
	=\ & \nbr{(\wb-\mb)^\ts(\widetilde{\Gb}^{(t)}-\Gb^{(t)})}_2\le~\ve~\nbr{\Vb\Vb^\ts(\wb-\mb)}_2\nbr{\Sigmab_\lambda\Sigmab^{-1}\Ub^\ts\Lb^{(t)}}_2.\label{eq:fda_iter2}
	\end{flalign}
\end{remark}

%%%%%%%%%%%%%%%%%%%%%%%%%%%%%%%%%%%%%%%%%%%%%%%%%%%%%%%%%%%%%%%%%
The next bound provides a critical inequality that can be used recursively to establish Theorem~\ref{thm:iterative2}.

\begin{proof}[Detailed proof of Lemma~\ref{lem:fda_recur}.]
From Algorithm~\ref{algo:iterative_rda}, we have for $j=1\ldots t-1$
	\begin{flalign}
	\Lb^{(j+1)}=&~\Lb^{(j)}-\lambda\Yb^{(j)}-\Ab\widetilde{\Gb}^{(j)}\nonumber\\
	=&~\Lb^{(j)}-(\Ab\Ab^\ts+\lambda\Ib_n)(\Ab\Sb\Sb^\ts\Ab^\ts+\lambda\Ib_n)^{-1}\Lb^{(j)}.\label{eq:fda_recur6}
	\end{flalign}
	Now, starting with the full SVD of $\Ab$, we get
	\begin{flalign}
	&~(\Ab\Ab^\ts+\lambda\Ib_n)(\Ab\Sb\Sb^\ts\Ab^\ts+\lambda\Ib_n)^{-1}\Lb^{(j)}\nonumber\\
	=&~\left(\Ub_f\Sigmab_f\Sigmab_f^\ts\Ub_f^\ts+\lambda\Ub_f\Ub_f^\ts\right)\left(\Ub_f\Sigmab_f\Vb_f^\ts\Sb\Sb^\ts\Vb_f\Sigmab_f^\ts\Ub_f^\ts+\lambda\Ub_f\Ub_f^\ts\right)^{-1}\Lb^{(j)}\nonumber\\
	=&~\Ub_f\left(\Sigmab_f\Sigmab_f^\ts+\lambda\Ib_n\right)\Ub_f^\ts\Ub_f\left(\Sigmab_f\Vb_f^\ts\Sb\Sb^\ts\Vb_f\Sigmab_f^\ts+\lambda\Ib_n\right)^{-1}\Ub_f^\ts\Lb^{(j)}\nonumber\\
	=&~\Ub_f\left(\Sigmab_f\Sigmab_f^\ts+\lambda\Ib_n\right)\left(\Sigmab_f\Vb_f^\ts\Sb\Sb^\ts\Vb_f\Sigmab_f^\ts+\lambda\Ib_n\right)^{-1}\Ub_f^\ts\Lb^{(j)}\nonumber\\
	=&~\Ub_f\begin{pmatrix}
	\Sigmab^2+\lambda\Ib_\rho & \mathbf{0}\\
	\mathbf{0} &\lambda\Ib_{n-\rho}
	\end{pmatrix}\begin{pmatrix}
	(\Sigmab\Vb^\ts\Sb\Sb^\ts\Vb\Sigmab+\lambda\Ib_\rho)^{-1} & \mathbf{0}\\
	\mathbf{0} & \frac{1}{\lambda}\Ib_{n-\rho}
	\end{pmatrix}\Ub_f^\ts\Lb^{(j)}\nonumber\\
	=&~\Ub_f\begin{pmatrix}
	(\Sigmab^2+\lambda\Ib_\rho)(\Sigmab\Vb^\ts\Sb\Sb^\ts\Vb\Sigmab+\lambda\Ib_\rho)^{-1} & \mathbf{0}\\
	\mathbf{0} &\Ib_{n-\rho}
	\end{pmatrix}\Ub_f^\ts\Lb^{(j)}\nonumber\\
	=&~\begin{pmatrix}
	\Ub & \Ub_{\perp}
	\end{pmatrix}\begin{pmatrix}
	(\Sigmab^2+\lambda\Ib_\rho)(\Sigmab\Vb^\ts\Sb\Sb^\ts\Vb\Sigmab+\lambda\Ib_\rho)^{-1} & \mathbf{0}\\
	\mathbf{0} &\Ib_{n-\rho}
	\end{pmatrix}\begin{pmatrix}
	\Ub^\ts\\
	\Ub_{\perp}^\ts
	\end{pmatrix}\Lb^{(j)}\nonumber\\
	=&~\Ub_{\perp}\Ub_{\perp}^\ts\Lb^{(j)}+\Ub(\Sigmab^2+\lambda\Ib_\rho)(\Sigmab\Vb^\ts\Sb\Sb^\ts\Vb\Sigmab+\lambda\Ib_\rho)^{-1}\Ub^\ts\Lb^{(j)}\label{eq:fdarecur1}\\
	=&~\Ub_{\perp}\Ub_{\perp}^\ts\Lb^{(j)}+\Ub(\Sigmab^2+\lambda\Ib_\rho)\left(\Sigmab\Sigmab_\lambda^{-1}\left(\Sigmab_\lambda\Vb^\ts\Sb\Sb^\ts\Vb\Sigmab_\lambda\right)\Sigmab_\lambda^{-1}\Sigmab+\lambda\Ib_\rho\right)^{-1}\Ub^\ts\Lb^{(j)}\nonumber\\
	=&~\Ub_{\perp}\Ub_{\perp}^\ts\Lb^{(j)}+\Ub(\Sigmab^2+\lambda\Ib_\rho)\left(\Sigmab\Sigmab_\lambda^{-1}\left(\Sigmab_\lambda^2+\Eb\right)\Sigmab_\lambda^{-1}\Sigmab+\lambda\Sigmab\Sigmab_\lambda^{-1}\Sigmab_\lambda\Sigmab^{-2}\Sigmab_\lambda\Sigmab_\lambda^{-1}\Sigmab\right)^{-1}\Ub^\ts\Lb^{(j)}\nonumber\\
	=&~\Ub_{\perp}\Ub_{\perp}^\ts\Lb^{(j)}+\Ub(\Sigmab^2+\lambda\Ib_\rho)\left(\Sigmab\Sigmab_\lambda^{-1}\left(\Sigmab_\lambda^2+\Eb+\lambda\Sigmab_\lambda\Sigmab^{-2}\Sigmab_\lambda\right)\Sigmab_\lambda^{-1}\Sigmab\right)^{-1}\Ub^\ts\Lb^{(j)}\nonumber\\
	=&~\Ub_{\perp}\Ub_{\perp}^\ts\Lb^{(j)}+\Ub(\Sigmab^2+\lambda\Ib_\rho)\left(\Sigmab\Sigmab_\lambda^{-1}\left(\Ib_\rho+\Eb\right)\Sigmab_\lambda^{-1}\Sigmab\right)^{-1}\Ub^\ts\Lb^{(j)}.\label{eq:fda_recur2}
	\end{flalign}
	Here, eqn.~\eqref{eq:fda_recur2} holds because $\Sigmab_\lambda\Vb^\ts\Sb\Sb^\ts\Vb\Sigmab_\lambda=\Sigmab_\lambda^2+\Eb$ and
	the fact that $\Sigmab_\lambda^2+\lambda\Sigmab_\lambda\Sigmab^{-2}\Sigmab_\lambda\in\RR{n}{n}$ is a diagonal matrix whose $i$th diagonal element satisfies
	\begin{flalign*} \left(\Sigmab_\lambda^2+\lambda\Sigmab_\lambda\Sigmab^{-2}\Sigmab_\lambda\right)_{ii}=\frac{\sigma_i^2}{\sigma_i^2+\lambda}+\frac{\lambda}{\sigma_i^2+\lambda}=1,
	\end{flalign*}
	for any $i=1\ldots \rho$.
	Thus, we have $\left(\Sigmab_\lambda^2+\lambda\Sigmab_\lambda\Sigmab^{-2}\Sigmab_\lambda\right)=\Ib_\rho$.
	Since $\|\Eb\|_2<1$, Proposition~\ref{eq:matsum}  implies that $(\Ib_\rho+\Eb)^{-1}$ exists and $$(\Ib_\rho+\Eb)^{-1}=~\Ib_\rho+\sum_{\ell=1}^{\infty}(-1)^\ell\Eb^\ell=\Ib_\rho+\Qb, $$
	where $\Qb=\sum_{\ell=1}^{\infty}(-1)^\ell\Eb^\ell$.

	Thus, we rewrite eqn.~\eqref{eq:fda_recur2} as
	\begin{flalign}
	&~(\Ab\Ab^\ts+\lambda\Ib_n)(\Ab\Sb\Sb^\ts\Ab^\ts+\lambda\Ib_n)^{-1}\Lb^{(j)}\nonumber\\
	=&~\Ub_{\perp}\Ub_{\perp}^\ts\Lb^{(j)}+\Ub(\Sigmab^2+\lambda\Ib_\rho)\Sigmab^{-1}\Sigmab_\lambda\left(\Ib_\rho+\Eb\right)^{-1}\Sigmab_\lambda\Sigmab^{-1}\Ub^\ts\Lb^{(j)}\nonumber\\
	=&~\Ub_{\perp}\Ub_{\perp}^\ts\Lb^{(j)}+\Ub(\Sigmab^2+\lambda\Ib_\rho)\Sigmab^{-1}\Sigmab_\lambda\left(\Ib_\rho+\Qb\right)\Sigmab_\lambda\Sigmab^{-1}\Ub^\ts\Lb^{(j)}\nonumber\\
	=&~\Ub_{\perp}\Ub_{\perp}^\ts\Lb^{(j)}+\Ub(\Sigmab^2+\lambda\Ib_\rho)\Sigmab^{-1}\Sigmab_\lambda^2\Sigmab^{-1}\Ub^\ts\Lb^{(j)}+\Ub(\Sigmab^2+\lambda\Ib_\rho)\Sigmab^{-1}\Sigmab_\lambda\Qb\Sigmab_\lambda\Sigmab^{-1}\Ub^\ts\Lb^{(j)}\nonumber\\
	=&~\Ub_{\perp}\Ub_{\perp}^\ts\Lb^{(j)}+\Ub\Ub^\ts\Lb^{(j)}+\Ub(\Sigmab^2+\lambda\Ib_\rho)\Sigmab^{-1}\Sigmab_\lambda\Qb\Sigmab_\lambda\Sigmab^{-1}\Ub^\ts\Lb^{(j)}\label{eq:fda_rec12}\\
	=&~(\Ub\Ub^\ts+\Ub_{\perp}\Ub_{\perp}^\ts)\Lb^{(j)}+\Ub(\Sigmab^2+\lambda\Ib_\rho)\Sigmab^{-1}\Sigmab_\lambda\Qb\Sigmab_\lambda\Sigmab^{-1}\Ub^\ts\Lb^{(j)}\nonumber\\
	=&~\Ub_f\Ub_f^\ts\Lb^{(j)}+\Ub(\Sigmab^2+\lambda\Ib_\rho)\Sigmab^{-1}\Sigmab_\lambda\Qb\Sigmab_\lambda\Sigmab^{-1}\Ub^\ts\Lb^{(j)}.\label{eq:fda_recur3}
	\end{flalign}
	Eqn.~\eqref{eq:fda_rec12} holds as $(\Sigmab^2+\lambda\Ib_\rho)\Sigmab^{-1}\Sigmab_\lambda^2\Sigmab^{-1}=\Ib_\rho$. Further, using the fact that $\Ub_f\Ub_f^\ts=\Ib_n$, we rewrite eqn.~\eqref{eq:fda_recur3} as
	\begin{flalign}
	&(\Ab\Ab^\ts+\lambda\Ib_n)(\Ab\Sb\Sb^\ts\Ab^\ts+\lambda\Ib_n)^{-1}\Lb^{(j)} =\Lb^{(j)}+\Ub(\Sigmab^2+\lambda\Ib_\rho)\Sigmab^{-1}\Sigmab_\lambda\Qb\Sigmab_\lambda\Sigmab^{-1}\Ub^\ts\Lb^{(j)}.\label{eq:fda_recur10}
	\end{flalign}
	Thus, combining eqns.~\eqref{eq:fda_recur6} and \eqref{eq:fda_recur10}
	\begin{flalign}	\Lb^{(j+1)}=~-\Ub(\Sigmab^2+\lambda\Ib_\rho)\Sigmab^{-1}\Sigmab_\lambda\Qb\Sigmab_\lambda\Sigmab^{-1}\Ub^\ts\Lb^{(j)}.\label{eq:fda_recur19}
	\end{flalign}
	Finally, using eqn.~\eqref{eq:fda_recur19}
	\begin{flalign}
%	\nbr{\Sigmab_\lambda\Sigmab^{-1}\Ub^\ts\Lb^{(j+1)}}_2=&~\nbr{\Sigmab_\lambda\Sigmab^{-1}\Ub^\ts\Ub(\Sigmab^2+\lambda\Ib_\rho)\Sigmab^{-1}\Sigmab_\lambda\Qb\Sigmab_\lambda\Sigmab^{-1}\Ub^\ts\Lb^{(j)}}_2\nonumber\\
%	=&~\nbr{\Sigmab_\lambda\Sigmab^{-1}(\Sigmab^2+\lambda\Ib_\rho)\Sigmab^{-1}\Sigmab_\lambda\Qb\Sigmab_\lambda\Sigmab^{-1}\Ub^\ts\Lb^{(j)}}_2\nonumber\\
\nbr{\Sigmab_\lambda\Sigmab^{-1}\Ub^\ts\Lb^{(j+1)}}_2=&~\nbr{\Sigmab_\lambda\Sigmab^{-1}\Ub^\ts\Ub(\Sigmab^2+\lambda\Ib_\rho)\Sigmab^{-1}\Sigmab_\lambda\Qb\Sigmab_\lambda\Sigmab^{-1}\Ub^\ts\Lb^{(j)}}_2\nonumber\\
=&~\nbr{\Sigmab_\lambda\Sigmab^{-1}(\Sigmab^2+\lambda\Ib_\rho)\Sigmab^{-1}\Sigmab_\lambda\Qb\Sigmab_\lambda\Sigmab^{-1}\Ub^\ts\Lb^{(j)}}_2\nonumber\\
=&~\nbr{\Qb\Sigmab_\lambda\Sigmab^{-1}\Ub^\ts\Lb^{(j)}}_2
\le\nbr{\Qb}_2\nbr{\Sigmab_\lambda\Sigmab^{-1}\Ub^\ts\Lb^{(j)}}_2\nonumber\\
\le&~\ve\,\nbr{\Sigmab_\lambda\Sigmab^{-1}\Ub^\ts\Lb^{(j)}}_2\,.\nonumber
	\end{flalign}
	where the third equality holds as $\Sigmab_\lambda\Sigmab^{-1}(\Sigmab^2+\lambda\Ib_\rho)\Sigmab^{-1}\Sigmab_\lambda=\Ib_\rho$ and the last two steps follow from sub-multiplicativity and eqn.~\eqref{eq:272} respectively. This concludes the proof.
\end{proof}

\paragraph{Proof of Theorem~\ref{thm:iterative2}.}
Applying Lemma~\ref{lem:fda_recur} iteratively, we get
\begin{flalign}
\nbr{\Sigmab_\lambda\Sigmab^{-1}\Ub^\ts\Lb^{(t)}}_2\le
&~\ve\,\nbr{\Sigmab_\lambda\Sigmab^{-1}\Ub^\ts\Lb^{(t-1)}}_2 \le \ldots
\le \ve^{t-1}\nbr{\Sigmab_\lambda\Sigmab^{-1}\Ub^\ts\Lb^{(1)}}_2.
\label{eq:fda_recur71}
\end{flalign}

Now, from eqn~\eqref{eq:fda_recur71}, we apply sub-multiplicativity to obtain
\begin{flalign}
\nbr{\Sigmab_\lambda\Sigmab^{-1}\Ub^\ts\Lb^{(1)}}_2=&~\nbr{\Sigmab_\lambda\Sigmab^{-1}\Ub^\ts\Omegab}_2
\le \nbr{\Sigmab_\lambda\Sigmab^{-1}}_2\nbr{\Ub^\ts}_2\nbr{\Omegab}_2
= \max_{1\le i\le\rho}(\sigma_i^2+\lambda)^{-\frac{1}{2}}
\le \lambda^{-\frac{1}{2}},
\label{eq:th11_last}
\end{flalign}
where we used the facts that $\nbr{\Ub^\ts}_2=1$,
$\Omegab^\ts\Omegab=\Ib_c$, and $\nbr{\Omegab}_2=1$.

Finally, combining eqns.~\eqref{eq:fda_iter2},~\eqref{eq:fda_recur71} and~\eqref{eq:th11_last}, we conclude
\begin{flalign*}
\nbr{(\wb-\mb)^\ts(\widehat{\Gb} -\Gb)}_2\le\frac{\ve^t}{\sqrt{\lambda}}~\nbr{\Vb\Vb^\ts(\wb-\mb)}_2,
\end{flalign*}
which completes the proof.
\qed
%!TEX root = arxiv_LDA.tex

\section{Proof of Theorem~\ref{thm:iterative_rda}}

\begin{lemma}\label{lem:struct2}
	%%%%%%%%%%%%%%%%%%%%%%%%%%%%%%%%%%%%%%%%%%%%%%%%%%%%%%%%%%%%%%%%%%%%%%%%%%%%%%
	For $j=1\ldots t$, let $\Lb^{(j)}$ and $\widetilde{\Gb}^{(j)}$ be the intermediate matrices in Algorithm~\ref{algo:iterative_rda}, $\Gb^{(j)}$ be the matrix defined in eqn.~\eqref{eqn:pdits} and $\Rb$ be defined as in Lemma~\ref{lem:alt_xopt}. Further, let $\Sb\in\RR{d}{s}$ be the sketching matrix and define $\widehat{\Eb}=\Vb^\ts\Sb\Sb^\ts\Vb-\Ib_{\rho}.$ If eqn.~\eqref{eq:struct1} is satisfied, i.e., $\|\widehat{\Eb}\|_2\le\frac{\ve}{2}$, then for all $j=1 \ldots t$, we have
	\begin{flalign}
	\nbr{(\wb-\mb)^\ts(\widetilde{\Gb}^{(j)}-\Gb^{(j)})}_2\le~\ve\nbr{\Vb\Vb^\ts(\wb-\mb)}_2\, \nbr{\Rb^{-1}\Sigmab^{-1}\Ub^\ts\Lb^{(j)}}_2 \,,
	\end{flalign}
	where  $\Rb=\Ib_\rho+\lambda\Sigmab^{-2}$.
\end{lemma}

\begin{proof}
	Note that $\Sigmab_\lambda^2=\Rb^{-1}$. Applying Lemma~\ref{lem:alt_xopt}, we can express
	%\smash{$\widetilde{\Gb}^{(j)}$} in terms of
	$\Gb^{(j)}$ as
	\begin{flalign}
	\Gb^{(j)}=\Vb\Rb^{-1}\Sigmab^{-1}\Ub^\ts\Lb^{(j)}.\label{eq:lev_rda_alt}
	\end{flalign}

	Next, rewriting eqn.~\eqref{eq:fda14} gives
	\begin{flalign}
	\widetilde{\Gb}^{(j)}=&~\Vb\Sigmab(\Sigmab\Vb^\ts\Sb\Sb^\ts\Vb\Sigmab+\lambda\Ib_\rho)^{-1}\Ub^\ts\Lb^{(j)}\\
	=&~\Vb\Sigmab(\Sigmab(\Ib_\rho+\widehat{\Eb})\Sigmab+\lambda\Ib_\rho)^{-1}\Ub^\ts\Lb^{(j)}
	= \Vb\Sigmab\Sigmab^{-1}(\Ib_\rho+\widehat{\Eb}+\lambda\Sigmab^{-2})^{-1}\Sigmab^{-1}\Ub^\ts\Lb^{(j)}\nonumber\\
	=&~\Vb(\Rb+\widehat{\Eb})^{-1}\Sigmab^{-1}\Ub^\ts\Lb^{(j)}
	= \Vb(\Rb(\Ib_\rho+\Rb^{-1}\widehat{\Eb}))^{-1}\Sigmab^{-1}\Ub^\ts\Lb^{(j)}\,.\label{eq:fda04}
	\end{flalign}

	Further, notice that
	\begin{flalign}
	\nbr{\Rb^{-1}\widehat{\Eb}}_2\le\nbr{\Rb^{-1}}_2\nbr{\widehat{\Eb}}_2\le\nbr{\Rb^{-1}}_2\cdot \frac{\ve}{2}=\left(\frac{\sigma_1^2}{\sigma_1^2+\lambda}\right)\frac{\ve}{2}\le\frac{\ve}{2}< 1.\label{eq:fda07}
	\end{flalign}
	Now, Proposition~\ref{eq:matsum} implies that $(\Ib_\rho+\Rb^{-1}\widehat{\Eb})^{-1}$ exists.
	Let $\widehat{\Qb}=\sum_{\ell=1}^{\infty}(-1)^\ell(\Rb^{-1}\widehat{\Eb})^\ell$, we have
	\begin{flalign*}
	(\Ib_\rho+\Rb^{-1}\widehat{\Eb})^{-1}=\Ib_\rho+\sum_{\ell=1}^{\infty}(-1)^{\ell}(\Rb^{-1}\widehat{\Eb})^{\ell}=\Ib_\rho+\widehat{\Qb}.
	\end{flalign*}
	Thus, we can rewrite eqn.~\eqref{eq:fda04} as
	\begin{flalign}
	\widetilde{\Gb}^{(j)}=&~\Vb(\Ib_\rho+\widehat{\Qb})\Rb^{-1}\Sigmab^{-1}\Ub^\ts\Lb^{(j)}\nonumber\\
	=&~\Vb\Rb^{-1}\Sigmab^{-1}\Ub^\ts\Lb^{(j)}+\Vb\widehat{\Qb}\Rb^{-1}\Sigmab^{-1}\Ub^\ts\Lb^{(j)}\nonumber\\
	=&~\Gb^{(j)}+\Vb\widehat{\Qb}\Rb^{-1}\Sigmab^{-1}\Ub^\ts\Lb^{(j)},\label{eq:fda05}
	\end{flalign}
	where eqn.~\eqref{eq:fda05} follows eqn.~\eqref{eq:lev_rda_alt}. Further, using eqn.~\eqref{eq:fda07}, we have
	\begin{flalign}
	\|\widehat{\Qb}\|_2=&~\nbr{\sum_{\ell=1}^{\infty}(-1)^{\ell}(\Rb^{-1}\widehat{\Eb})^{\ell}}_2\le\sum_{\ell=1}^{\infty}\nbr{(\Rb^{-1}\widehat{\Eb})^{\ell}}_2\nonumber\\
	\le&~\sum_{\ell=1}^{\infty}\nbr{\Rb^{-1}\widehat{\Eb}}_2^{\ell}\le~\sum_{\ell=1}^{\infty}(\frac{\ve}{2})^{\ell}=~\frac{\ve/2}{1-\ve/2}\le~\ve,\label{eq:fda06}
	\end{flalign}
	where we used the triangle inequality, sub-multiplicativity of the spectral norm, and the fact that $\ve \le 1$.
	Next, we combine eqns.~\eqref{eq:fda05} and~\eqref{eq:fda06} to get
	\begin{flalign} \nbr{(\wb-\mb)^\ts(\widetilde{\Gb}^{(j)}-\Gb^{(j)})}_2=&~\nbr{(\wb-\mb)^\ts\Vb\widehat{\Qb}\Rb^{-1}\Sigmab^{-1}\Ub^\ts\Lb^{(j)}}_2\nonumber\\
	\le&~\nbr{(\wb-\mb)^\ts\Vb}_2\nbr{\widehat{\Qb}}_2\nbr{\Rb^{-1}\Sigmab^{-1}\Ub^\ts\Lb^{(j)}}_2\nonumber\\
	\le&~\ve~\nbr{(\wb-\mb)^\ts\Vb}_2\nbr{\Rb^{-1}\Sigmab^{-1}\Ub^\ts\Lb^{(j)}}_2\nonumber\\
	=&~\ve~\nbr{(\wb-\mb)^\ts\Vb\Vb^\ts}_2\nbr{\Rb^{-1}\Sigmab^{-1}\Ub^\ts\Lb^{(j)}}_2\nonumber\\
	=&~\ve~\nbr{\Vb\Vb^\ts(\wb-\mb)}_2\nbr{\Rb^{-1}\Sigmab^{-1}\Ub^\ts\Lb^{(j)}}_2,\label{eq:fda015}
	\end{flalign}
	where the first inequality follows from sub-multiplicativity  and the second last equality holds due to the unitary invariance of the  spectral norm. This concludes the proof.
\end{proof}

\begin{remark}
	Repeated application of Lemma~\ref{lem:induction} and Lemma~\ref{lem:struct2} yields:
	\begin{flalign}
	\nbr{(\wb-\mb)^\ts(\widehat{\Gb} -\Gb)}_2
	=&~\nbr{(\wb-\mb)^\ts\big(\sum_{j=1}^{t}\widetilde{\Gb}^{(j)}-\Gb\big)}_2 \nonumber\\
	=&~\nbr{(\wb-\mb)^\ts(\widetilde{\Gb}^{(t)}-\big(\Gb-\sum_{j=1}^{t-1}\widetilde{\Gb}^{(j)}\big))}_2\nonumber\\
	=&~\nbr{(\wb-\mb)^\ts(\widetilde{\Gb}^{(t)}-\Gb^{(t)})}_2\nonumber\\
	\le&~\ve~\nbr{\Vb\Vb^\ts(\wb-\mb)}_2 \, \nbr{\Rb^{-1}\Sigmab^{-1}\Ub^\ts\Lb^{(t)}}_2.\label{eq:iter2}
	\end{flalign}
\end{remark}

The next bound provides a critical inequality that can be used recursively in order to establish Theorem~\ref{thm:iterative_rda}.
\begin{lemma}\label{lem:recur}
	Let $\Lb^{(j)}$, $j=1 \ldots t$, be the matrices of Algorithm~\ref{algo:iterative_rda} and $\Rb$ is as defined in Lemma~\ref{lem:alt_xopt}. For any $j=1 \ldots t-1$, define $\widehat{\Eb}=\Vb^\ts\Sb\Sb^\ts\Vb-\Ib_{\rho}.$ If eqn.~\eqref{eq:struct1} is satisfied \ie $\|\widehat{\Eb}\|_2\le\frac{\ve}{2}$, then
	\begin{flalign}
	&\nbr{\Rb^{-1}\Sigmab^{-1}\Ub^\ts\Lb^{(j+1)}}_2\le~\ve\,\nbr{\Rb^{-1}\Sigmab^{-1}\Ub^\ts\Lb^{(j)}}_2.\label{eq:recur}
	\end{flalign}
\end{lemma}

\begin{proof} From Algorithm~\ref{algo:iterative_rda}, we have for $j=1 \ldots t-1$
	\begin{flalign}
	\Lb^{(j+1)}=&~\Lb^{(j)}-\lambda\Yb^{(j)}-\Ab\widetilde{\Gb}^{(j)}\nonumber\\
	=&~\Lb^{(j)}-(\Ab\Ab^\ts+\lambda\Ib_n)(\Ab\Sb\Sb^\ts\Ab^\ts+\lambda\Ib_n)^{-1}\Lb^{(j)}.
	\label{eq:recur6}
	\end{flalign}

	Now, rewriting eqn.~\eqref{eq:fdarecur1}, we have
	\begin{flalign}
	&~(\Ab\Ab^\ts+\lambda\Ib_n)(\Ab\Sb\Sb^\ts\Ab^\ts+\lambda\Ib_n)^{-1}\Lb^{(j)}\nonumber\\
	=&~\Ub_{\perp}\Ub_{\perp}^\ts\Lb^{(j)}+\Ub(\Sigmab^2+\lambda\Ib_\rho)(\Sigmab\Vb^\ts\Sb\Sb^\ts\Vb\Sigmab+\lambda\Ib_\rho)^{-1}\Ub^\ts\Lb^{(j)}\nonumber\\
	=&~\Ub_{\perp}\Ub_{\perp}^\ts\Lb^{(j)}+\Ub(\Sigmab^2+\lambda\Ib_\rho)(\Sigmab(\Ib_\rho+\widehat{\Eb})\Sigmab+\lambda\Ib_\rho)^{-1}\Ub^\ts\Lb^{(j)}\nonumber\\
	=&~\Ub_{\perp}\Ub_{\perp}^\ts\Lb^{(j)}+\Ub(\Sigmab^2+\lambda\Ib_\rho)\Sigmab^{-1}(\Ib_\rho+\widehat{\Eb}+\lambda\Sigmab^{-2})^{-1}\Sigmab^{-1}\Ub^\ts\Lb^{(j)}.\label{eq:recur2}
	\end{flalign}
	Here, eqn.~\eqref{eq:recur2} holds because $(\Ib_\rho+\widehat{\Eb}+\lambda\Sigmab^{-2})$ is invertible since it is a positive definite matrix. In addition, using the fact that $\Rb=(\Ib_\rho+\lambda\Sigmab^{-2})$, we rewrite eqn.~\eqref{eq:recur2} as
	\begin{flalign}
	&~(\Ab\Ab^\ts+\lambda\Ib_n)(\Ab\Sb\Sb^\ts\Ab^\ts+\lambda\Ib_n)^{-1}\Lb^{(j)}\nonumber\\
	=&~\Ub_{\perp}\Ub_{\perp}^\ts\Lb^{(j)}+\Ub(\Sigmab^2+\lambda\Ib_\rho)\Sigmab^{-1}(\Rb+\widehat{\Eb})^{-1}\Sigmab^{-1}\Ub^\ts\Lb^{(j)}\nonumber\\
	=&~\Ub_{\perp}\Ub_{\perp}^\ts\Lb^{(j)}+\Ub(\Sigmab^2+\lambda\Ib_\rho)\Sigmab^{-1}\left(\Rb(\Ib_\rho+\Rb^{-1}\widehat{\Eb})\right)^{-1}\Sigmab^{-1}\Ub^\ts\Lb^{(j)}\nonumber\\
	=&~\Ub_{\perp}\Ub_{\perp}^\ts\Lb^{(j)}+\Ub(\Sigmab^2+\lambda\Ib_\rho)\Sigmab^{-1}(\Ib_\rho+\Rb^{-1}\widehat{\Eb})^{-1}\Rb^{-1}\Sigmab^{-1}\Ub^\ts\Lb^{(j)}\nonumber\\
	=&~\Ub_{\perp}\Ub_{\perp}^\ts\Lb^{(j)}+\Ub(\Sigmab^2+\lambda\Ib_\rho)\Sigmab^{-1}(\Ib_\rho+\widehat{\Qb})\Rb^{-1}\Sigmab^{-1}\Ub^\ts\Lb^{(j)}\nonumber\\
	=&~\Ub_{\perp}\Ub_{\perp}^\ts\Lb^{(j)}+\Ub(\Sigmab^2+\lambda\Ib_\rho)\Sigmab^{-1}\Rb^{-1}\Sigmab^{-1}\Ub^\ts\Lb^{(j)}+\Ub(\Sigmab^2+\lambda\Ib_\rho)\Sigmab^{-1}\widehat{\Qb}\Rb^{-1}\Sigmab^{-1}\Ub^\ts\Lb^{(j)}\nonumber\\
	=&~(\Ub\Ub^\ts+\Ub_{\perp}\Ub_{\perp}^\ts)\Lb^{(j)}+\Ub(\Sigmab^2+\lambda\Ib_\rho)\Sigmab^{-1}\widehat{\Qb}\Rb^{-1}\Sigmab^{-1}\Ub^\ts\Lb^{(j)}\nonumber\\
	=&~\Ub_f\Ub_f^\ts\Lb^{(j)}+\Ub(\Sigmab^2+\lambda\Ib_\rho)\Sigmab^{-1}\widehat{\Qb}\Rb^{-1}\Sigmab^{-1}\Ub^\ts\Lb^{(j)}.\label{eq:recur3}
	\end{flalign}
	The second and third equalities follow from Proposition~\ref{eq:matsum} (using eqn.~\eqref{eq:fda07}) and the fact that $\Rb^{-1}$ exists. Further, $\widehat{\Qb}$ is as defined as in Lemma~\ref{lem:struct2}.  Moreover, the second last equality holds as $(\Sigmab^2+\lambda\Ib_\rho)\Sigmab^{-1}\Rb^{-1}\Sigmab^{-1}=\Ib_\rho$. Now, using the fact that $\Ub_f\Ub_f^\ts=\Ib_n$, we rewrite eqn.~\eqref{eq:recur3} as
	\begin{flalign}
	&(\Ab\Ab^\ts+\lambda\Ib_n)(\Ab\Sb\Sb^\ts\Ab^\ts+\lambda\Ib_n)^{-1}\Lb^{(j)}\nonumber\\
	&~~~~~~~~~~~~~~~~~~~~~~~~~~=\Lb^{(j)}+\Ub(\Sigmab^2+\lambda\Ib_\rho)\Sigmab^{-1}\widehat{\Qb}\Rb^{-1}\Sigmab^{-1}\Ub^\ts\Lb^{(j)}.\label{eq:recur10}
	\end{flalign}
	Thus, combining, eqns.~\eqref{eq:recur6} and \eqref{eq:recur10}, we have
	\begin{flalign}
	\Lb^{(j+1)}=~-\Ub(\Sigmab^2+\lambda\Ib_\rho)\Sigmab^{-1}\widehat{\Qb}\Rb^{-1}\Sigmab^{-1}\Ub^\ts\Lb^{(j)}.\label{eq:recur19}
	\end{flalign}

	Finally, from eqn.~\eqref{eq:recur19}, we obtain
	\begin{flalign}
	\nbr{\Rb^{-1}\Sigmab^{-1}\Ub^\ts\Lb^{(j+1)}}_2=&~\nbr{\Rb^{-1}\Sigmab^{-1}\Ub^\ts\Ub(\Sigmab^2+\lambda\Ib_\rho)\Sigmab^{-1}\widehat{\Qb}\Rb^{-1}\Sigmab^{-1}\Ub^\ts\Lb^{(j)}}_2\nonumber\\
	=&~\|\Rb^{-1}\Sigmab^{-1}(\Sigmab^2+\lambda\Ib_\rho)\Sigmab^{-1}\widehat{\Qb}\Rb^{-1}\Sigmab^{-1}\Ub^\ts\Lb^{(j)}\|_2\nonumber\\
	=&~\|\widehat{\Qb}\Rb^{-1}\Sigmab^{-1}\Ub^\ts\Lb^{(j)}\|_2\le\nbr{\widehat{\Qb}}_2\nbr{\Rb^{-1}\Sigmab^{-1}\Ub^\ts\Lb^{(j)}}_2\nonumber\\
	\le&~\ve\nbr{\Rb^{-1}\Sigmab^{-1}\Ub^\ts\Lb^{(j)}}_2,
	\end{flalign}
	where the third equality holds as $\Rb^{-1}\Sigmab^{-1}(\Sigmab^2+\lambda\Ib_\rho)\Sigmab^{-1}=\Ib_\rho$ and the last two steps follow from sub-multiplicativity and eqn.~\eqref{eq:fda06} respectively. This concludes the proof.
\end{proof}

\paragraph{Proof of Theorem~\ref{thm:iterative_rda}.}
Applying Lemma~\ref{lem:recur} iteratively, we have
\begin{flalign}
\nbr{\Rb^{-1}\Sigmab^{-1}\Ub^\ts\Lb^{(t)}}_2\le&~\ve\nbr{\Rb^{-1}\Sigmab^{-1}\Ub^\ts\Lb^{(t-1)}}_2 \le \ldots \le \ve^{t-1}\nbr{\Rb^{-1}\Sigmab^{-1}\Ub^\ts\Lb^{(1)}}_2.
\label{eq:recur71}
\end{flalign}

Now, from eqn~\eqref{eq:recur71} and noticing that $\Lb^{(1)} = \Omegab$ by definition, we have
\begin{flalign}
\nbr{\Rb^{-1}\Sigmab^{-1}\Ub^\ts\Lb^{(1)}}_2
% =&~\nbr{\Rb^{-1}\Sigmab^{-1}\Ub^\ts\Omegab}_2
\le \nbr{\Rb^{-1}\Sigmab^{-1}}_2\nbr{\Ub^\ts}_2\nbr{\Omegab}_2
= \max_{1\le i\le\rho}\left\{\frac{\sigma_i}{\sigma_i^2+\lambda}\right\}\le \frac{1}{2\sqrt{\lambda}},
\label{eq:th5_last}
\end{flalign}
where we used sub-multiplicativity and the facts that $\nbr{\Ub^\ts}_2=1$, $ \Omegab^\ts\Omegab=\Ib_c$, and
$\nbr{\Omegab}_2=1$.
The last step in eqn.~\eqref{eq:th5_last} holds since for all $i=1 \ldots \rho$,
\begin{flalign}
(\sigma_i-\sqrt{\lambda})^2\ge 0 \quad \Rightarrow \quad \sigma_i^2+\lambda\ge 2\sigma_i\sqrt{\lambda} \quad \Rightarrow \quad \frac{\sigma_i}{\sigma_i^2+\lambda}\le \frac{1}{2\sqrt{\lambda}} .
\end{flalign}
Finally, combining eqns.~\eqref{eq:iter2},~\eqref{eq:recur71} and~\eqref{eq:th5_last}, we obtain
\begin{flalign*}
\nbr{(\wb-\mb)^\ts(\widehat{\Gb} -\Gb)}_2\le\frac{\ve^t}{2\sqrt{\lambda}}~\nbr{\Vb\Vb^\ts(\wb-\mb)}_2,
\end{flalign*}
which concludes the proof.
\qed
%!TEX root = arxiv_LDA.tex

%%
\section{Sampling-based approaches}\label{sxn:appendix:sampling}

We now discuss how to satisfy the conditions of eqns.~(\ref{eq:struct2}) or~(\ref{eq:struct1}) by \textit{sampling}, \ie, selecting a small number of features.
%Towards that end, consider Algorithm~\ref{algo:1} for the construction of the sampling-and-rescaling matrix %$\Sb$.
%
%We also derive Theorem~\ref{thm:matmul} (see Section~\ref{sxn:proof_sampling} for the proof), which may be of independent interest and is a strengthening of Theorem~4.2 of~\cite{Holod15}.

%\footnote{We do note that Theorem~\ref{thm:matmul} is implicit in~\cite{Co17}.}

%\begin{minipage}{.5\linewidth}
\begin{algorithm}[H]
	\caption{Sampling-and-rescaling matrix}\label{algo:1}%
	\begin{algorithmic}
		\State \textbf{Input:} Sampling probabilities $p_i$, $i=1,\dots, d$; \\
			\qquad \quad number of sampled columns $s \ll d$;
		\State $\Sb \gets \zero_{d \times s}$;

		\For{$t=1$ \textbf{to} $s$}
		\State Pick $i_t\in\left\{1,\ldots ,d\right\}$ with $\PP(i_t=i)=p_i$;
		\State $\Sb_{i_tt}=1/\sqrt{s\,p_{i_t}}$;
		\EndFor

		\State \textbf{Output:} Return $\Sb$;
	\end{algorithmic}
\end{algorithm}
%\end{minipage}
%
\hspace{.01\linewidth}
Finally, the next result appeared in~\cite{CYD18} as Theorem~3 and is a strengthening of Theorem~4.2 of \cite{Holod15}, since the sampling complexity $s$ is improved to depend only on $\nbr{\Zb}_F^2$ instead of the stable rank of $\Zb$ when $\|\Zb\|_2\le 1$. We also note that Lemma~\ref{lem:matmul} is implicit in \cite{Co17}\,.
%
%\begin{minipage}{.47\linewidth}
\begin{lemma}\label{lem:matmul}
	Let $\Zb \in \mathbb{R}^{d \times n}$ with $\nbr{\Zb}_2\leq 1$ and let $\Sb$ be constructed by Algorithm~\ref{algo:1} with  %$p_i=\nbr{\Zb_{i*}}_2^2/\nbr{\Zb}_F^2$ for  $i=1,\dots, d$. Let $\delta$ be a failure probability and let $\ve \in (0,1]$. If the number of %sampled columns $s$ satisfies
	\[
	s\ge \frac{8\nbr{\Zb}_F^2}{3\,\ve^2}\ln\left(\frac{4~(1+\nbr{\Zb}_F^2)}{\delta}\right),
	\]
	then, with probability at least  $1-\delta$,
	\begin{flalign*}
	\nbr{\Zb^\ts\Sb\Sb^\ts\Zb-\Zb^\ts\Zb}_2\le\ve.
	\end{flalign*}
\end{lemma}
%\end{minipage}

\par

Applying Lemma~\ref{lem:matmul} with $\Zb=\Vb\Sigmab_{\lambda}$, we can satisfy the condition of eqn.~(\ref{eq:struct2}) using the sampling probabilities $p_i = \nbr{(\Vb\Sigmab_{\lambda})_{i*}}_2^2/d_{\lambda}$ (recall that $\nbr{\Vb\Sigmab_{\lambda}}_F^2=d_{\lambda}$ and $\nbr{\Vb\Sigmab_{\lambda}}_2\leq 1$). It is easy to see that these probabilities are exactly proportional to the column ridge leverage scores of the design matrix $\Ab$. Setting $s = \Ocal(\ve^{-2} d_{\lambda} \ln d_{\lambda})$ suffices to satisfy the condition of eqn.~(\ref{eq:struct2}). We note that approximate ridge leverage scores also suffice and that their computation can be done efficiently without computing $\Vb$~\cite{Co17}. Finally, applying Lemma~\ref{lem:matmul} with $\Zb=\Vb$ we can satisfy the condition of eqn.~(\ref{eq:struct1}) by simply using the sampling probabilities $p_i = \nbr{\Vb_{i*}}_2^2/\rho$ (recall that $\nbr{\Vb}_F^2=\rho$ and $\nbr{\Vb}_2=1$), which correspond to the column leverage scores of the design matrix $\Ab$. Setting $s = \Ocal(\ve^{-2}\rho\ln\rho)$ suffices to satisfy the condition of eqn.~(\ref{eq:struct1}). We note that approximate leverage scores also suffice and that their computation can be done efficiently without computing $\Vb$~\cite{DM12}.

%\input{Proof_of_sampling}
%!TEX root = arxiv_LDA.tex

\section{Additional experimental results}
\label{sxn:appendix-expts}

As noted in Section~\ref{sxn:conclusions},
we conjecture that using different sampling matrices in each iteration of Algorithm~\ref{algo:iterative_rda} (\ie, introducing new ``randomness'' in each iteration) could lead to improved bounds for our main theorems.
We evaluate this conjecture empirically by comparing the performance of
Algorithm~\ref{algo:iterative_rda} using either a single sketching matrix $\Sb$
(the setup in the main paper) or sampling (independently) a new sketching matrix at every iteration $j$.

\begin{figure*}[htbp]
\centering
\subfloat{
    \includegraphics[width=0.19\columnwidth,keepaspectratio]{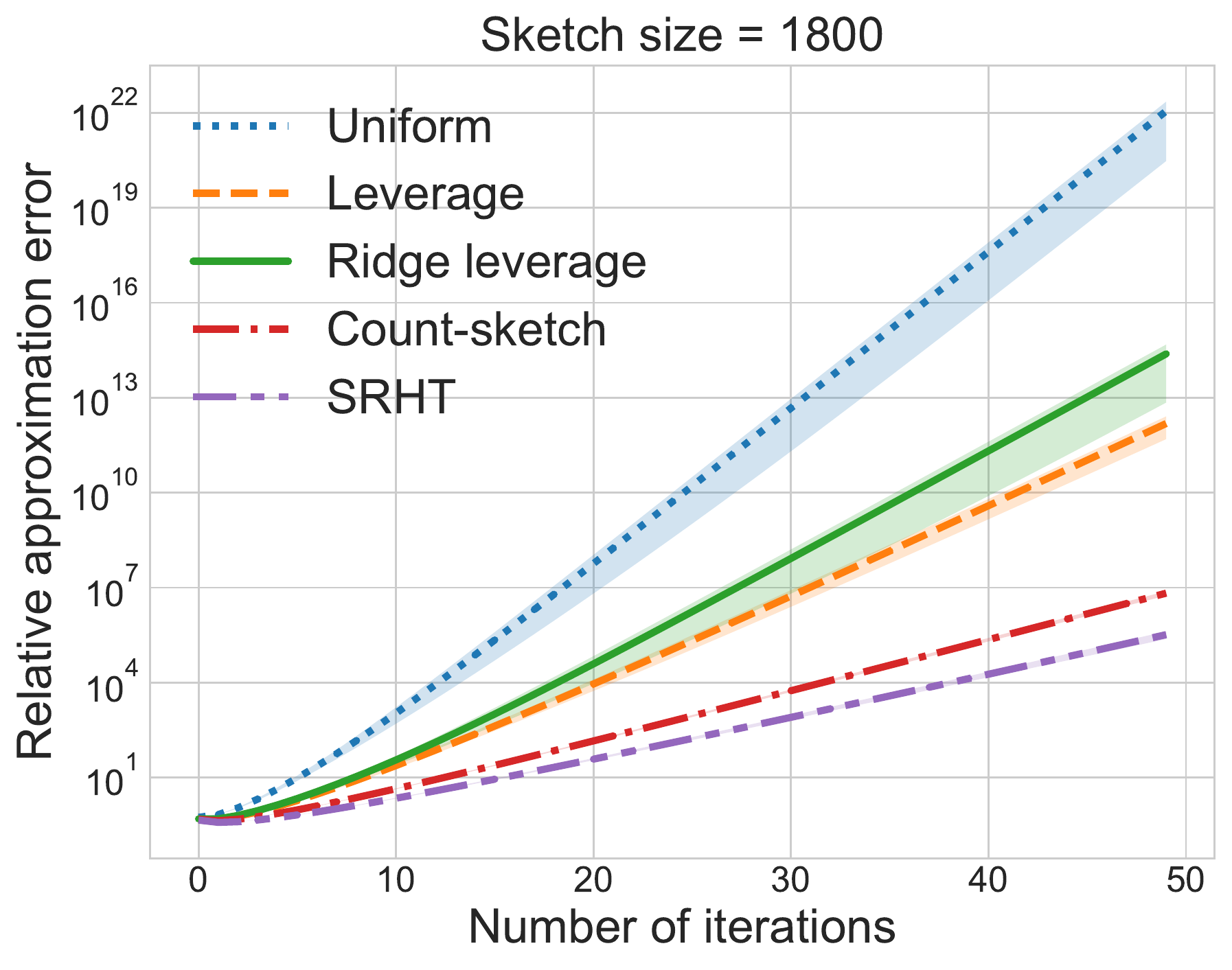}}
\subfloat{
    \includegraphics[width=0.19\columnwidth,keepaspectratio]{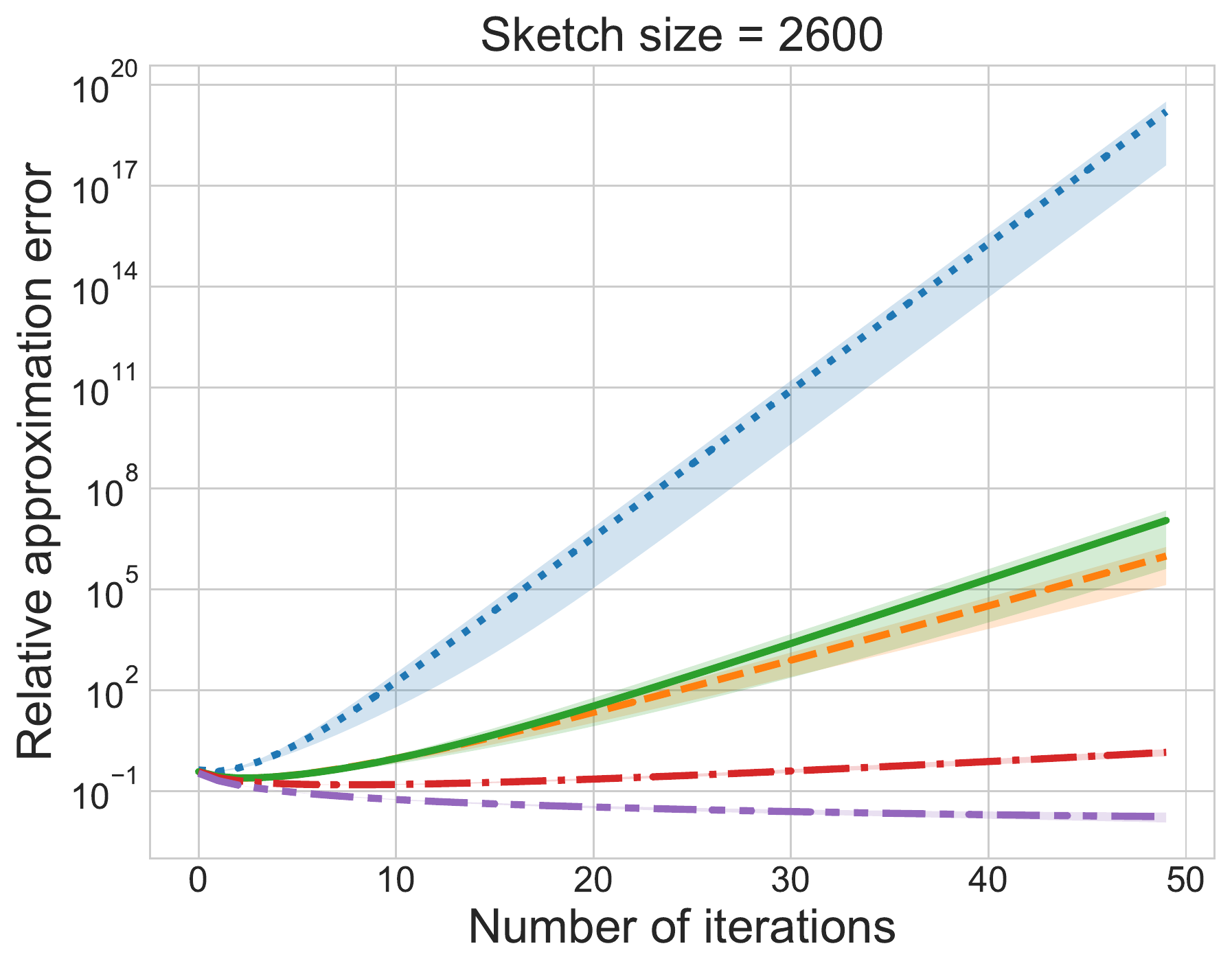}}
\subfloat{
    \includegraphics[width=0.19\columnwidth,keepaspectratio]{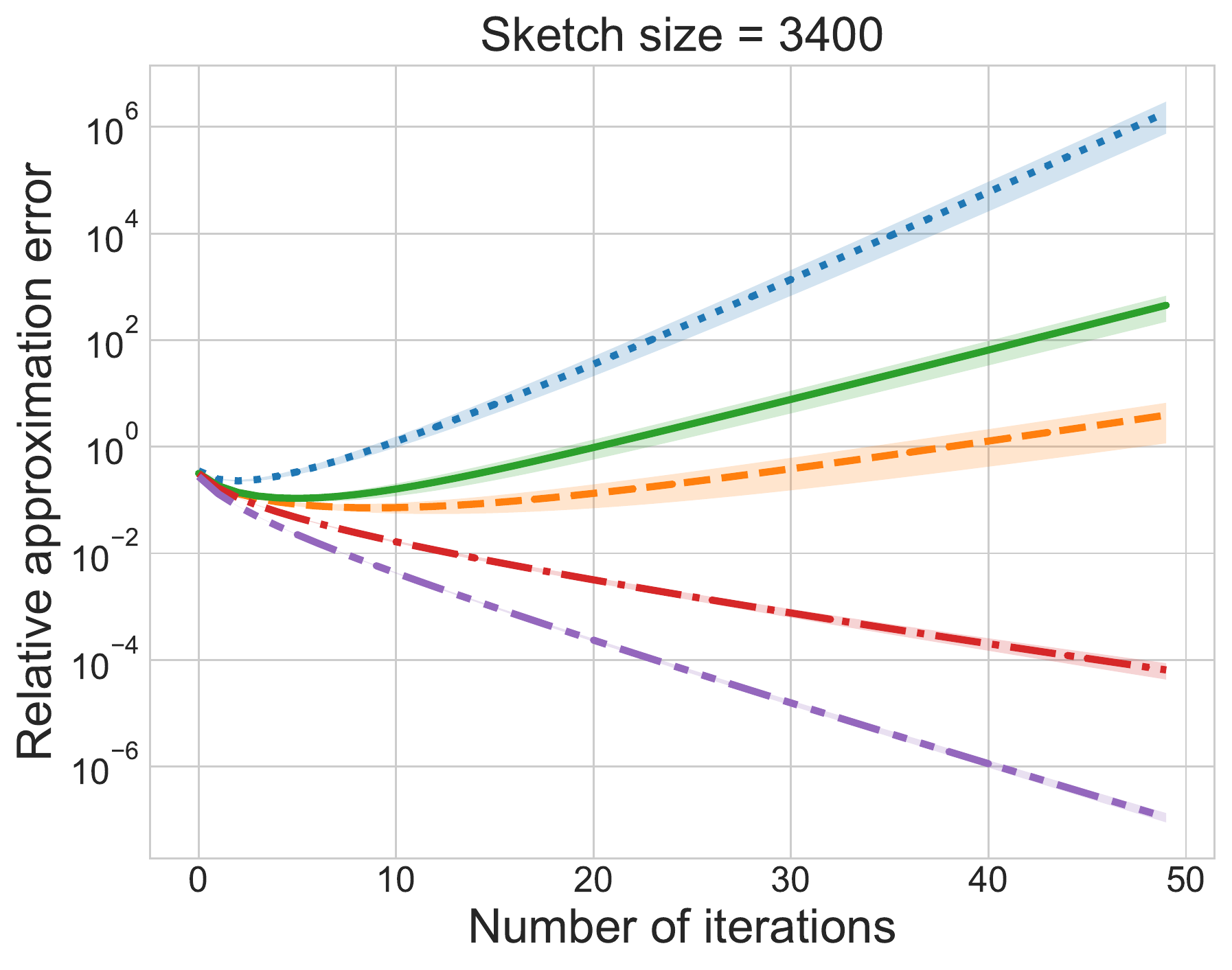}}
\subfloat{
    \includegraphics[width=0.19\columnwidth,keepaspectratio]{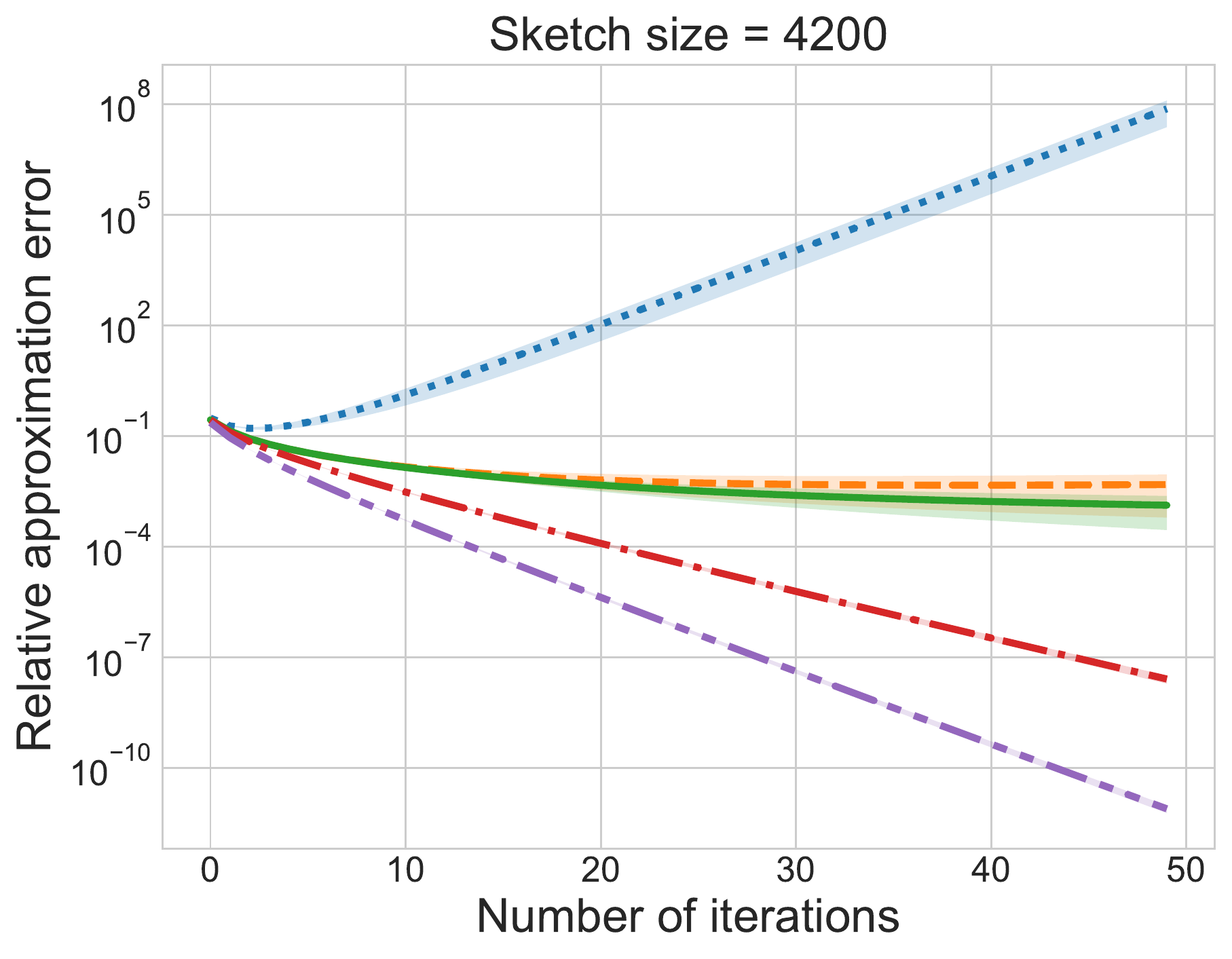}}
\subfloat{
    \includegraphics[width=0.19\columnwidth,keepaspectratio]{single-sketch/ORL-iter-err-rel-lmbd10-ncol5}}
\vspace{-8pt}
\addtocounter{subfigure}{-5}
\subfloat[$s = 1800$]{
    \includegraphics[width=0.19\columnwidth,keepaspectratio]{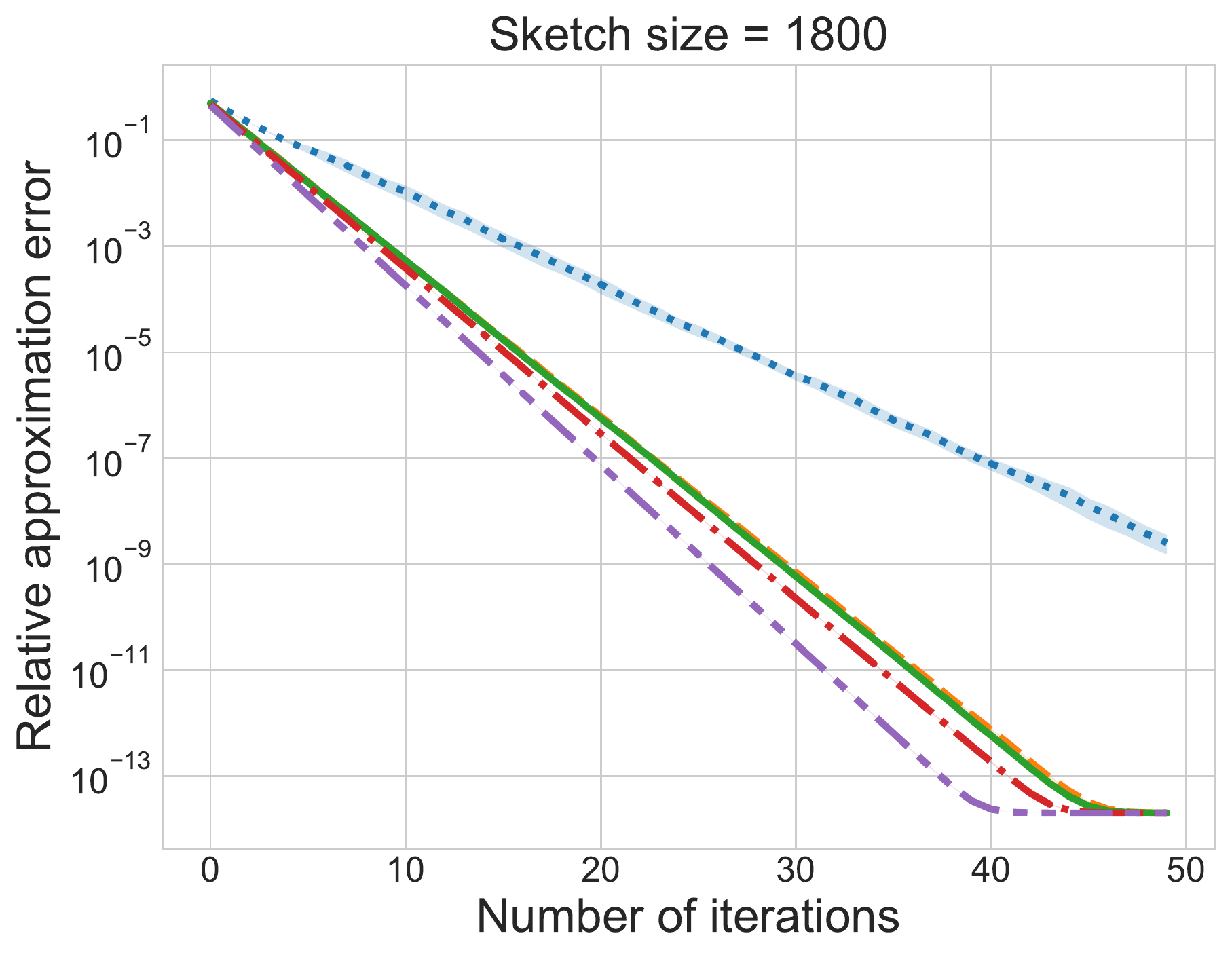}}
\subfloat[$s = 2600$]{
    \includegraphics[width=0.19\columnwidth,keepaspectratio]{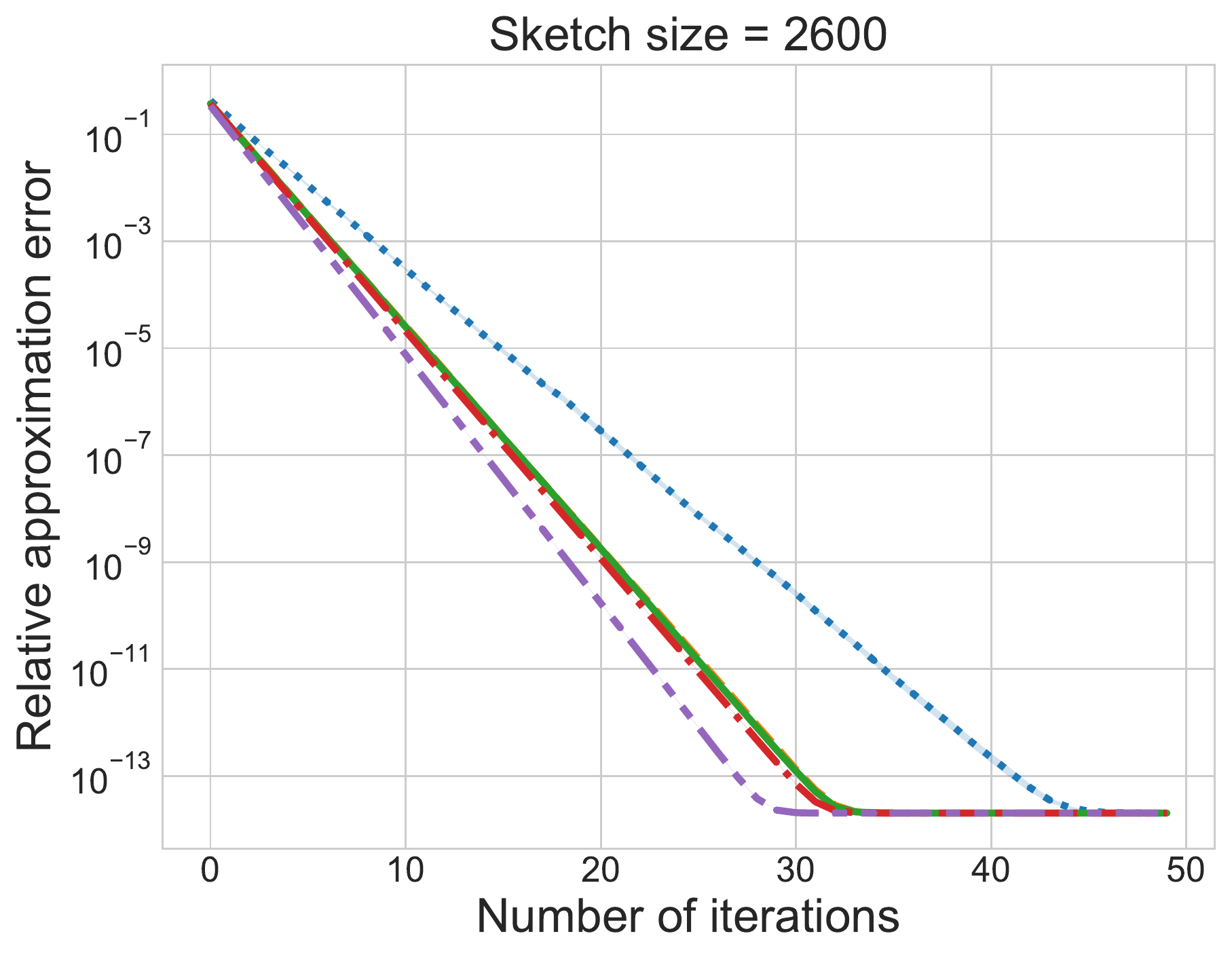}}
\subfloat[$s = 3400$]{
    \includegraphics[width=0.19\columnwidth,keepaspectratio]{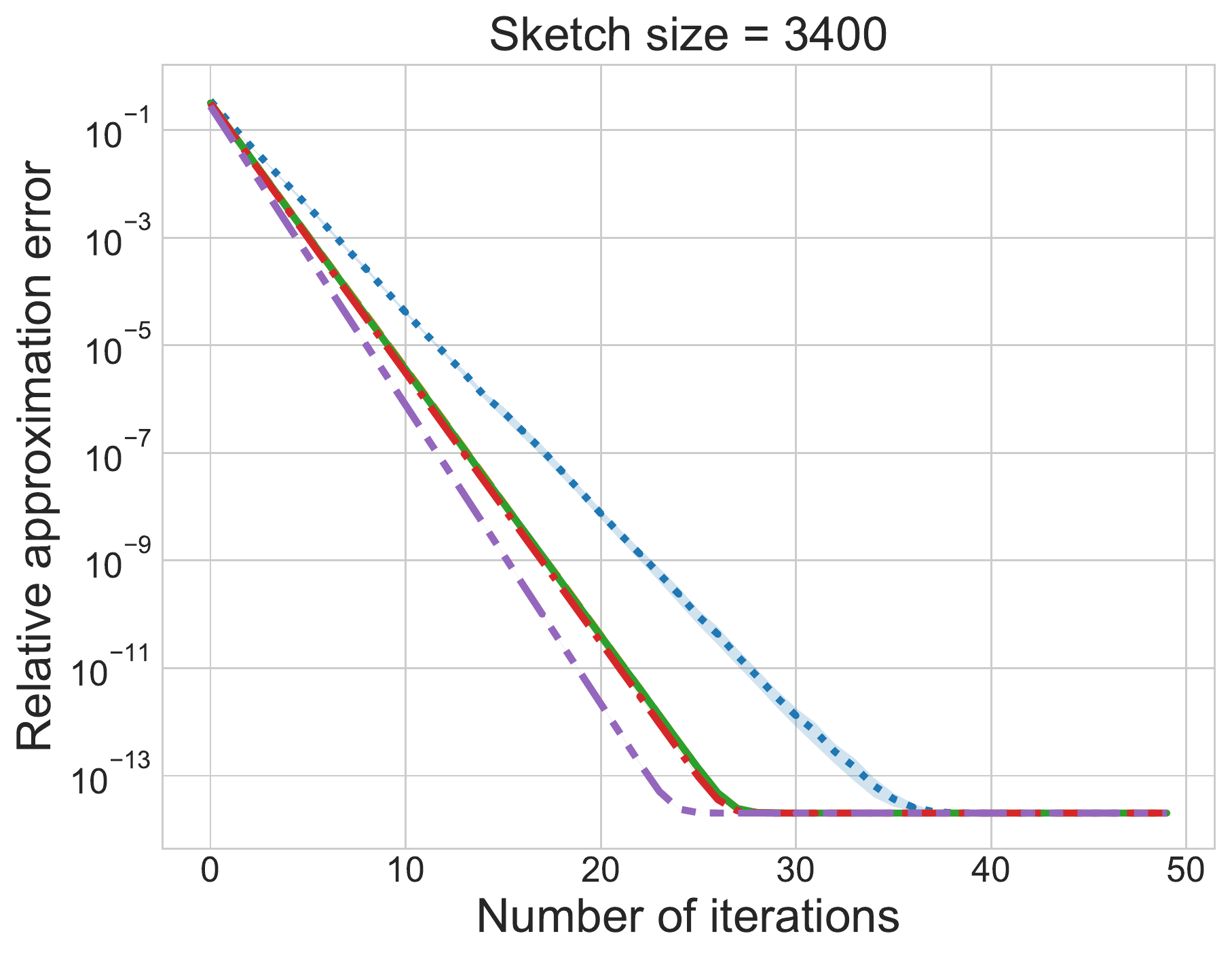}}
\subfloat[$s = 4200$]{
    \includegraphics[width=0.19\columnwidth,keepaspectratio]{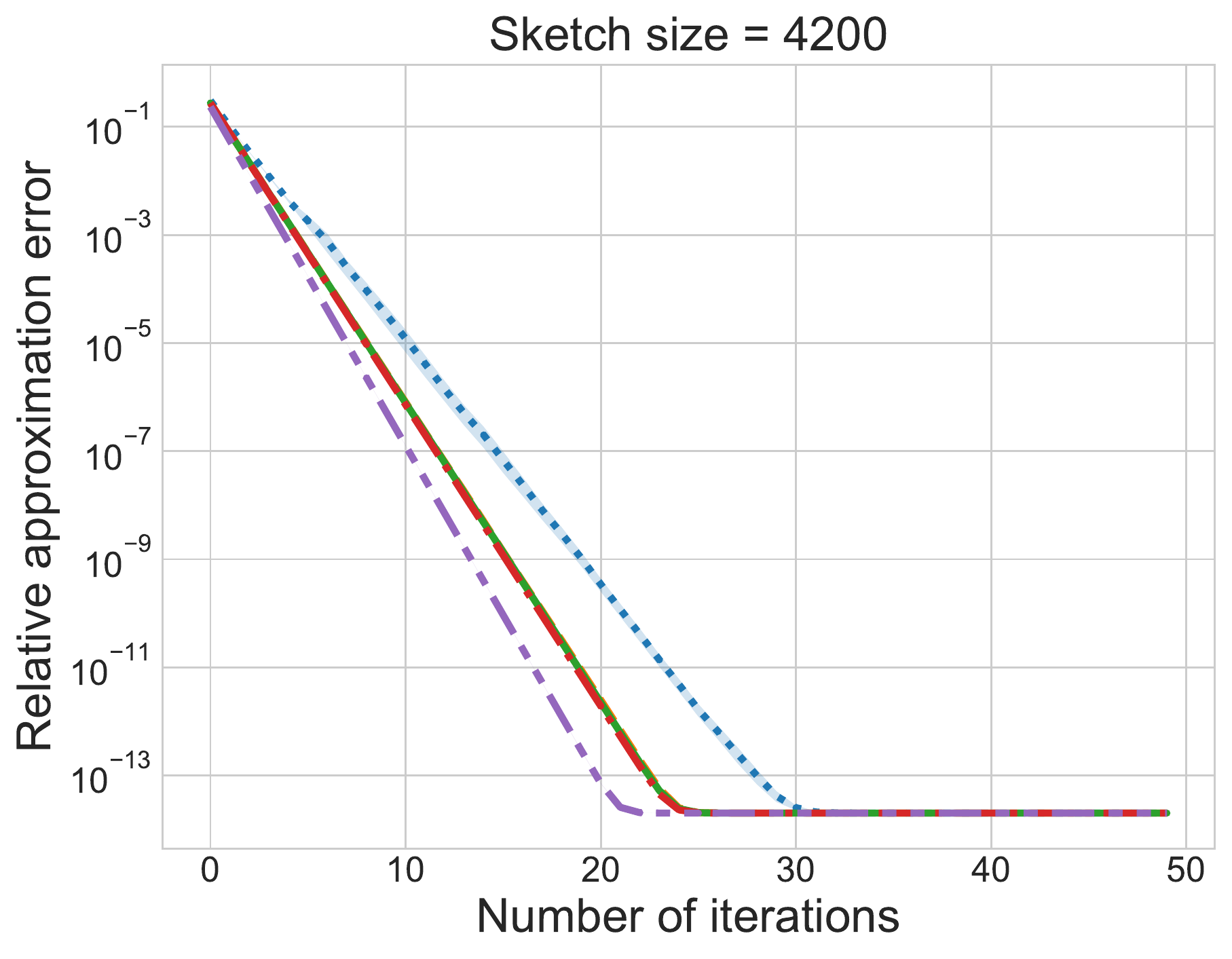}}
\subfloat[$s = 5000$]{
    \includegraphics[width=0.19\columnwidth,keepaspectratio]{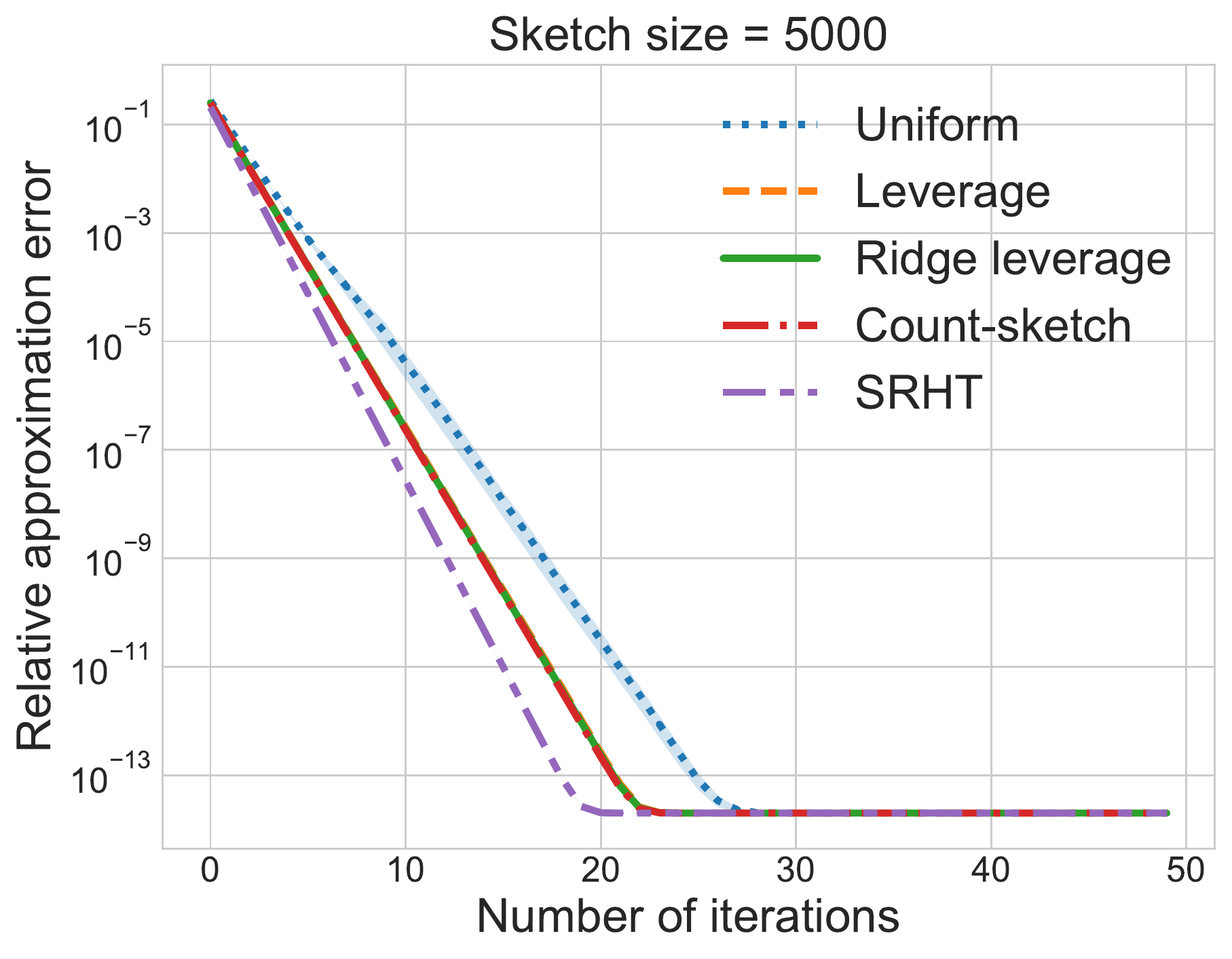}}
\caption{
Relative approximation error vs.~number of iterations on ORL dataset for increasing sketch size~$s$. \emph{Top row}: using a single sampling-and-rescaling matrix $\Sb$ throughout the iterations. \emph{Bottom row}: sample a new $\Sb_j$ at every iteration $j$.
Errors are on log-scale.}
\label{fig:appendix-ORL}
\end{figure*}

\begin{figure*}[htbp]
\centering
\subfloat{
    \includegraphics[width=0.19\columnwidth,keepaspectratio]{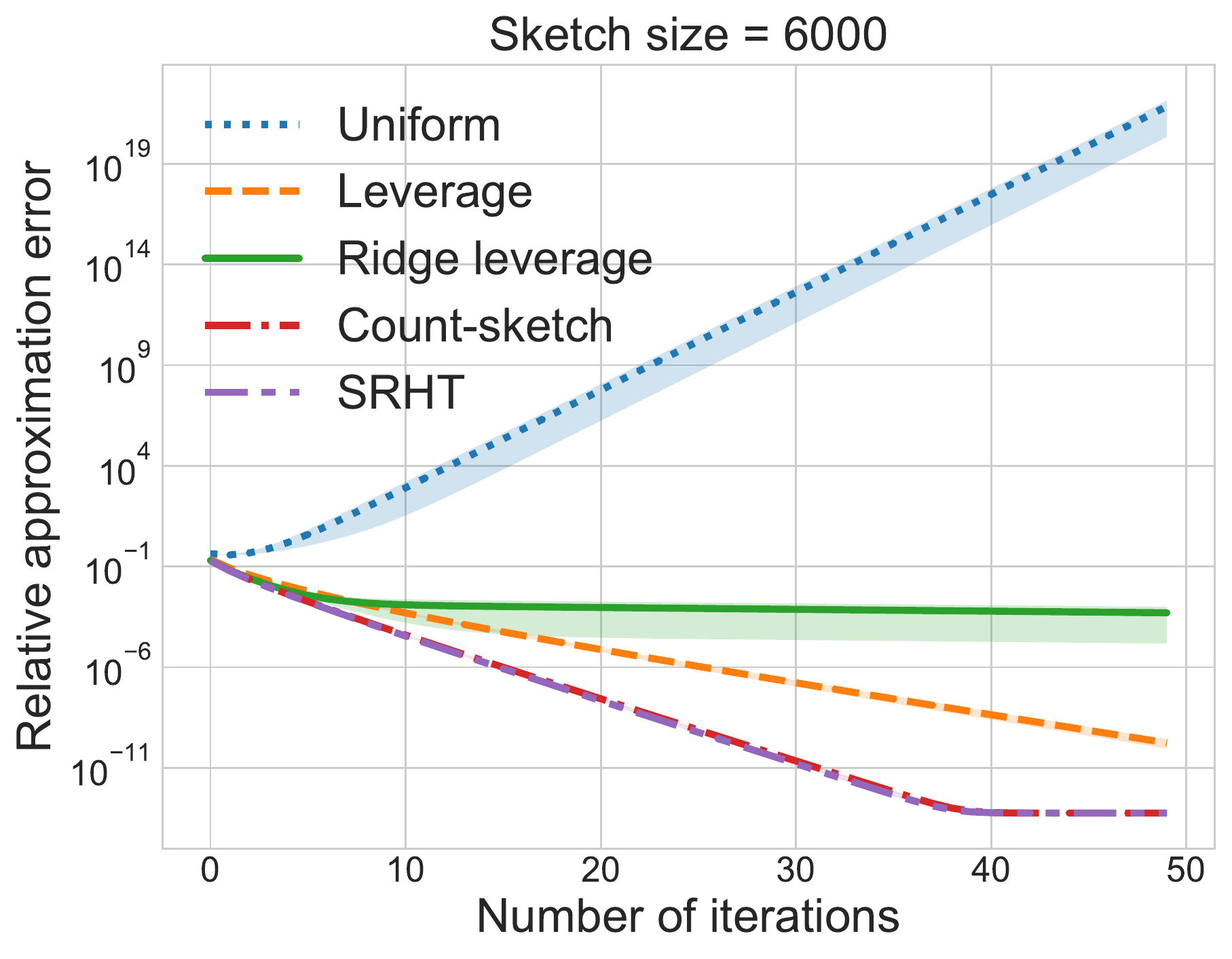}}
\subfloat{
    \includegraphics[width=0.19\columnwidth,keepaspectratio]{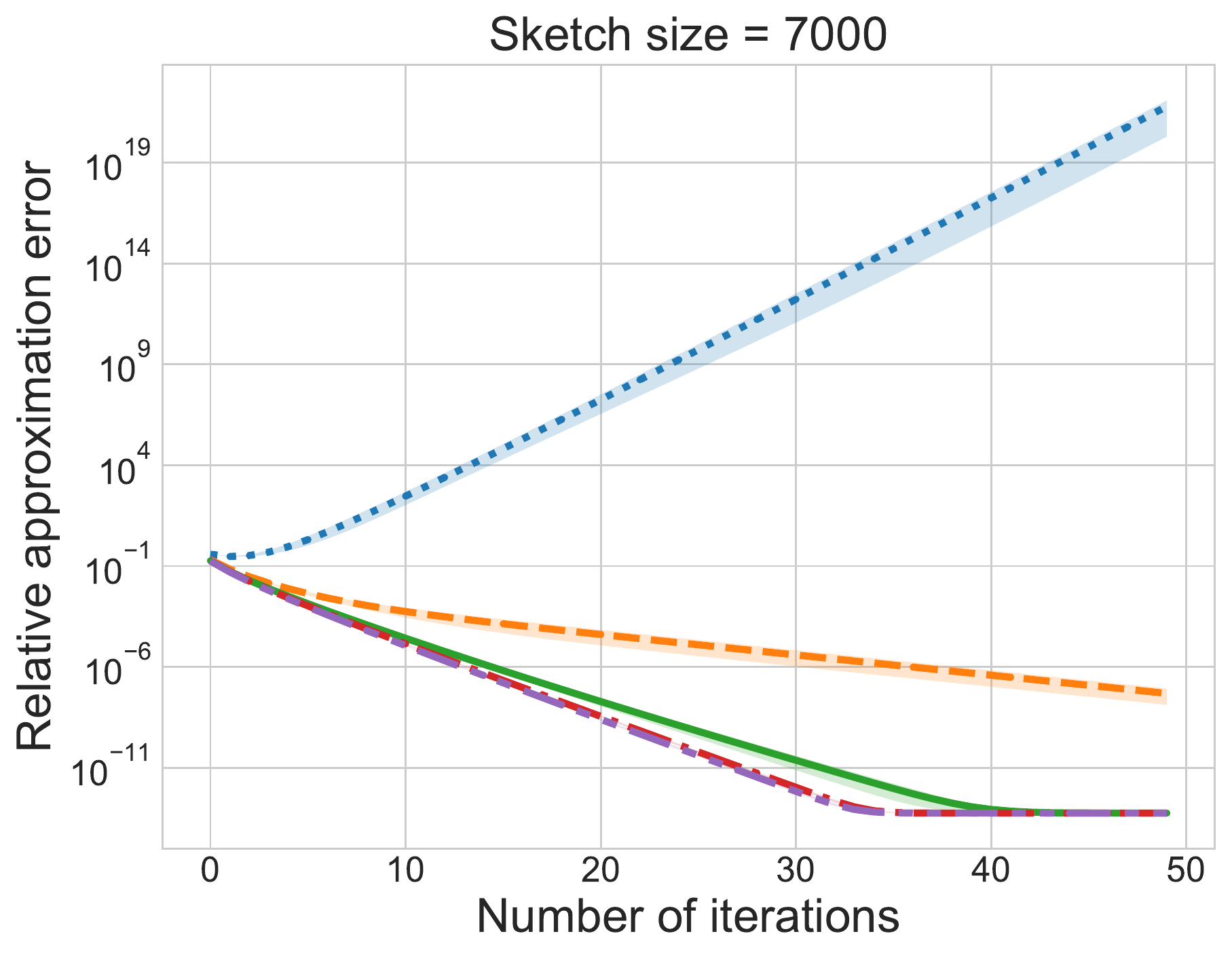}}
\subfloat{
    \includegraphics[width=0.19\columnwidth,keepaspectratio]{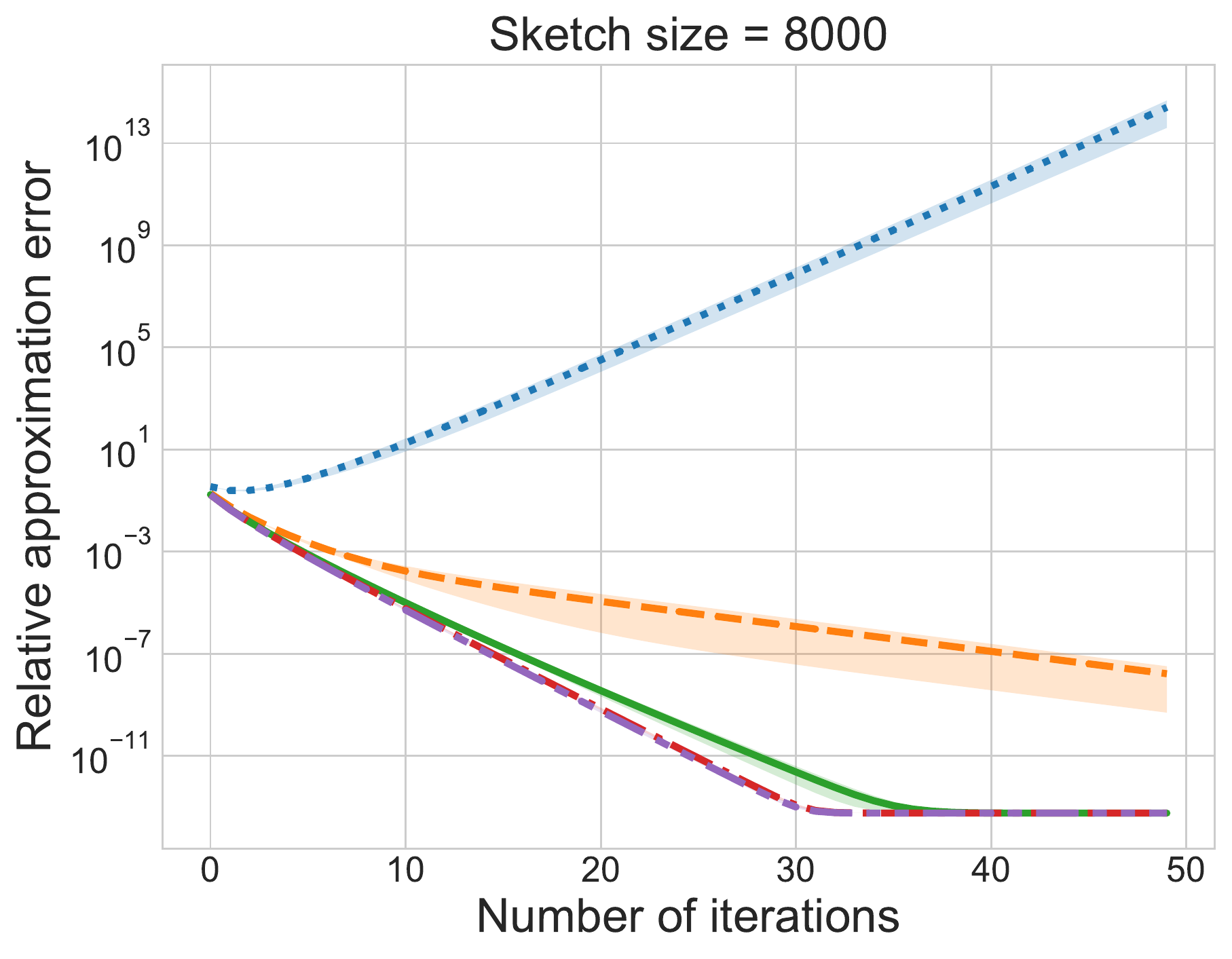}}
\subfloat{
    \includegraphics[width=0.19\columnwidth,keepaspectratio]{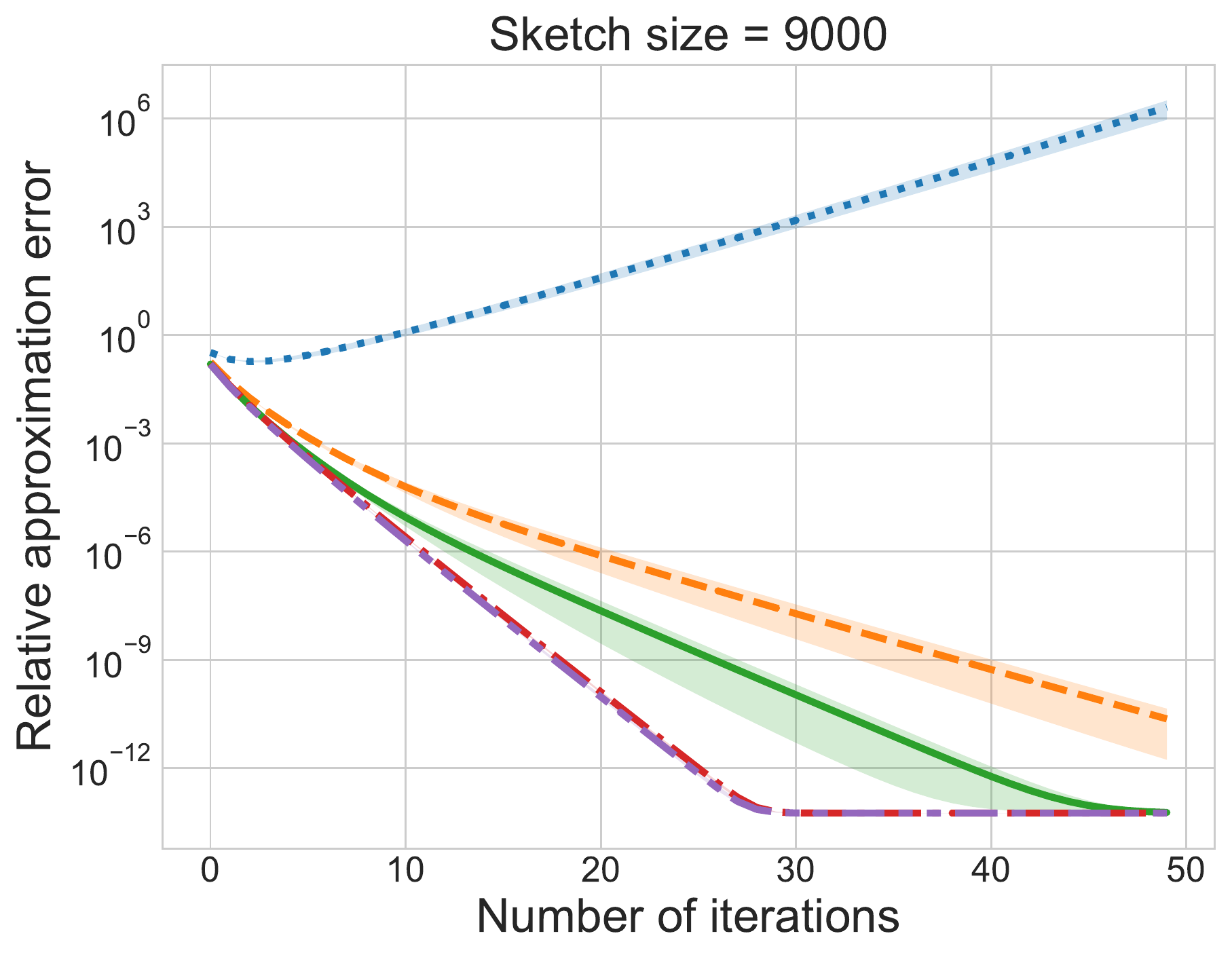}}
\subfloat{
    \includegraphics[width=0.19\columnwidth,keepaspectratio]{single-sketch/PEMS-iter-err-rel-lmbd10-ncol5}}
\vspace{-8pt}
\addtocounter{subfigure}{-5}
\subfloat[$s = 6000$]{
    \includegraphics[width=0.19\columnwidth,keepaspectratio]{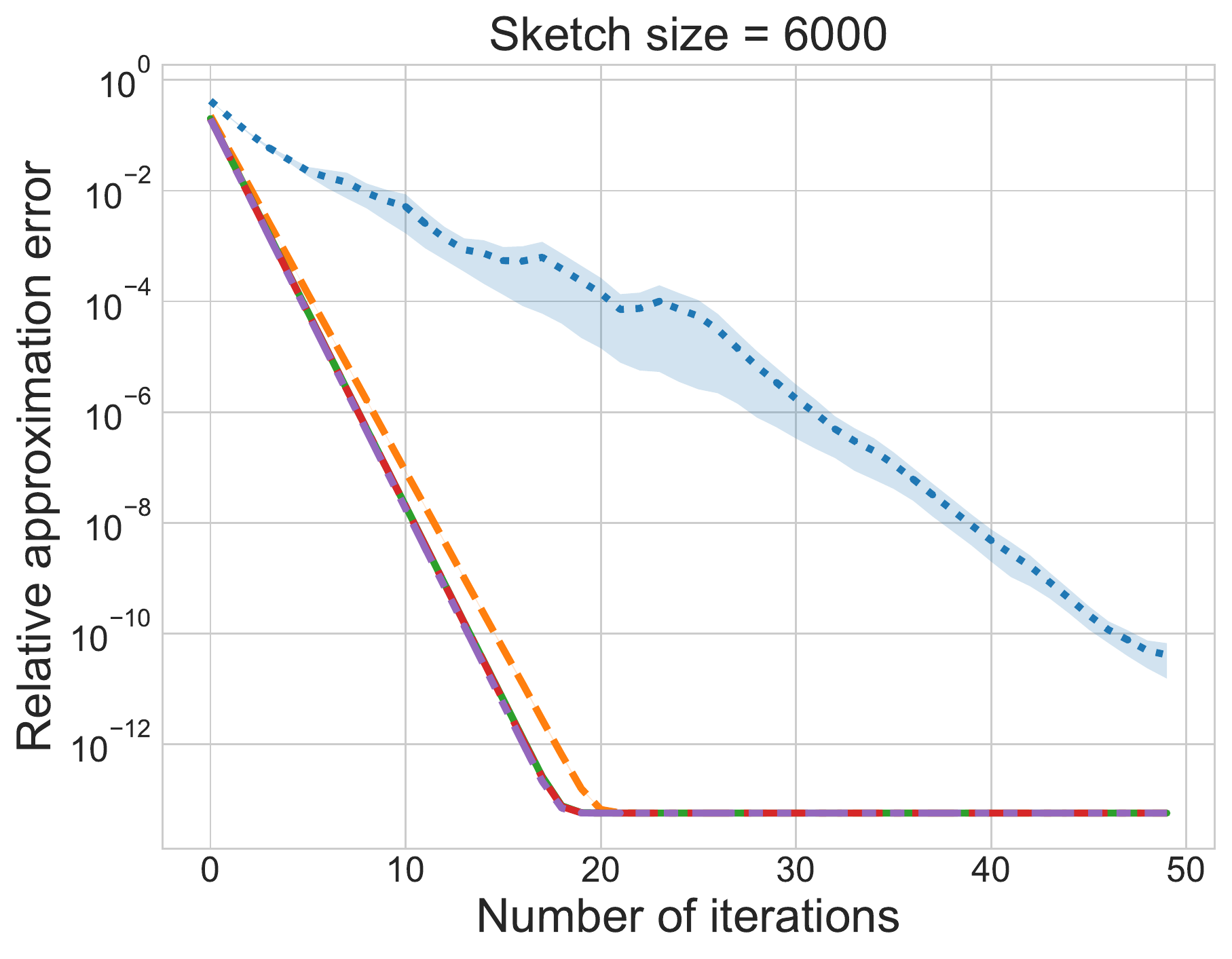}}
\subfloat[$s = 7000$]{
    \includegraphics[width=0.19\columnwidth,keepaspectratio]{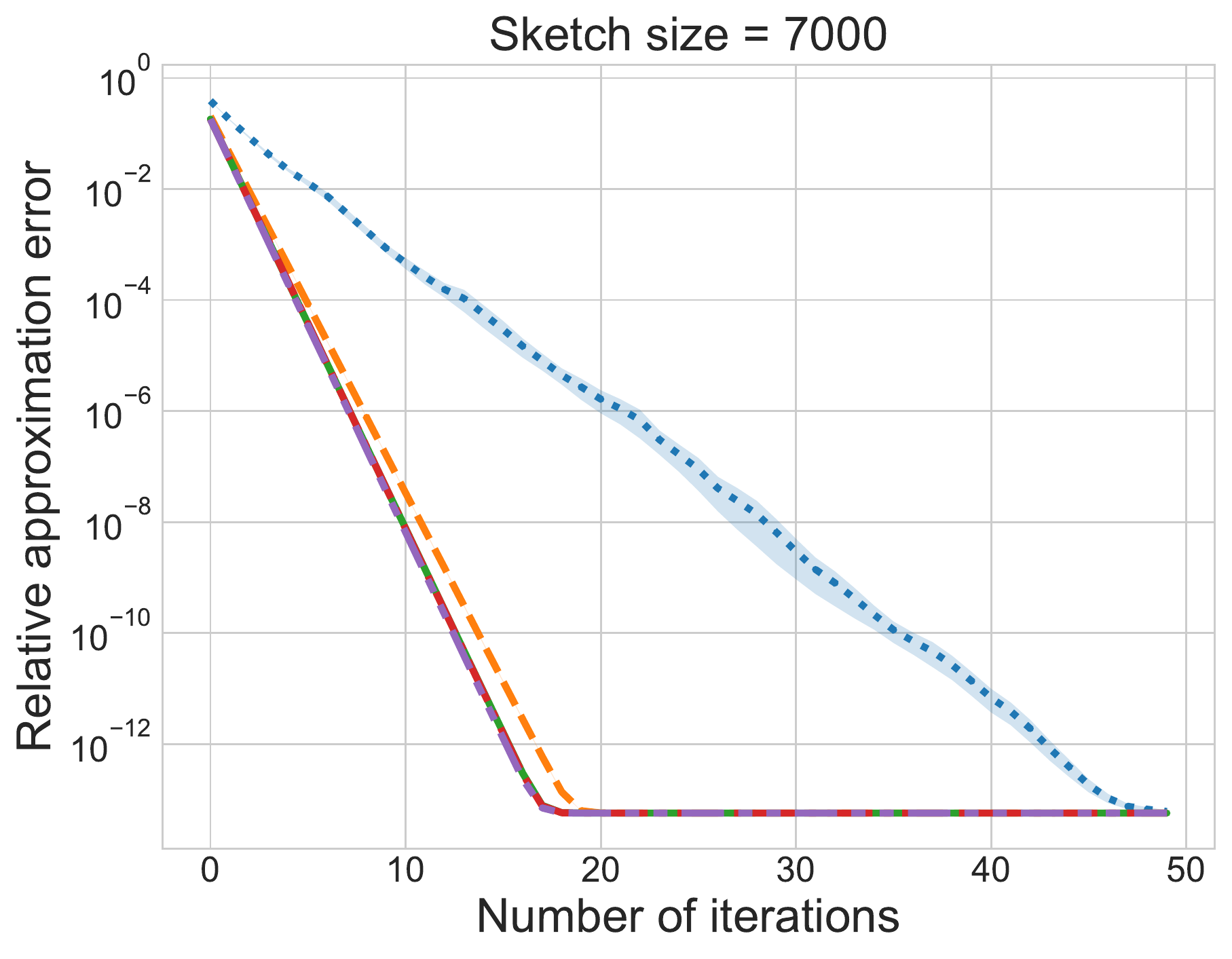}}
\subfloat[$s = 8000$]{
    \includegraphics[width=0.19\columnwidth,keepaspectratio]{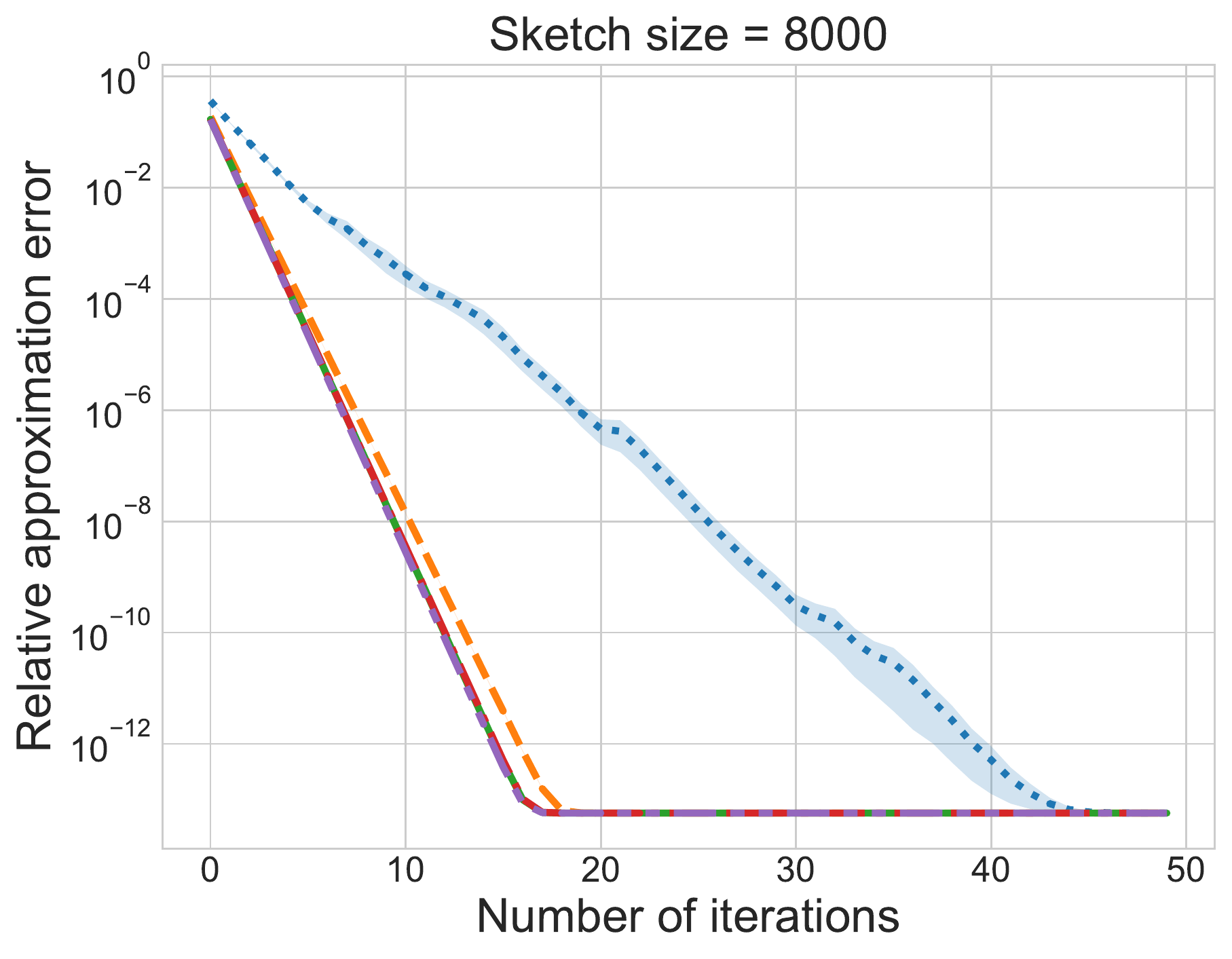}}
\subfloat[$s = 9000$]{
    \includegraphics[width=0.19\columnwidth,keepaspectratio]{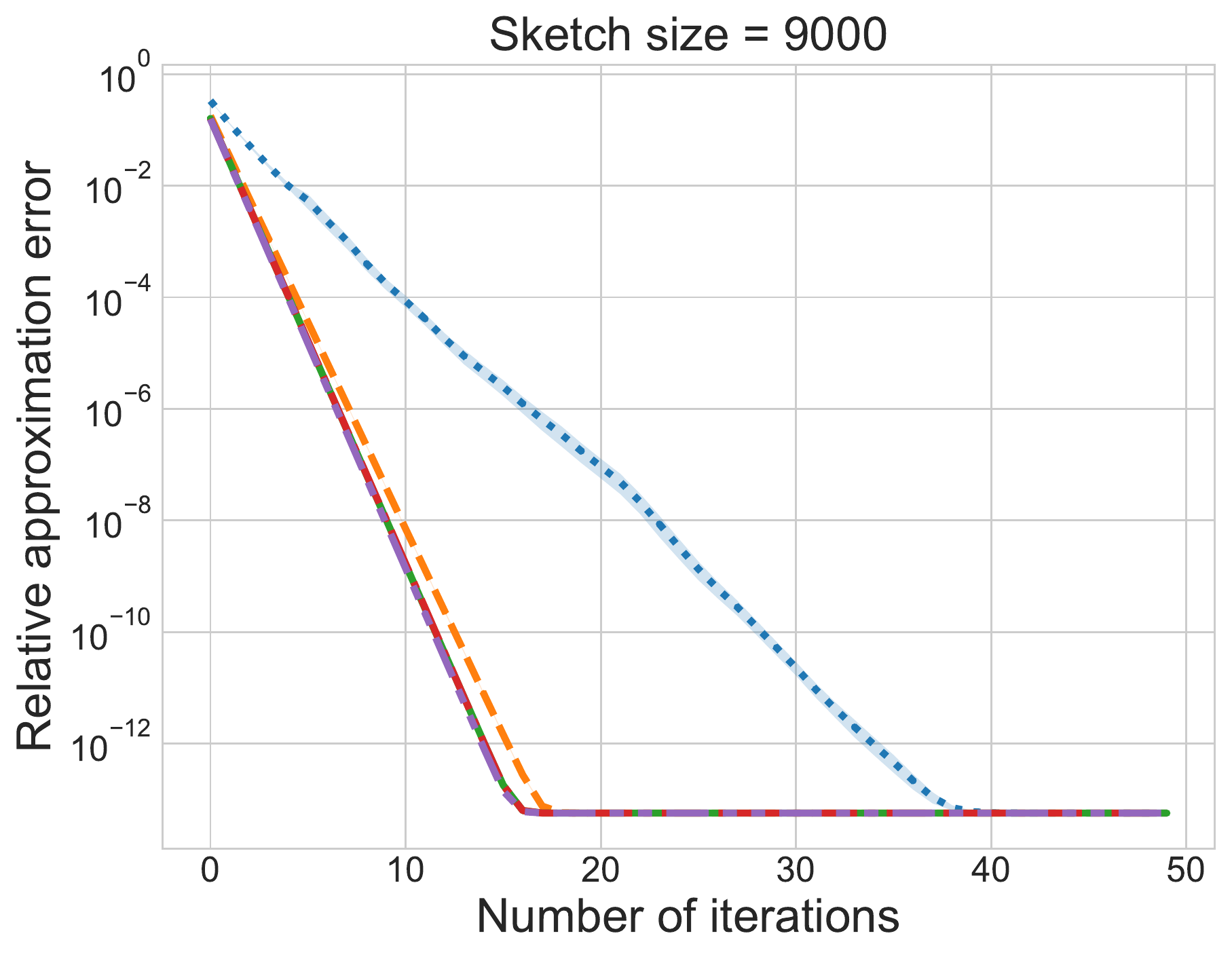}}
\subfloat[$s = 10000$]{
    \includegraphics[width=0.19\columnwidth,keepaspectratio]{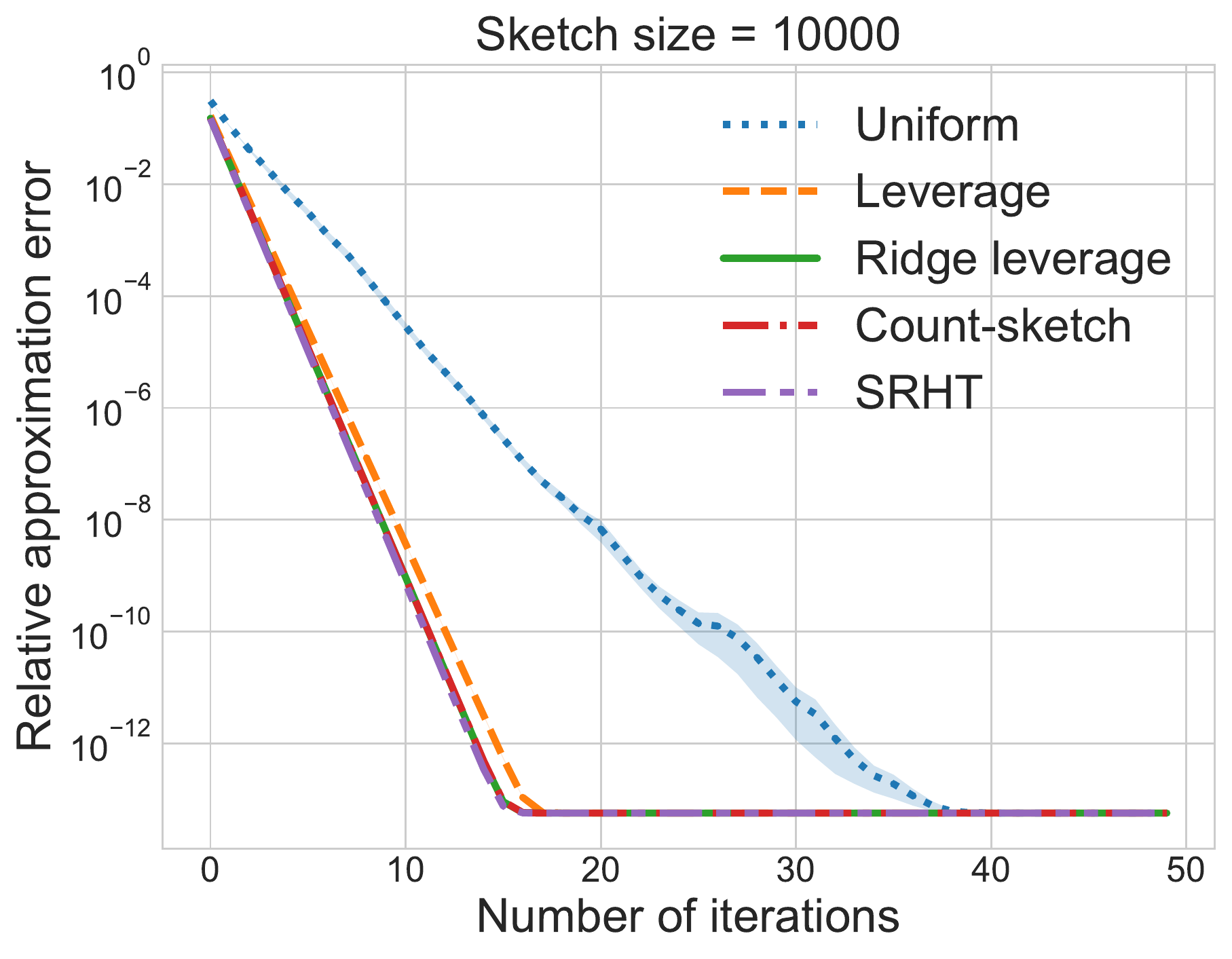}}
\caption{
Relative approximation error vs.~number of iterations on PEMS for increasing sketch size~$s$. \emph{Top row}: using a single sampling-and-rescaling matrix $\Sb$ throughout the iterations. \emph{Bottom row}: sample a new $\Sb_j$ at every iteration $j$.
Errors are on log-scale.}
\label{fig:appendix-PEMS}
\end{figure*}

\begin{figure*}[htbp]
\centering
\subfloat[ORL; single~$\Sb$]{
    \includegraphics[width=0.24\columnwidth,keepaspectratio]{single-sketch/ORL-size-err-rel-lmbd10}}
\subfloat[ORL; multiple~$\Sb_j$]{
    \includegraphics[width=0.24\columnwidth,keepaspectratio]{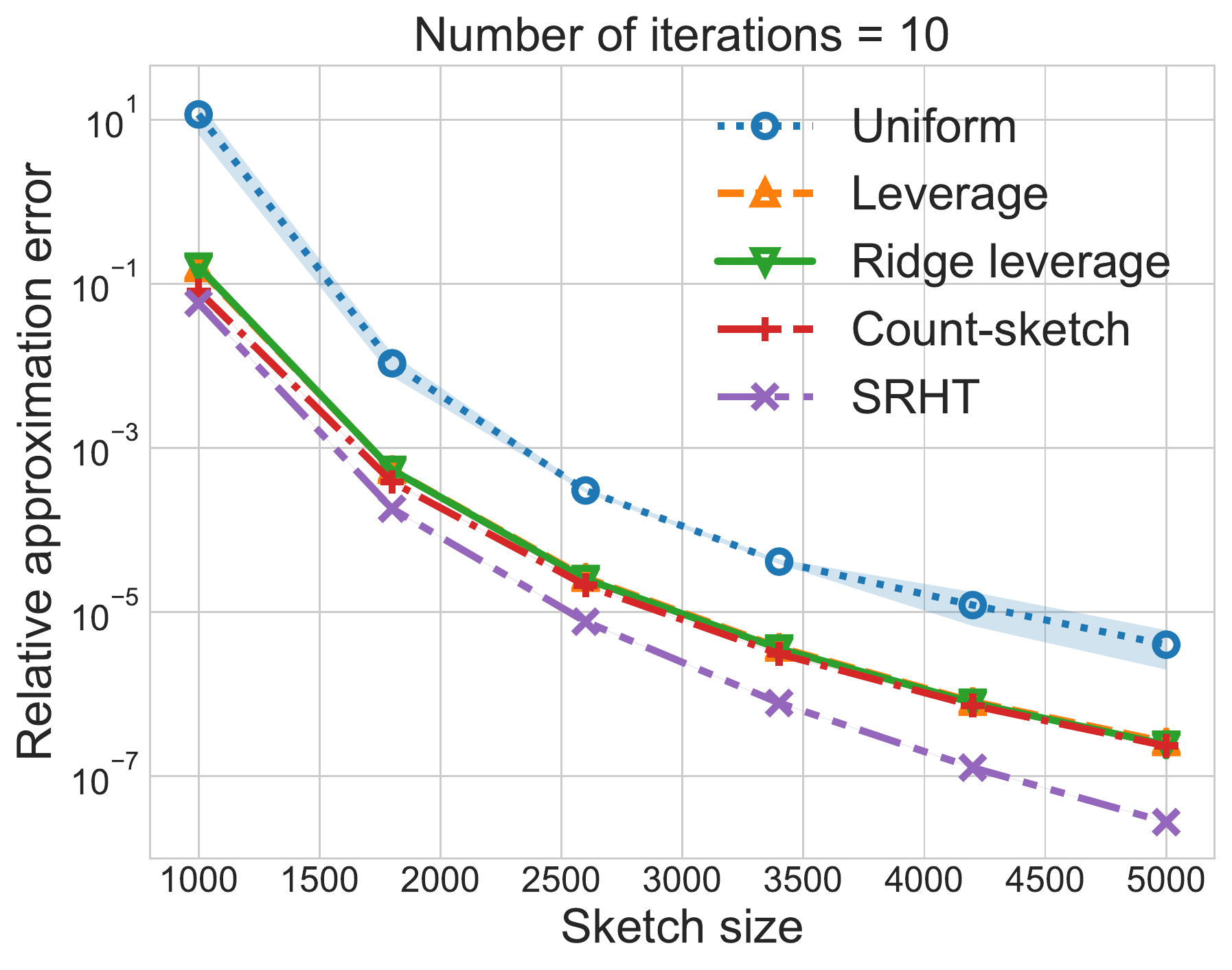}}
\subfloat[PEMS; single~$\Sb$]{
    \includegraphics[width=0.24\columnwidth,keepaspectratio]{single-sketch/PEMS-size-err-rel-lmbd10}}
\subfloat[PEMS; multiple~$\Sb_j$]{
    \includegraphics[width=0.24\columnwidth,keepaspectratio]{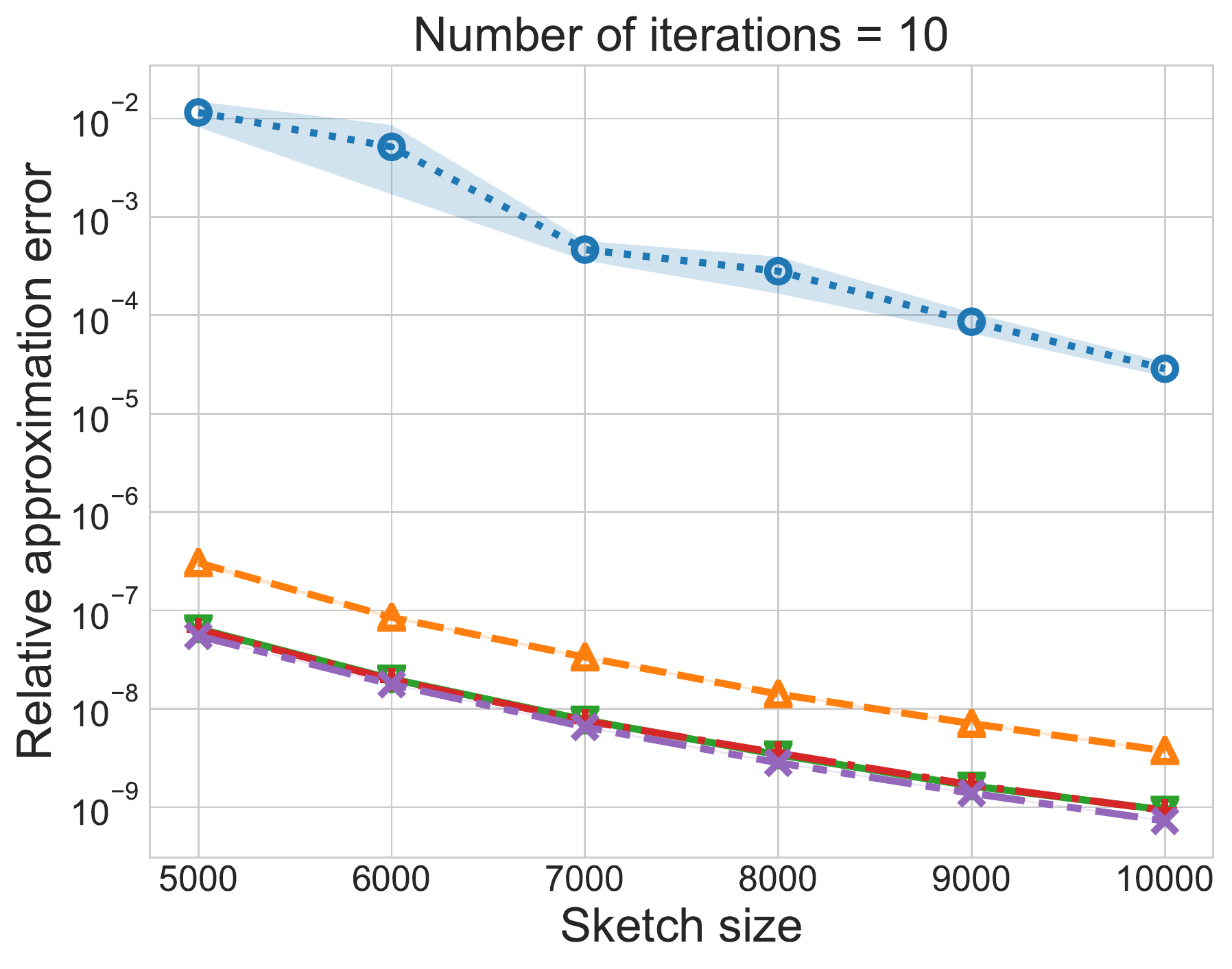}}
\caption{
Relative approximation error vs.~sketch size on ORL and PEMS after 10 iterations. \emph{Single~$\Sb$}: using a single sampling-and-rescaling matrix $\Sb$ throughout the iterations. \emph{Multiple~$\Sb_j$}: sample a new $\Sb_j$ at every iteration $j$. Errors are on log-scale; note the difference in magnitude of the approximation errors across plots.}
\label{fig:appendix-size}
\end{figure*}

Figures~\ref{fig:appendix-ORL}~and~\ref{fig:appendix-PEMS} show the relative approximation error vs.~number of iterations on the ORL and PEMS datasets for increasing sketch sizes.
Figure~\ref{fig:appendix-size} plots the relative approximation error vs.~sketch size after 10 iterations of Algorithm~\ref{algo:iterative_rda} were run.
We observe that using a newly sampled sketching matrix at every iteration enables faster convergence as the iterations progress, and also reduces the sketch size $s$ necessary for Algorithm~\ref{algo:iterative_rda} to converge.

%!TEX root = nips_LDA.tex

\end{appendices}

\end{document}